\documentclass[journal]{IEEEtran}
\usepackage{amsmath}
\usepackage{siunitx}
\usepackage{graphicx}
\usepackage{mdwtab}
\usepackage{booktabs}
\usepackage{multirow}
\usepackage{multicol}
\usepackage{xcolor}
\usepackage{subcaption}
\usepackage{amssymb}
\usepackage{hyperref}
\usepackage{cite}
\usepackage[justification=centering]{caption}
\pdfoutput=1

\usepackage[normalem]{ulem}
\usepackage[ruled,noresetcount,noend]{algorithm2e}
\usepackage{algpseudocode}
\usepackage{xcolor}
\definecolor{darkgreen}{rgb}{0,.4,0}
\definecolor{darkcyan}{rgb}{0,.4,.4}
\newcommand{\REMOVE}[1]%
          {{\color{red}\sout{#1}}}

\newcommand{\COMMENT}[1]%
          {{\color{darkgreen}\textbf{{PG:}} {#1}}}
          
\begin{document}
\title{DocBinFormer: A Two-Level Transformer Network for Effective Document Image Binarization}

\author{Risab~Biswas,
        Swalpa~Kumar~Roy,~\IEEEmembership{Member,~IEEE},
        Ning Wang,
        Umapada Pal,~\IEEEmembership{Senior~Member,~IEEE}, \\
        Guang-Bin Huang
        
\thanks{R. Biswas is with Optiks Innovations Pvt. Ltd (P360), Mumbai, Maharashtra 400066, India (e-mail: risab.biswas@p360.com).}

\thanks{S. K. Roy is with the Department of Computer Science and Engineering, Alipurduar Government Engineering and Management College, West Bengal 736206, India (e-mail: swalpa@cse.jgec.ac.in).}

\thanks{N. Wang is with Chongqing College of Mobile Communication, China (e-mail: lemon-807@foxmail.com).}

\thanks{U. Pal is with the CVPR Unit, Indian Statistical Institute, Kolkata, West Bengal 700108, India (e-mail: umapada@isical.ac.in).}

\thanks{G. B. Huang is with the School of Electrical and Electronic Engineering, Nanyang Technological University,  50 Nanyang Ave, Singapore 639798 (e-mail: egbhuang@ntu.edu.sg).}
}

\markboth{Preprint Submitted to Arxiv}%
{Biswas \MakeLowercase{\textit{et al.}}: DocBinFormer: A Two-Level Transformer Network for Effective Document Image Binarization}

\maketitle
\begin{abstract}
In real life, various degradation scenarios exist that might damage document images, making it harder to recognize and analyze them, thus binarization is a fundamental and crucial step for achieving the most optimal performance in any document analysis task. We propose DocBinFormer (Document Binarization Transformer), a novel two-level vision transformer (TL-ViT) architecture based on vision transformers for effective document image binarization. The presented architecture employs a two-level transformer encoder to effectively capture both global and local feature representation from the input images. These complimentary bi-level features are exploited for efficient document image binarization, resulting in improved results for system-generated as well as handwritten document images in a comprehensive approach. With the absence of convolutional layers, the transformer encoder uses the pixel patches and sub-patches along with their positional information to operate directly on them, while the decoder generates a clean (binarized) output image from the latent representation of the patches. Instead of using a simple vision transformer block to extract information from the image patches, the proposed architecture uses two transformer blocks for greater coverage of the extracted feature space on a global and local scale. The encoded feature representation is used by the decoder block to generate the corresponding binarized output. Extensive experiments on a variety of DIBCO and H-DIBCO benchmarks show that the proposed model outperforms state-of-the-art techniques on four metrics. The source code will be made available at {\url{https://github.com/RisabBiswas/DocBinFormer}}.
\end{abstract}

\begin{IEEEkeywords}
Document Image Binarization, Document Image Enhancement, Vision Transformer, and Deep Learning.
\end{IEEEkeywords}

\IEEEpeerreviewmaketitle
\section{Introduction}
\label{sec:Intro}
\IEEEPARstart{D}{ocument} image analysis studies have top priority for the preservation and readability of document images, particularly those with historical significance. In historical circumstances, document records typically include essential information that stretches back decades or even centuries \cite{megyesi2019decode}. Streaks, blots, artifacts, pen strokes, ink bleed, inadequate maintenance, aging effects, warping effects during acquisition, uneven lighting, etc. can all impede the conservation of document records. Owing to different kinds of degradation, binarization of document images is still a challenging task. The subsequent downstream data processing tasks, such as word segmentation, optical character recognition (OCR), information discovery, document layout analysis (DLA), etc., may be significantly impacted by these aberrations. Therefore, document image binarization, which is an image-to-image translation problem can be a good solution as a preprocessing step for the downstream tasks. Although this issue has been a longstanding challenge, its relevance and significance persist, especially when dealing with historical documents and various types of degradation. For instance, even state-of-the-art OCR engines like AWS Textract OCR face challenges with degraded images. Take an example of OCR on sample \#6 of the DIBCO 2012 dataset. The OCR results on sample \#6, Line\#5 for GT, degraded and binarized images are - L5 GT OCR - \textit{``Jack. Has he tried the flying machine''}, OCR L5 (Degraded Input) - \textit{``Jack. Has he tried the \textcolor{red}{jflying} machine \textcolor{red}{E}''}, OCR L5 Binarized (Ours) - \textit{``Jack. Has he tried the flying machine''}, OCR L5 Binarized (DE-GAN\cite{souibgui2020gan}) - \textit{``Jack. Has he tried the \textcolor{red}{jflying} \textcolor{red}{-} machine''}. The part indicated in \textcolor{red}{red} shows the wrong OCR output. This highlights how document binarization can substantially improve OCR accuracy. Additionally, the continuous release of more challenging datasets by ICDAR underscores the ongoing relevance and complexity of the document binarization problem, making it a crucial research area for document analysis and recognition. As the demands for accurate document processing continue to grow, developing and advancing robust document binarization techniques remains a pivotal research goal to ensure the effective handling of a wide range of document types and conditions.

Early document binarization approaches included a variety of thresholding techniques~\cite{otsu1979threshold}, \cite{sauvola2000adaptive}, including identifying one (or more) suitable thresholds for categorizing pixels in the image as pertaining to the foreground or background. However, these techniques faced technical limitations related to their sensitivity to variations in document lighting conditions, noise, and complex backgrounds, and they were not robust, often resulting in suboptimal binarization results. Moreover, these methods could not be easily generalized for different types of document degradations, further limiting their applicability in diverse document processing scenarios. To overcome these problems, researchers have turned to deep learning. They use neural networks to understand the complex non-linear relationship in degraded/clean document images. The most popular neural networks for this task are the Convolutional Neural Networks or CNNs~\cite{lecun1995convolutional}. These networks have become the go-to choice in the field of document binarization. While CNN-based models have shown promising performance improvements, they have many limitations. One of the limitations is that the convolutional operation they use is not ideal for capturing complex and long-range relationships within a document. In document binarization, it's crucial to understand information from distant parts of the document because text in degraded documents can be stretched out with different sizes, shapes, and textures. CNNs struggle with this, as they primarily focus on local information. Generative adversarial networks (GAN) \cite{goodfellow2014generative}, especially contrastive GAN (cGAN) are also considered a good solution for image generation problems. However, one significant challenge with GAN based approach is the potential for mode collapse, where the generator produces a limited diversity of outputs, failing to explore the full range of possible image variations. This can result in a lack of diversity and richness in the generated images, limiting their practical applicability in certain scenarios. Additionally, there can be information loss during the generative process, where fine-grained details and subtle variations in the original data may not be fully preserved in the generated images. Techniques like regularization and architectural improvements are often used to mitigate these issues, but some degree of information loss is inherent in the generative process of GANs. Because of the recent accomplishment of transformers in natural language processing (NLP) \cite{vaswani2017attention}, their potential to address issues concerning computer vision has gained traction. The self-attention mechanism aids in capturing global relationships among contextual features. 

Despite the promising performance of ViT, one key area for improvement is the handling of spatial hierarchies and local features. Unlike CNNs, which operate directly on pixel arrays and are inherently capable of capturing local features through convolutional kernels, ViTs divide an image into fixed-size patches and process them independently. While the self-attention mechanism helps in capturing long-range dependencies, ViTs may struggle with efficiently encoding fine-grained spatial information and local context, particularly when dealing with objects or patterns at varying scales within an image. This limitation potentially hinders their performance in tasks such as document binarization where precise spatial relationships and local details are crucial. Consequently, addressing these challenges to enhance ViT's ability to handle spatial hierarchies and local features remains an active area of research and a potential avenue for improvement. To address this challenge, many hybrid approaches have been proposed that combine CNN and ViT such as Twins~\cite{chu2021twins}, Focal Transformer~\cite{yang2021focal}, which has gained attention as a potential solution to address the challenges associated with ViTs handling of spatial hierarchies and local features. By leveraging the strengths of CNNs in capturing local patterns and fine-grained details, and the global context modeling capabilities of ViTs through self-attention mechanisms, this hybrid model attempts to strike a balance between local and global feature extraction. However, while this hybrid approach shows promise and CNNs are undoubtedly good for local feature extraction in image processing tasks, it may not always be the optimal choice depending on the task at hand, especially for tasks like document image enhancement that has very diverse modalities. CNNs typically employ a fixed receptive field, which may hinder their ability to capture local details effectively, especially when dealing with historical documents with varying text sizes and complex layouts. Additionally, the pooling operations commonly used in CNN architectures can lead to a loss of fine-grained local information, potentially impacting the quality of binarization results. Finally, when striving for a comprehensive understanding of the document's global context or when dealing with complex document structures, CNNs may fall short, as their focus on local features might not adequately leverage global relationships. Therefore, while CNNs offer local information extraction strengths, their limitations must be carefully considered when selecting the most suitable architecture for the task of document image binarization.

The proposed architecture is inspired by the idea of incorporating both local and global features in image processing through self-attention mechanisms, with the goal of enhancing binarization results. To achieve this, we propose an approach that leverages attention mechanisms to extract both global and local features from images. This approach addresses the challenge of recovering local information from images, especially when reducing patch sizes, and underscores the importance of local patch attention for constructing high-performance visual transformers. During the encoding stage, we generate local representations for each patch and combine them with the global representation of the input image. Our proposed architecture has achieved state-of-the-art results in addressing document image binarization challenges, as demonstrated by our research findings. The presented approach diverges from TNT~\cite{han2021transformer} in the way it efficiently processes and aggregates global and local information. While taking inspiration from TNT, we were motivated to redesign and adapt them for low-level vision tasks like document image binarization. This adaptation was driven by the need to overcome the limitations associated with TNT, particularly its computational inefficiency, making our proposed TL-ViT, known as \textbf{DocBinFormer}, a unique and effective solution for document image binarization.

In the proposed network, the degraded input image is first split into patches before entering the encoder section. The image patches are further broken down into sub-patches. Each image patch is attached to a latent token representation during the encoding process. Post the encoding process, the decoder produces enhanced image patches which are finally rearranged to create the final binarized output image. In this methodology, the transformer encoder performs the feature extraction using the multi-head self-patch and sub-patch attention mechanisms. Unlike the previously proposed models for document image binarization using a vision transformer, which was usually limited to extracting only the global contextual features while not being able to retrieve the intricate details that can only be captured by a local attention mechanism. The architecture of \textbf{DocBinFormer} ensures that the patch's local structure is not corrupted by ViT's simple use of a conventional transformer to process the patch sequence. Instead, we employed TL-ViT to discover both the patch and the sub-patch information from an image. Using sub-patches in a two-level transformer architecture can be a better choice for capturing local information compared to a standard CNN in scenarios like document binarization. While CNNs are indeed widely used and effective, there are specific reasons why utilizing attention mechanism at a sub-patch level would be more advantageous for local information capture. The fixed-size convolutional kernels in CNNs might be too large to capture fine details, or too small to consider the broader local context in an image. Moreover, CNNs can capture local information at one fixed scale, and it might not be optimal for all parts of an image. The static receptive fields are also uniform across all regions of an image with parameters being shared across the entire image, which might limit their ability to effectively capture highly localized features. To overcome these challenges, we propose using the sub-patch level attention, as they are more effective than CNNs due to their dynamic adaptability and context-awareness. Unlike CNNs, attention mechanisms adjust their focus based on the specific content of the input data. The adaptability of the attention mechanism allows them to precisely target and capture intricate details that can exhibit significant variations in scale, context, and content within an image. It also aids in the understanding of how this sub-patch level feature representation fits into the broader image context. This adaptability becomes particularly crucial when dealing with document images containing variations in text size, font, background texture, etc.

Sub-patch level transformers too can adaptively focus on different regions within the document, adjusting their attention to the specific characteristics of text and background within the sub-patch. This also allows for the selection of an appropriate spatial granularity for local information capture. In the sub-patch transformer approach, you can have multiple levels of local processing, each with its own attention mechanism. This can facilitate the model to capture local details at different levels of granularity.  Sub-patches can adapt their receptive fields (the area of input they consider) dynamically, depending on the task and the content of the image.  Contrary to the CNN, the sub-patches of the local transformer encoder have their own attention mechanisms, which reduces the need for extensive parameter sharing. Therefore, the TL-ViT approach in document image binarization proves to be advantageous because it offers adaptability, multi-scale processing, and attention mechanisms that can better capture the local information essential for the accurate segmentation of text from the background.

The inclusion of patch-level attention combined with sub-patch encoded features is a novel and effective approach for achieving global-local interaction in the proposed work. This strategy allows us to capture both the broader, global context of an image and the fine-grained, local details simultaneously. While sub-patch transformers excel at capturing local intricacies, patch-level attention ensures that the model is aware of the overarching structures and relationships within the image. This synergy between global and local information processing is a key innovation in our approach, enhancing the model's ability to understand the interplay between different elements in the image. As a result, our work achieves a more comprehensive and nuanced understanding of the visual content, making it well-suited for a wide range of computer vision tasks where global-local interaction is essential for accurate interpretation and decision-making. This combined approach of patch-level attention and sub-patch processing proves highly successful for the document binarization problem due to its unique ability to address the specific challenges presented by degraded document images. By employing patch-level attention, the model gains an understanding of the global structure of the document, identifying regions containing text and non-text elements—a crucial aspect of accurate binarization in documents with complex layouts. Meanwhile, sub-patch processing enables the model to focus on the fine-grained details of the text, accommodating variations in text sizes, fonts, and styles. This adaptability is vital for precise text segmentation and maintaining text integrity. Additionally, the approach's capacity to handle noise, artifacts, and multi-scale text information, while retaining contextual awareness, results in more robust and accurate document binarization, making it particularly effective for this specialized task. Our experimental results provide compelling evidence to support the above hypothesis regarding the effectiveness of patch and sub-patch level feature interaction in our proposed method. These results demonstrate that our approach outperforms nearly all the CNN-based methods that were examined. This achievement underscores the significance of our novel approach and its ability to capture and leverage local and sub-local information, surpassing the performance of conventional CNN-based approaches.

Previous Vision Transformer (ViT)-based methods like DocEnTr\cite{souibgui2022docentr} have demonstrated limitations in effectiveness due to their primary focus on patch-level or global feature representation, neglecting the vital sub-patch-level details present in document images. These approaches struggle to capture fine-grained local information, adapt to variations in text size and style, and effectively handle intricate details within patches, which are essential for accurate text localization and recognition in tasks like document image binarization. Our proposed method addresses the limitations discussed above, enabling more precise text localization, improved adaptation to document variability, and enhanced context-aware decision-making, thereby surpassing the performance of previous ViT-based approaches.

The contributions of the proposed methodology to this end are summarized below:

\begin{itemize}
    \item We introduce \textbf{DocBinFormer}, an end-to-end, simple and yet effective binarization technique. To the best of our knowledge, this is the first two-level transformer architecture designed for document binarization where global and local feature interaction has been captured in this manner.
    
    \item The proposed network not only targets global features but also leverages the effectiveness of local information and combines them in a simple and efficient way, using the multi-head self-patch and sub-patch attention which is one of the novelties of this work. The architecture utilizes the attention mechanism instead of a ViT-CNN hybrid approach.
    
    \item In the proposed TL-ViT network at \textit{Level 1}, we capture the patch-level feature representations, and at \textit{Level 2}, the model extracts information from the sub-patches, which provides a local representation of the patches, resulting in significantly precise binarization.
    
    \item The paper provides a thorough and detailed case study to demonstrate the significance of \textbf{DocBinFormer} architecture both from a qualitative and quantitative perspective. Extensive experiments are conducted on the DIBCO and the H-DIBCO datasets to validate the efficacy of the proposed model for document binarization in comparison to a variety of peer models. The experimental results show that the DocBinFormer model has promising performance advantages when binarizing documents with different degrees of degradation. The paper also includes an ablation study of the proposed transformer encoder network.
\end{itemize}

The remaining paper is organized as follows: In Section~\ref{sec:Works} we examine the related research for document image binarization and state-of-the-art. The proposed methodology is described in Section~\ref{sec:Method} where we illustrate the details of the proposed network. Section~\ref{sec:Exp} provides the experimental results and comparison, including the ablation studies. Finally, a comprehensive conclusion is drawn in Section~\ref{sec:Con}.

\section{Related Works}
\label{sec:Works}
In this section, we will discuss different techniques for document image binarization. We discuss the historical methods as well as the most recent developments in this field. The following subsections illustrate the different approaches in depth:

\subsection{Thresholding Based Techniques}
The earliest traditional approaches for document image binarization were based on thresholding. The use of these threshold values was to categorize the pixels in the document image as foreground (black) or background (white) \cite{otsu1979threshold}, \cite{sauvola2000adaptive}. For cases dealing with low-quality images, local adaptive approaches like \cite{niblack1985introduction}, and perform better. There are also deep learning-based techniques that are inspired by these thresholding techniques which we will see in the coming subsection. A hybrid technique proposed by \cite{gatos2009icdar} was among the first. Niblack's \cite{niblack1985introduction} approach is used to calculate the backdrop of an image using inpainting, followed by an image normalization to adjust the background fluxes. This adjusted image is then subjected to Otsu's approach to eliminating background noise. Finally, the two binarization outputs are integrated at the linked component level.

\begin{figure*}[ht]
    \centering
    \includegraphics[clip=true, trim = 00 00 00 00, width=1.02\textwidth]{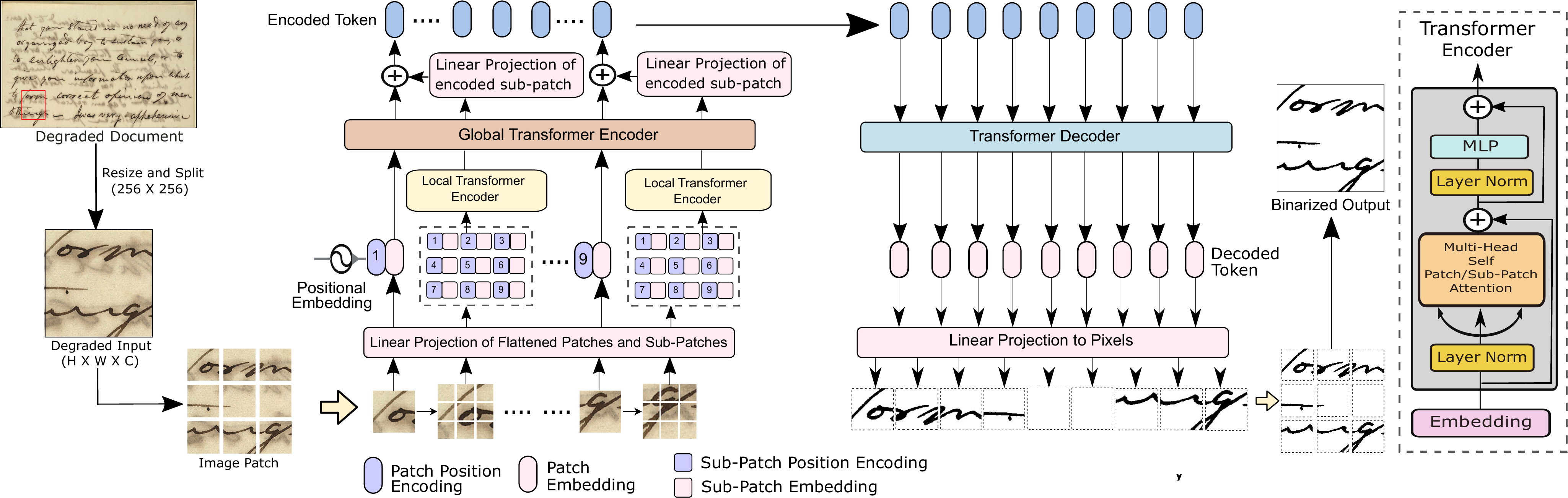} 
    \captionsetup{justification=centering}
    \caption{\centering Architecture of the proposed network for document image binarization. The encoding stage is shown on the left, and the decoding stage is shown on the right. The transformer encoder block is graphically illustrated as well.}
    \label{fig:TNTBinarization}
\end{figure*}

\subsection{Deep Learning Based Techniques}
Deep learning \cite{lecun2015deep} has found places in solving a wide variety of computer vision problems like large-scale image classification \cite{biswas2020identification}, object detection~\cite{zhao2019object}, image segmentation \cite{biswas2023polypsam}, and earth observation~\cite{roy2022multimodal} etc. Deep learning-based solutions were also employed to address the problem of document image binarization by learning the augmentation straight from raw data. Thresholding-based approaches have also been combined with the recent deep-learning networks to improve the quality of binarization. DeepOtsu \cite{he2019deepotsu}, and SauvolaNet \cite{li2021sauvolanet} are a few examples of the same. Pastor \textit{et. al}~\cite{pastor2015insights} utilizes CNNs for solving binarization problems. \cite{tensmeyer2017document} proposed a scalable fully convolutional network (FCN) for document image binarization. Furthermore, generative models (GAN) are also popularly used for various image restoration problems. A conditional GAN technique was described in \cite{souibgui2020gan} for several improvement tasks. A comparable cGAN-based technique was also presented in~\cite{zhao2019document}. DocDiff~\cite{Yang2023DocDiffDE} is the first diffusion-based framework created expressly for issues with document image enhancement, like document binarization and deblurring.

\subsection{Vision Transformer Based Techniques}
Transformers have been driving developments in deep learning applications in recent years. Transformer-based designs initially showed significant success in natural language tasks like translation and embedding, outperforming earlier sequence modeling techniques. This drives numerous efforts to use them for vision tasks such as classification \cite{dosovitskiy2020image}, object identification \cite{carion2020end}, document interpretation \cite{xu2020layoutlmv2}, multi-modal remote sensing \cite{roy2022multimodal}, and so on. Transformers have also been utilized for tasks like restoration \cite{liang2021swinir} and image dewarping \cite{li2021selfdoc}. However, the architectures proposed in most of the techniques continue to rely on CNN as a feature extractor before proceeding to the transformer step. CNN might also recreate the resultant image. Souibgui \textit{et al.}~\cite{souibgui2022docentr} proposed a fully transformer-based approach for document image enhancement, without the need for any CNN. Motivated by the capabilities and remarkable success of the vision transformer (ViT) framework, and with the aim to overcome the specific challenges posed by convolutional neural networks (CNNs), this research explores the potential of ViT-based models in the domain of document binarization. We introduce an innovative ViT-based model named \textbf{DocBinFormer}, designed to excel in the challenging task of binarizing degraded or complex documents.

\section{Proposed Methodology}
\label{sec:Method}
The proposed methodology is an efficient document image binarization framework that employs a two-level vision transformer (TL-ViT) based encoder-decoder architecture. The entire network can be divided into two stages - \textit{Encoding} and \textit{Decoding}, respectively. Fig.~\ref{fig:TNTBinarization} illustrates the architecture of the \textbf{DocBinFormer} network. The components of the architecture are discussed in detail from Subsection~\ref{sub_sec:Patch/Sub-Patch} to \ref{sub_sec:Loss/Optimization}:
\subsection{Patch / Sub-Patch and Position Embedding}
\label{sub_sec:Patch/Sub-Patch}
Let the degraded 2D input image be $\mathbf{X} \in \rm \mathbb{R}^{(H\times{W}\times{C})}$, where ($\rm H$,$\rm W$) is the spatial resolution of the image i.e. height and width, respectively and $C$ be the number of color channels. Since we are dealing with a binarization problem, we decided to choose a grey-scale input image, thus $\rm C$ = 1. Accordingly, the input image becomes $\mathbf{X}\in \mathbb{R}^{(\rm H\times{W}\times{1})}$. In the beginning, the input image is split into $\mathbf{n_{patch}}$ number of non-overlapping patches and then flattened into a latent embedding space using a learnable linear projection. The patch representation matrix can be illustrated below:
\begin{equation}
   \rm \mathbf{X_{patch}} = [x^{(1)},x^{(2)},\ldots,x^{(n_{patch})}] \in \mathbb{R}^{(n_{patch}\times{(p^2)}\times{1})}
\end{equation}
where, $\mathbf{X_{patch}}$ is the patch embeddings matrix, ($\rm p\times p$) is the resolution of each patch, and $\rm n_{patch} = {(H\times{W})/{p^2}}$ is the total number of patches, which also serves as the effective input sequence length and let $\rm{D}$ be the dimension of each patch embedding. This is similar to the patch representation mentioned in the ViT \cite{dosovitskiy2020image}.

Similarly, each patch is further divided into $\mathbf{n_{sub\mathunderscore patches}}$ number of sub-patches, each of size ($\rm s\times s$) with $\rm s$ being a hyper-parameter and its variational impact on the model's performance is exemplified in the Section~\ref{sub_sec:Ablation}. The sub-patches are thereafter flattened into a latent embedding space using a learnable linear projection to create the sub-patch embeddings, $\mathbf{X_{sub\mathunderscore patch}} \in \mathbb{R}^{\rm (n_{sub\mathunderscore patch}\times{(s^2)}\times{1})}$ and $\rm{D'}$ is the dimension of each sub-patch embeddings, where $\rm s<<p$. Therefore, we can say a patch is composed of a sequence of sub-patches:
\begin{equation}
\label{ref:patch}
\mathbf{x}^{\rm (i)} \rightarrow [\rm x^{(i,1)},x^{(i,2)},x^{(i,3)},\ldots, x^{(i,n_{sub\mathunderscore patches})}]
\end{equation}
where $\rm x^{(i,j)}\in \mathbb{R}^{(s\times{s}\times{1})}$ is the $\rm {j^{th}}$ sub-patch of the $\rm {i^{th}}$ patch of the input image $\mathbf{X}$, and $\rm j$ = 1,2,$\ldots$,$\rm n_{sub\mathunderscore patches}$. 

To simplify this idea, consider $\rm x^{(1)}$ as the $1^{\rm st}$ patch of the input image, $\mathbf{X}$ and $\rm x^{(1,1)}$ be the $1^{\rm st}$ sub-patch of the $1^{\rm st}$ patch. The sequence of patch-level representations is stored in the patch embedding whereas, the pixel-level representation is stored in the sub-patch matrix. Since we are dealing with the binarization problem but not the classification task thus ignoring the widely used classification (CLS) token, ${\rm X^{CLS}}$.

Next, to ensure that the images preserve their positional information, positional embeddings are linearly added to both the sequence of image patches and sub-patches. The positional embedding vectors for both of them are randomly initialized and are learnable parameters. The equation given below shows the positional embeddings for the patch representations:   
\begin{equation}
\label{ref:patch_pos_embedding}
\mathbf{X'_{patch}} = {\rm X_{patch}} \oplus \rm{PE}_{Patch}
\end{equation}
where $\rm {PE}_{Patch} \in \mathbb{R}^{(n_{patch} \times \rm D)}$ and $\oplus$ represent the patch positional encodings and element-wise addition, respectively. Similarly, each sub-patch embedding also comprises a sub-patch positional embedding:
\begin{equation}
\label{ref:subpatch_pos_embedding}
\mathbf{X'_{sub\mathunderscore patch}} = {\rm X_{sub\mathunderscore patch}} \oplus \rm{PE}_{SubPatch}
\end{equation}
where $\rm {PE}_{SubPatch} \in \mathbb{R}^{(n_{sub\mathunderscore patch} \times \rm{D'})}$. The patch position encoding allows keeping the global spatial information while the sub-patch position encoding maintains the local relative features. If we do not offer the positioning information to the transformer, it will have no notion about the image sequence (which comes first and the images that follow it) and there is a chance of generating arbitrary images during the decoding stage that will have no relation with the input image. $\mathbf{X'_{patch}}$ and $\mathbf{X'_{sub\mathunderscore patch}}$ represent the patch and the sub-patch embeddings which comprises the positional encoding as well, shown in Eqs.~(\ref{ref:patch_pos_embedding}) and (\ref{ref:subpatch_pos_embedding}) respectively. These will be the input to the transformer encoder block, which we will discuss in Subsection~\ref{sub_sec:Encoder}.

\subsection{Multi-Head Self Patch and Sub-Patch Attention}
\label{sub_sec:Encoder}
The proposed model employs two levels of transformer encoder network for making better interaction between the local and global feature dependencies within a patch. In the first level, the transformer encoder processes the visual information extracted from each patch ($\mathbf{X'_{patch}}$) whereas in the second level, the transformer encoder utilizes the visual sub-patches ($\mathbf{X'_{sub\mathunderscore patch}}$) to extract the fine-grained details. As illustrated in the proposed network diagram, Fig.~\ref{fig:TNTBinarization}, the transformer block that processes the patches is termed the global transformer block, which is followed by the transformer block that is responsible for processing the sub-patches and is termed the local transformer block. Both the encoders have a similar internal structure of the self attention block where patches $\mathbf{X'_{patch}}$ and sub-patches $\mathbf{X'_{sub\mathunderscore patch}}$, shown in Eq.~(\ref{ref:patch_pos_embedding}) and (\ref{ref:subpatch_pos_embedding}) are the input to the global transformer block and the local transformer block, respectively. There might exist one or more transformer encoder blocks. Each transformer encoder comprises a multi-head self-patch or sub-patch attention \cite{vaswani2017attention} block to capture both global and local feature dependencies. We divide and discussed the whole processing into 3 sub-subsections, i.e., tokenization of the image patches (Subsection~\ref{subsubsection:Stage1}), tokenization of the image sub-patches (Subsection~\ref{subsubsection:Stage2}), and finally creating the encoded feature representation (Subsection~\ref{subsubsection:Stage3}).

\subsubsection{Global Transformer Encoder Block}
\label{subsubsection:Stage1}
{\bf Step 1:} At the beginning, the overall patch, $\mathbf{X'_{patch}}$ passes through a layer normalization layer (LN) and then enters the multi-head self patch-attention (MPA) block of the transformer encoder. The objective of this step is to capture the long-range contextual information from the image patches. In the self-attention module, the layer normalised patch matrix, $\mathbf{X''_{patch}}$ is linearly transformed into three parts, i.e., queries $\mathbf{Q}\in \mathbb{R}^{\rm (n_{patch} \times D_k)}$, keys $\mathbf{K}\in \mathbb{R}^{\rm (n_{patch}\times{D}_k)}$ and values $\mathbf{V}\in \mathbb{R}^{\rm (n_{patch}\times{D}_v)}$ where $\rm D_k$ is the dimension of $\mathbf{Q}$ and $\mathbf{K}$, and $\rm D_v$ is the dimension of $\mathbf{V}$, which are computed by the following equation: 
\begin{equation}
\mathbf{Q} = \rm X''_{patch}{W_Q}, \hskip 10pt \mathbf{K} = X''_{patch}{W_K}, \hskip 10pt \mathbf{V} = X''_{patch}{W_V}
\label{eqn:qkv}
\end{equation}
where, $\mathbf{W_Q}\in \mathbb{R}^{\rm (D\times{D_k})}$, $\mathbf{W_K}\in \mathbb{R}^{\rm (D\times{D_k})}$, $\mathbf{W_V}\in \mathbb{R}^{\rm (D\times{D_v})}$ are the parameter matrices.

{\bf Step 2:} Next, we use the following equation to compute the self-attention (SA) operation which will be utilised in Step 3:
\begin{equation}
\rm SA(\rm Q,K,V) = Softmax(\frac{{\rm Q}{K}^T}{\sqrt{\rm D_k}})V
\label{eqn:attn}
\end{equation}
Here, the scaling term ${\sqrt{\rm D_k}}$, negates the softmax function's modest gradients.

{\bf Step 3:} The MPA performs $\mathbf{m}$ self-attention operation (SA) in parallel, where $\mathbf{m}$ is also a hyperparameter. The output values of each self-attention head are concatenated and linearly projected to generate the final output. The following mathematical representation will illustrate the above step more clearly: 
\begin{equation}
\rm MPA(\rm {Q},{K},{V}) = Concat({\rm {SA}}^1,{SA}^2,\ldots, {SA}^{\rm m})\rm W_O
\end{equation}
where, 
\begin{equation}
\rm{SA}^{(\rm m)} = Attention(\rm Q W_Q, K W_K, V W_V)
\end{equation}
where, $\mathbf{W_O}\in \mathbb{R}^{\rm ({m}\cdot D_v\times{D})}$ is the projection matrix, the Attention operation is shown in Eq.~(\ref{eqn:attn}) and the other parameters are shown in Eq.~(\ref{eqn:qkv}). In the proposed transformer network, we chose ${\rm m}$ = 8, which means we have 8 self-attention heads working in parallel inside each transformer encoder block, the same is mentioned in the column, attention heads in Table~\ref{Tab:hyper}.

{\bf Step 4:} As graphically illustrated in Fig.~\ref{fig:TNTBinarization}, the output of MPA is a matrix, $\mathbf{Y_{patch}}$ of size $(\rm {n_{patch}} \times D)$. This is further passed through a linear layer (weights $\mathbf{W_l} \in \mathbb{R}^{\rm D \times \rm D}$) :
\begin{equation}
    \mathbf{Y_{patch}} = \rm MPA(\mathbf{X''_{patch}})\mathbf{W_l}
    \label{equ:mpa_out}
\end{equation}

Finally, the output of the multi-head self-patch attention network, containing the feature representation, $\mathbf{Y_{patch}}$ is passed through a layer normalization layer (\rm LN) followed by a multi-layered perceptron block (\rm MLP) to obtain the final output of the global encoder block, $\mathbf{Y'_{patch}} \in \mathbb{R}^{\rm(n_{patch} \times \rm D)}$ : 
\begin{equation}
    \mathbf{Y'_{patch}} = \mathbf{Y_{patch}} \oplus \rm MLP(LN(\mathbf{Y_{patch}}))
    \label{equ:enco_out}
\end{equation}

\subsubsection{Local Transformer Encoder Block}
\label{subsubsection:Stage2}
Similar to the global encoder block that processed the image patches, we will utilise the local transformer encoder block to process the image sub-patches. The input to this stage would be the sub-patch embeddings, $\mathbf{X'_{sub\mathunderscore patch}}$, shown in Eq.~(\ref{ref:subpatch_pos_embedding}). The objective of this step is to capture the local contextual information from the image sub-patches. We utilise the multi-head self sub-patch attention (MSPA) block inside the local encoder block to process the sub-patches. We follow the same steps to process the image sub-patches, as shown in Subsection \ref{subsubsection:Stage1}. The other components i.e. (LN) and (MLP) are exactly the same as prior. 
Similar to the previous steps, at the end of step 4, we will have the output of the local transformer encoder block as $\mathbf{Y'_{sub\mathunderscore patch}}$ containing the local feature representation of the patches, whose size is $(\rm n_{sub\mathunderscore patch} \times {D'})$.
\subsubsection{Combine Global and Local Contextual Information}
\label{subsubsection:Stage3}
The level 1 and level 2 encoders captured in the above steps contain the global and local feature dependencies independently. In order to get the most out of this architecture, we combine the local and global information at this final step of the encoding stage. This is achieved by adding the global and the local information captured from the patch and sub-patch embeddings of an image, respectively. At first, the output obtained from Subsection~(\ref{subsubsection:Stage2}), which represents the encoded sub-tokens, $\mathbf{Y'_{sub\mathunderscore patch}}$, is reshaped into the domain of the number of patches i.e. $\rm (n_{patch} \times {D'})$ by passing through a linear layer followed by a layer normalization. This step is essential so that the below addition is possible. An element-wise addition is performed between this reshaped matrix and the output obtained from Eq.~(\ref{equ:enco_out}):
\begin{equation}
    \mathbf{Y_{encoder}} = \mathbf{Y'_{patch}} \oplus \rm LN(MLP(\mathbf{Y'_{sub\mathunderscore patch}}))
    \label{equ:trans_out}
\end{equation}
$\mathbf{Y_{encoder}} \in \mathbb{R}^{\rm (n_{patch} \times \rm D)}$ shown in the Eq.~(\ref{equ:trans_out}), is the final encoded feature representation that contains both the patch-level as well as the pixel-level feature information. This will be passed to the second stage i.e. decoding, discussed below.

\subsection{Binarization using Transformer Decoder}
\label{sub_sec:Decoder}
At the end of Section~\ref{subsubsection:Stage3}, we have the extracted features from the degraded image in the form of encoded tokens, which will be processed by the decoder to generate the corresponding binarized output. In this section, we will discuss the decoding stage, which is also the final step of the proposed binarization framework. $\mathbf{Y_{encoder}}$, the encoded image representation becomes the input to the transformer block used in the decoder section as illustrated in Fig.~\ref{fig:TNTBinarization}. The decoder processes the encoded tokens in the same way the global transformer does in the encoding stage. The encoded tokens undergo the same steps as described in Subsection~\ref{subsubsection:Stage1}, during the decoding stage to obtain $\mathbf{Y_{decoder}}$. The output of the decoded patches is passed through a linear layer of dimension equal to that of the patch tokens in order to project them into the expected pixel values required to generate the final binarized output denoted by $\mathbf{Y_{output}}$. 

In the decoding stage, we need not be concerned about the sub-patches, since the purpose of the sub-patches was to extract the local information, which is already embedded in our encoded tokens at the end of Section~\ref{sub_sec:Encoder}. The output of this stage is the enhanced (binarized) pixel representation of each of the input images.  The binarized patches are finally rearranged to create the actual binarized output of the corresponding degraded input image.

\subsection{Loss and Optimization Function}
\label{sub_sec:Loss/Optimization}
In order to train the proposed model, we use \emph{Mean Square Error} (MSE) loss. Since the MSE gives more weight to the inaccuracies owing to the squaring component of the function, it is effective for assuring that the model does not comprise any outlier predictions with significant errors. For training the proposed model, we calculate the MSE loss between the output image obtained from the decoder i.e. the model output and their corresponding ground truth image. Let the ground truth (GT) be $\mathbf{{X}}_{\rm GT}$. Note that the GT images need not be divided into patches or sub-patches. The below equation illustrates the above calculation mathematically: 
\begin{equation}
\rm{MSE} = {\frac{1} {n}}\sum_{i=1}^{n}(\mathbf{{X}}_{GT}-\mathbf{Y'_{output}})^2
\end{equation}
For optimizing the model, we are using {Weight decay Adam Optimizer (AdamW)}\cite{kingma2014adam} as an optimization function. This can help prevent the over-fitting of the model.
\section{Experimental Analysis}
\label{sec:Exp}
\subsection{Dataset and Evaluation Metrics}
We employed a range of DIBCO and H-DIBCO tools for training, testing, and validation. We used DIBCO/H-DIBCO 2009~\cite{5277767}, 2010~\cite{5693650}, 2011~\cite{6065249}, 2012\cite{6424498}, 2013~\cite{6628857}, 2014~\cite{6981120}, 2016~\cite{7814134}, 2017~\cite{8270159}, 2018~\cite{8583809} \& 2019~\cite{8978205}. These comprise both printed and handwritten document images. For our experiment purpose, we use the leave-one-out cross-validation technique for evaluation. That means we consider all images of one particular (H-) DIBCO year as a testing dataset, keeping all remaining documents from other years as the training dataset. In the next subsections, we will evaluate the model numerically and qualitatively, as well as analyze its performance in various scenarios. We employ traditional metrics like - peak signal-to-noise ratio (PSNR), F-Measure (FM), pseudo-F-measure (Fps), and distance reciprocal distortion (DRD) to validate the performance of the binarisation model. These metrics assess the visual resemblance of the output binarised images with the ground truth ones. An ablation analysis was performed, details of which can be found in Section~\ref{sub_sec:Ablation}, for various settings of the hyper-parameters to validate the proposed two-level strategy's efficacy, and the settings yielding the highest PSNR values were used for evaluating the performance of the proposed model. In the coming sub-sections, we will evaluate the model both quantitatively and qualitatively and also analyze its performance in different scenarios. 

\subsection{Model Configuration and Implementation Details}
This section will cover some implementation specifics as well as the metrics that were used to evaluate the model's effectiveness. Each degraded image and its corresponding GT image are split into resized images of size ($256\times256\times1$) for training, testing, and validation purposes. We can change the magnitudes of this resize operation based on our experimental needs, e.g. ($256\times256$) or ($512\times512$). 
We ran extensive experiments by altering the hyper-parameters such as the number of heads in the self-attention layer, encoder dimension, sub-patch size, etc. The grid search technique is used to determine the final optimal value of hyper-parameters of the model. For our experimental analysis, we chose to divide the degraded image into ($16\times16$) patches and further performed splitting on each patch ($8\times8$) sub-patch. Our final model's hyperparameters are mentioned in Table~\ref{Tab:hyper}. The model’s training parameters are - optimizer = AdamW with learning rate = 1.5e-4, eps=1e-08, and weight-decay = 0.05. The batch size is kept at 16 and we trained all versions of the model for 200 epochs. All coding implementations for training and inference are carried out in PyTorch using a single NVIDIA A100 GPU with 80 GB VRAM.

 
\begin{table*}[!ht]
\centering
\caption{Hyper-parameter settings of the proposed model}
\footnotesize
\begin{tabular}{|l|c|c|c|c|c|c|}
\toprule
    \textbf{Transformer Block}   &\textbf{Patch Size}     &\textbf{Layers}        &\textbf{Dim}       &\textbf{Attention Heads}   &\textbf{MLP Dim}    &\textbf{Params} \\ \midrule
    Global Encoder     &$16\times16$     &6      &768     &8   &2048    &30.7~\rm{M}  \\
    Local Encoder    &$8\times8$        &4     &256      &8   &2048   &6.9~\rm{M}  \\
    Decoder   &$16\times16$  &1   &768  &1  &256 &38.1~\rm{M}  \\
    \hline
\end{tabular}
\label{Tab:hyper}
\end{table*}
 
\begin{table*}[ht!]
\centering
\caption{Comparative results of the proposed method and other state-of-the-art methods on the DIBCO 2009-2011 datasets.}
\footnotesize
\begin{tabular}{|l|c|c|c|r||c|c|c|r||c|c|c|r|}
\toprule
\textbf{Method} & \multicolumn{4}{c||}{\textbf{DIBCO 2009}} & \multicolumn{4}{c||}{\textbf{DIBCO 2010}} & \multicolumn{4}{c|}{\textbf{DIBCO 2011}} \\
\midrule
 & \textbf{PSNR$\uparrow$} & \textbf{FM$\uparrow$} & \textbf{Fps$\uparrow$} & \textbf{DRD$\downarrow$} & \textbf{PSNR$\uparrow$} & \textbf{FM$\uparrow$} & \textbf{Fps$\uparrow$} & \textbf{DRD$\downarrow$} & \textbf{PSNR$\uparrow$} & \textbf{FM$\uparrow$} & \textbf{Fps$\uparrow$} & \textbf{DRD$\downarrow$} \\
\midrule
Otsu \cite{otsu1979threshold} & 15.31 & 78.60 & 80.50 & 22.56 & 17.53 & 85.50 & 90.69 & 4.05 & 15.72 & 82.10 & 84.80 & 9.02 \\
Sauvola \cite{sauvola2000adaptive} & 16.32 & 85.00 & 89.50 & 6.93 & 15.86 & 74.00 & 83.39 &7.26 & 15.60 & 82.10 & 87.70 & 8.03 \\
Winner \cite{5277767} & 18.66 & 91.24 & - & - & 19.78 & 91.50 & 93.58 & - & 17.84 & 88.74 & - & 5.36 \\
Gib~\cite{8517161} & 19.26 & 92.50 &94.40 &2.41 & 19.14 & 90.00 & 92.88 &2.75 & 18.29 & 90.33 &93.82 &2.99 \\
RDD~\cite{6373726} & 19.41 & 93.02 & 94.61 & 2.64 & 19.78 & 91.36 & 93.18 & 2.42 & 17.71 & 87.83 & 90.24 & 4.66 \\
DD-GAN~\cite{De2020DocumentIB} & 19.73 & 93.25 &94.67  & 2.81 & 22.20 & 95.18 & 96.77 &1.27 & 20.39 & 93.69 & 95.35 &1.85 \\
Tensmeyer\textit{ et al.}~\cite{tensmeyer2017document} & 18.43 & 89.76 & 92.59 & 4.89 & 19.13 & 89.42 & 91.11 & 3.77 & 20.11 & 93.60 & 95.70 & 1.85 \\
Calvo-Zaragoza~\textit{ et al.}~\cite{calvo2019selectional} & 19.51 & 91.98 & 93.81 & 3.64 & 22.29 & 95.38 & 96.27 & 1.28 & 19.55 & 92.77 & 95.68 & 2.52 \\
cGANs~\cite{10.1016/j.patcog.2019.106968} & 20.30 & 94.10 &95.26  & 1.82 & 21.12 & 94.03 & 95.39 &1.58 & 20.26 & 93.81 & 95.70 &1.81 \\
Howe~\cite{Howe2013DocumentBW} & 20.43 & 94.04 &95.06  & 2.10 & 21.08 & 93.59 & 94.81 & 1.72 & 19.01 & 90.79 & 92.28 & 4.46 \\
DeepOtsu~\cite{HE2019379} & 20.59 & 93.97 &95.31  & 1.77 & 20.28 & 91.35 & 94.20 &2.36 & 19.90 & 93.40 & 95.80 &1.90 \\
DocEnTr \cite{souibgui2022docentr} & \textcolor{brown}{21.32} & \textcolor{brown}{95.50} & \textcolor{red}{96.70} & \textcolor{blue}{0.18} & \textcolor{brown}{21.90} & \textcolor{brown}{95.10} & \textcolor{brown}{96.85} & \textcolor{red}{0.13} & \textcolor{brown}{20.81} & \textcolor{brown}{94.37} & \textcolor{brown}{96.15} & \textcolor{brown}{1.63} \\
$\rm {D^{2}BFormer}$~\cite{wang2022novel} & \textcolor{red}{21.33} & \textcolor{red}{95.51} & \textcolor{brown}{95.93} & \textcolor{brown}{1.42} & \textcolor{red}{23.10} & \textcolor{red}{96.23} & \textcolor{red}{97.49} & \textcolor{brown}{0.95} & \textcolor{red}{21.27} & \textcolor{red}{94.82} & \textcolor{red}{96.62} & \textcolor{red}{1.56} \\
\midrule
\textbf{DocBinFormer} & \textcolor{blue}{22.18} & \textcolor{blue}{96.26} & \textcolor{blue}{97.69} & \textcolor{blue}{0.09} & \textcolor{blue}{23.46} & \textcolor{blue}{96.61} & \textcolor{blue}{97.53} & \textcolor{blue}{0.23} & \textcolor{blue}{22.14} &\textcolor{blue}{96.18} & \textcolor{blue}{97.51} & \textcolor{blue}{0.44} \\
\bottomrule
\end{tabular}
\label{Tab:ComparativeResults_2009-2011}
\end{table*}

\begin{table*}[ht!]
\centering
\caption{Comparative results of the proposed method and other state-of-the-art methods on the DIBCO 2012-2014 datasets.}
\footnotesize
\begin{tabular}{|l|c|c|c|r||c|c|c|r||c|c|c|r|}
\toprule
\textbf{Method} & \multicolumn{4}{c||}{\textbf{DIBCO 2012}} & \multicolumn{4}{c||}{\textbf{DIBCO 2013}} & \multicolumn{4}{c|}{\textbf{DIBCO 2014}} \\
\midrule
 & \textbf{PSNR$\uparrow$} & \textbf{FM$\uparrow$} & \textbf{Fps$\uparrow$} & \textbf{DRD$\downarrow$} & \textbf{PSNR$\uparrow$} & \textbf{FM$\uparrow$} & \textbf{Fps$\uparrow$} & \textbf{DRD$\downarrow$} & \textbf{PSNR$\uparrow$} & \textbf{FM$\uparrow$} & \textbf{Fps$\uparrow$} & \textbf{DRD$\downarrow$} \\
\midrule
Otsu \cite{otsu1979threshold} & 14.98 & 74.87 & 76.84 & 26.77 & 16.63 & 80.07 & 82.89 & 11.00 & 18.70 & 91.60 & 95.53 & 2.66 \\
Sauvola \cite{sauvola2000adaptive} & 16.89 & 80.88 & 86.90 & 6.51 & 17.02 & 82.68 &88.12  & 7.25 & 17.56 &84.18 & 88.47 & 4.95 \\
Gib~\cite{8517161} & 19.34 & 90.99 &92.75 &3.09 & 19.58 & 91.14 &94.75 &2.77 &19.93 & 94.00 &96.48 &1.79 \\
RDD~\cite{6373726}  & 19.55 &89.76 & 89.61 & 4.19 & 19.59 &87.70 & 88.15 & 4.21 & 20.31 &94.38 & 95.94 & 1.95 \\
Howe~\cite{Howe2013DocumentBW} & 20.43 & 94.04 &95.06  & 2.10 & 21.08 & 93.59 & 94.81 &3.18 & 19.01 & 90.79 & 92.28 &1.08 \\
DeepOtsu~\cite{HE2019379} & 20.93 & 91.46 &93.76  &2.60 & 20.34 & 90.09 & 92.96 &3.15 & 21.53 & 95.90 & 97.62 &1.21 \\
Tensmeyer\textit{ et al.}~\cite{tensmeyer2017document} & 20.60 & 92.53 & 96.67 & 2.48 & 20.71 & 93.17 & 96.81 & 2.21 & 20.76 & 91.96 & 94.78 & 2.72 \\
Calvo-Zaragoza~\textit{ et al.}~\cite{calvo2019selectional} & 22.07 & 95.03 & 96.04 & 1.76 &22.57 & 95.25 &96.56 & 1.72 & 21.26 & 95.81 & 96.78 & 1.00 \\
Winner \cite{5277767} & 21.80 & 89.47 &90.18 &3.44 & 20.68 & 92.12 & 94.19 &3.10 & \textcolor{brown}{22.66} & \textcolor{brown}{96.88} & 97.65 & \textcolor{brown}{0.90} \\
cGANs~\cite{10.1016/j.patcog.2019.106968} & 21.91 & 94.96 &96.15  & 1.55 & 22.23 & 95.28 & 96.47 &1.39 & 22.12 & 96.41 & 97.55 &1.07 \\
DD-GAN~\cite{De2020DocumentIB}  &\textcolor{brown}{22.34} & \textcolor{brown}{95.21} &\textcolor{brown}{96.32}  & \textcolor{brown}{1.52} & 22.43 & {95.10} & {96.52} &1.75 & 22.60 & 93.69 & \textcolor{brown}{97.66} &1.27 \\
DocEnTr~\cite{souibgui2022docentr} & 22.29 & 95.31 & 96.29 &1.60 & \textcolor{red}{23.24} & \textcolor{red}{96.42} & \textcolor{brown}{97.69} & \textcolor{red}{0.14} & 20.81 & 94.37 & 96.15 & 1.63 \\
$\rm {D^{2}BFormer}$~\cite{wang2022novel} & \textcolor{red}{23.35} & \textcolor{red}{96.28} & \textcolor{red}{96.90} &\textcolor{red}{1.26} & \textcolor{brown}{22.80} & \textcolor{brown}{95.96} &\textcolor{red}{96.96} &\textcolor{brown}{1.30} & \textcolor{blue}{23.91} & \textcolor{blue}{97.70} & \textcolor{blue}{98.44} &\textcolor{red}{0.64} \\
\midrule
\textbf{DocBinFormer} & \textcolor{blue}{23.96} &\textcolor{blue}{96.80} & \textcolor{blue}{98.27} & \textcolor{blue}{0.05} & \textcolor{blue}{24.10} & \textcolor{blue}{97.13} & \textcolor{blue}{98.34} & \textcolor{blue}{0.07} & \textcolor{red}{23.48} &\textcolor{red}{97.47} & \textcolor{red}{98.43} & \textcolor{blue}{0.12} \\
\bottomrule
\end{tabular}
\label{Tab:ComparativeResults_2012-2014}
\end{table*}

\begin{table*}[ht!]
\centering
\caption{Comparative results of the proposed method and other state-of-the-art methods on the DIBCO 2016-2018 datasets.}
\footnotesize
\begin{tabular}{|l|c|c|c|r||c|c|c|r||c|c|c|r|}
\toprule
\textbf{Method} & \multicolumn{4}{c||}{\textbf{DIBCO 2016}} & \multicolumn{4}{c||}{\textbf{DIBCO 2017}} & \multicolumn{4}{c|}{\textbf{DIBCO 2018}} \\
\midrule
 & \textbf{PSNR$\uparrow$} & \textbf{FM$\uparrow$} & \textbf{Fps$\uparrow$} & \textbf{DRD$\downarrow$} & \textbf{PSNR$\uparrow$} & \textbf{FM$\uparrow$} & \textbf{Fps$\uparrow$} & \textbf{DRD$\downarrow$} & \textbf{PSNR$\uparrow$} & \textbf{FM$\uparrow$} & \textbf{Fps$\uparrow$} & \textbf{DRD$\downarrow$} \\
\midrule
Otsu \cite{otsu1979threshold} & 17.82 & 86.74 & 88.86 & 5.44 & 13.85 & 77.68 & 77.94 & 15.75 & 9.81 & 51.67 & 53.33 & 58.34\\
Sauvola \cite{sauvola2000adaptive} & 17.15 & 83.68 & 87.48 & 6.21 & 14.96 & 79.83 & 87.09 &7.09 & 13.72 & 64.82 & 70.25 & 16.01 \\
Gib~\cite{8517161} & 19.18 & \textcolor{brown}{91.15} &93.03 &\textcolor{brown}{3.20} & 15.78 & 84.89 &88.47 &6.61 &15.12 & 76.63 & 81.13 &11.72  \\
RDD~\cite{6373726} & 17.64 & 84.75 &88.94 & 5.64 & 16.69 & 87.31 &89.77 & 5.80 & 16.45 & 81.48 & 85.02 & 8.32  \\
cGANs~\cite{10.1016/j.patcog.2019.106968} & \textcolor{blue}{19.64} & 91.66 & 94.58 &2.82 & 17.83 & 90.73 &92.58  & 3.58 & 18.37 & 90.60 & 95.39 &4.58 \\
DeepOtsu~\cite{HE2019379} & 19.02 & 90.57 &\textcolor{brown}{94.00} &\textcolor{brown}{3.20} &18.04 & 90.35 &93.67  &3.31 &18.19 & 88.06 & 91.13 &4.63  \\
Tensmeyer\textit{ et al.}~\cite{tensmeyer2017document} & 18.67 & 89.52 & 93.76 & 3.76 & 15.95 & 84.52 & 87.25 & 6.59 & 12.96 & 66.33 & 68.57 & 23.98 \\
Calvo-Zaragoza~\textit{ et al.}~\cite{calvo2019selectional} & 18.79 & 90.72 & 92.62 & 3.28 & 18.44 & 91.67 & 93.04 & 3.24 & 18.52 & 88.17 & 91.11 & 4.69 \\
Winner \cite{5277767} & 18.29 & 88.47 & 91.71 &3.93 & 18.28 & 91.04 & 92.86 & 3.40 & 19.11 & 88.34 & 90.24 & 4.92 \\
DD-GAN~\cite{De2020DocumentIB}  & 18.89 & 89.70 & 93.87 &3.57 & 18.34 & 90.98 &92.65  & 3.44  &17.32  & 85.03 &90.55 &6.05 \\
Howe~\cite{Howe2013DocumentBW} & 18.05 & 87.47 &92.28 &5.35 & 18.52 & 90.10 &90.95  &5.12 &16.67 & 80.84 & 82.85 &11.96 \\
DocEnTr \cite{souibgui2022docentr} & - & - & - & - & \textcolor{brown}{19.11} & \textcolor{brown}{92.53} & \textcolor{red}{95.15} &\textcolor{brown}{2.37} & \textcolor{blue}{19.47} & \textcolor{red}{92.53} & \textcolor{red}{95.15} & \textcolor{red}{2.37} \\
DocDiff~\cite{Yang2023DocDiffDE} & - & - & - & - &- & - & - & - & 17.92 & 88.11 & 90.43 & -  \\
$\rm {D^{2}BFormer}$~\cite{wang2022novel} &\textcolor{brown}{19.24} &\textcolor{red}{90.96} &\textcolor{red}{93.30} &\textcolor{red}{3.22} & \textcolor{red}{19.35} & \textcolor{red}{93.52} & \textcolor{brown}{95.09} & \textcolor{red}{2.12} & \textcolor{brown}{18.91} & \textcolor{brown}{88.84} & \textcolor{brown}{93.42} & \textcolor{brown}{3.99}  \\
\midrule
\textbf{DocBinFormer} & \textcolor{red}{19.26} & \textcolor{blue}{91.99} & \textcolor{blue}{95.74} & \textcolor{blue}{0.10} & \textcolor{blue}{20.93} & \textcolor{blue}{95.55} & \textcolor{blue}{97.21} & \textcolor{blue}{0.07} & \textcolor{red}{19.02} & \textcolor{blue}{93.88} & \textcolor{blue}{96.13} & \textcolor{blue}{0.13} \\
\bottomrule
\end{tabular}
\label{Tab:ComparativeResults_2016-2018}
\end{table*}

\subsection{Quantitative Analysis and Comparison}
\label{sub_sec:QuantAnalysis}
After selecting the best hyperparameters and training the model, we run experiments on a variety of datasets from both the DIBCO and H-DIBCO and evaluate the results with those obtained using other state-of-the-art techniques. Tables~\ref{Tab:ComparativeResults_2009-2011}-\ref{Tab:ComparativeResults_2016-2018} and Table~\ref{Tab:Comp_2019} show the binarization performance of all compared methods in terms of FM, Fps, DRD, and PSNR on the ten degraded document datasets along with the performance of the proposed methodology at the end of each table. The results of the best model are highlighted in \textcolor{blue}{Blue} and the second-best and third-best are highlighted in \textcolor{red}{Red} and \textcolor{brown}{Brown} respectively in each table. From Tables~\ref{Tab:ComparativeResults_2009-2011}-\ref{Tab:ComparativeResults_2016-2018} and Table~\ref{Tab:Comp_2019}, we can see that the DocBinFormer obtains a strong competitive performance on the ten benchmark datasets. The proposed model beats all the other methods in DIBCO 2009, 2010, 2011, 2012, 2016, and 2017. It is very competitive for DIBCO 2014, 2018 and 2019. These improvements can be quantified as - (1) For DIBCO 2009, we notice that the performance of the model outperforms the previous best model-based method by 0.85 and 3.98\% absolute improvement and percentage improvement, respectively, on PSNR. (2) We also calculated the percentage improvement from the last best method in terms of PSNR for other DIBCO sets. The results of which are - 1.55\% for 2010, 4.09\% for 2011, 2.61\% for 2012, 3.70\% for 2013, and 8.16\% for 2017. Our method is second best for the DIBCO 2014, 2016, and 2018 and third best in DIBCO 2019 in terms of PSNR. Our main conclusion is divided into these sections where we compare the results to different classes of methods: (1) \textbf{Comparison with traditional methods} - When compared to the methods such as Otsu or Sauvola, we conjecture that the main reason is the tailor-designed prior on degraded documents in the traditional models, resulting in a limited ability to handle multiple degradations and thus being less robust. However, the DocBinFormer model possesses an end-to-end learning framework, resulting in a strong ability to capture abundant document features and achieve accurate binarization results to be a more robust solution. (2) \textbf{Comparison with GAN-based methods} - When compared to the architectures such as cGANs, that utilize a GAN-based method, DocBinFormer yields 9.26\% on 2009, 9.37\% on 2010, 9.27\% on 2011, 9.35\% on 2012, 8.41\% on 2013,6.41\% on 2014,  17.38\% in 2017, 3.53\% in 2018, and 14.49\% in terms of PSNR. These experimental results demonstrate that the DocBinFormer is able to capture the intrinsic distribution of the source datasets quite effectively, which the GAN-based, model could not. We think the main reason can be attributed to the two-level attention that helped in capturing the intricate details of the image. (3) \textbf{Within the DIBCO 2016 dataset}, it is observed that the cGANs model achieved the highest PSNR performance. This success can potentially be attributed to the inherent flexibility of adversarial training techniques and the integration of multi-scale information. These attributes empower cGANs to augment the training dataset in an unsupervised fashion, thereby enhancing binarization performance. However, it is worth noting that, as highlighted earlier, when considering the percentage improvement achieved by DocBinFormer across all the remaining datasets, it becomes evident that this model possesses superior generalization capabilities compared to GAN-based models in a broader context.
\begin{table}[htbp]
\caption{\centering Comparative results of the proposed method and other state-of-the-art methods on the DIBCO 2019 dataset.}
\centering
\resizebox{\columnwidth}{!}{
\begin{tabular}{|l|c|c|c|r|}
\toprule
\textbf{Method} & \textbf{PSNR$\uparrow$}        &\textbf{FM$\uparrow$}        &\textbf{Fps$\uparrow$}   &\textbf{DRD$\downarrow$}   
\\ \midrule
Otsu \cite{otsu1979threshold} & 12.67 & 63.87 & 60.24 & 16.87 \\
Sauvola \cite{sauvola2000adaptive} & 12.66 & 63.82 & 60.18 & 16.91 \\
Gib~\cite{8517161} & 14.16 & 71.67 &70.83 &16.68 \\
RDD~\cite{6373726}  & 12.03 & 61.98 & 58.81 & 30.13 \\
cGANs~\cite{10.1016/j.patcog.2019.106968} & 12.80 & 60.77 & 60.69 &17.94 \\
DeepOtsu~\cite{HE2019379} &14.44 & 60.75 & 59.91 &13.46 \\
Tensmeyer\textit{ et al.}~\cite{tensmeyer2017document} & 12.38 & 59.14 & 57.72 & 23.96 \\
Calvo-Zaragoza~\textit{ et al.}~\cite{calvo2019selectional} & 12.35 & 58.82 & 56.00 & 19.65 \\
Winner \cite{5277767} & 14.48 & \textcolor{red}{72.87} &\textcolor{red}{72.15} & 16.24 \\
DD-GAN~\cite{De2020DocumentIB}  &14.43 &57.96 & 57.30 &12.21\\
Howe~\cite{Howe2013DocumentBW} & 12.11 & 31.58 & 30.34 &26.14 \\
DocEnTr \cite{souibgui2022docentr} & 13.85 & 59.00 & 60.00 & \textcolor{red}{0.30} \\
DocDiff~\cite{Yang2023DocDiffDE} &\textcolor{blue}{15.14} & \textcolor{blue}{73.38} & \textcolor{blue}{75.12} & - \\
$\rm {D^{2}BFormer}$~\cite{wang2022novel} & \textcolor{red}{15.05} & \textcolor{brown}{67.63} & \textcolor{brown}{66.69} & \textcolor{brown}{10.59} \\
\midrule
\textbf{DocBinFormer} & \textcolor{brown}{14.49} &60.31 & 64.00 & \textcolor{blue}{0.21}
\\ \hline 
\end{tabular}}
\label{Tab:Comp_2019}
\end{table}

(4) \textbf{Comparison with CNN-based methods} - When compared to the architectures that utilize CNNs such as DeepOtsu, Tensmeyer \textit{et al.}~\cite{tensmeyer2017document} or Calvo-Zaragoza \textit{et al.}~\cite{calvo2019selectional}, and the results presented in the paper shows that DocBinFormer proves more effective when compared to CNN based models. We conjecture that the main reason is that the CNN-based models tend to focus on local-aware semantic features, our two-level ViT architecture effectively captures both global and local characteristics in document images. This comprehensive feature extraction capability contributes to the superior and more precise binarization results achieved by our method compared to conventional CNN-based approaches.(5) \textbf{Comparison with ViT-based methods} - In our quantitative comparison with ViT-based approaches, including DocEnTr, it becomes evident that our proposed TL-ViT architecture outperforms the traditional ViT model in all the datasets except for DIBCO 2018. The rationale behind this notable performance difference lies in the fundamental distinction in feature extraction capabilities. While the vanilla ViT predominantly concentrates on global feature representation, thereby potentially overlooking crucial local features, the presented architecture excels by effectively capturing both global and local characteristics within document images. This comprehensive approach to feature extraction proves instrumental in achieving superior and more effective results compared to the traditional ViT model, which primarily focuses on global information. (6) \textbf{Comparison with ViT-CNN hybrid methods} - At last, we also compared our method with CNN-ViT hybrid models. The reason is primarily to demonstrate that though the hypothesis that the CNN-ViT combination can effectively capture global and local features holds true to some extent, because of the limitations of CNNs discussed earlier in this paper this hybrid model might not be the best choice. The experimental results show evidently that our proposed method is much more effective than CNN-ViT fusion models such as $\rm {D^{2}BFormer}$.

In conclusion, we evidently demonstrated that {\textbf{DocBinFormer}} surpasses the prior state-of-the-art algorithms for the document image binarization problem on the majority of the DIBCO and the H-DIBCO datasets.

\begin{figure*}[!ht]
\centering
\captionsetup{justification=centering}
\begin{tabular}{ccc}
\includegraphics[width=0.3\linewidth]{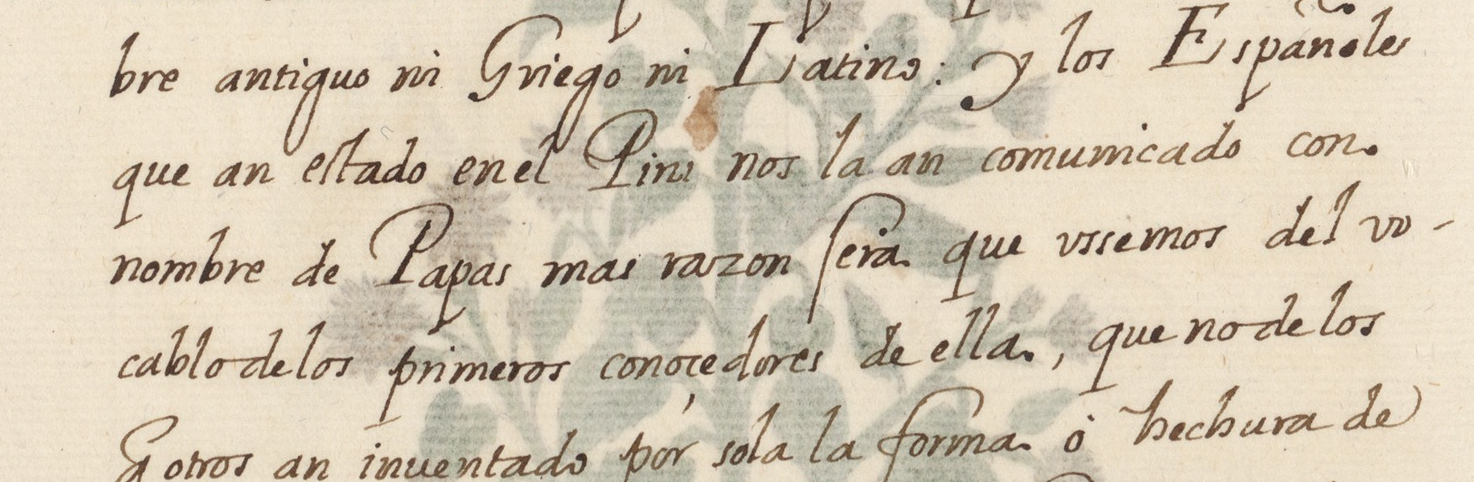} &
\includegraphics[width=0.3\linewidth]{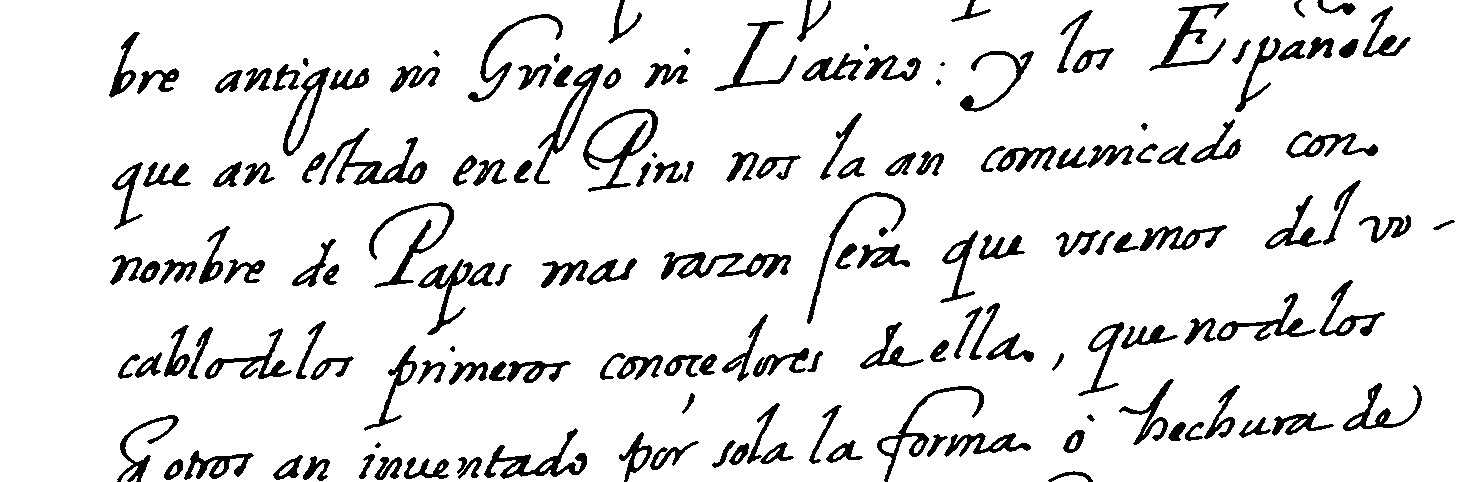} &
\includegraphics[width=0.3\linewidth]{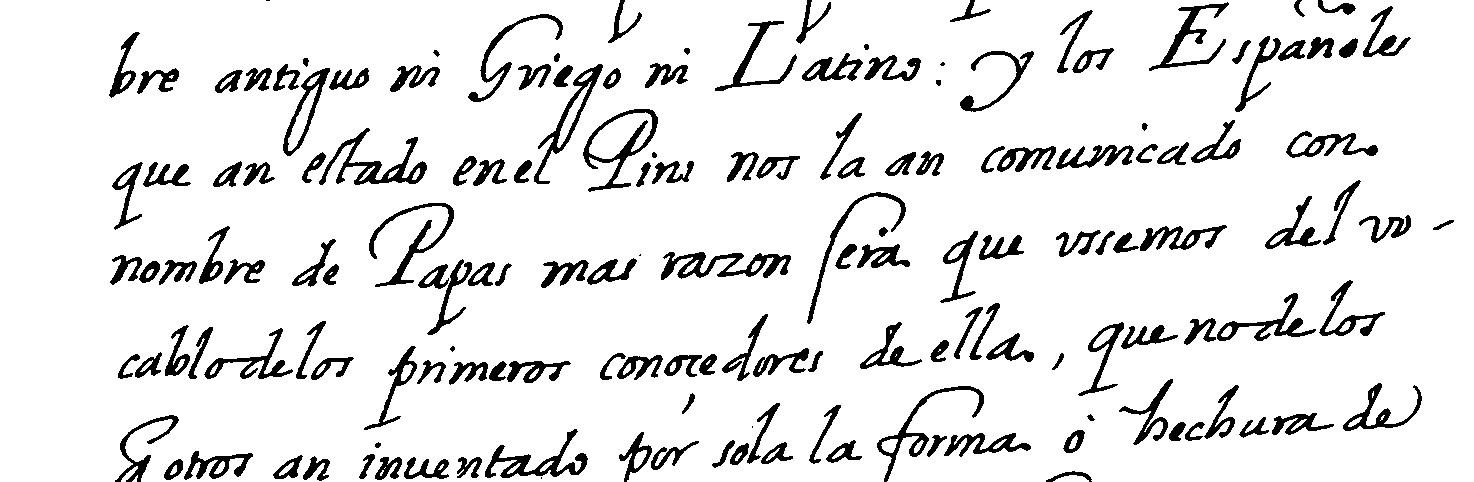} \\ 
& &  (\textcolor{blue}{PSNR = 22.84}) \\
\includegraphics[width=0.3\linewidth]{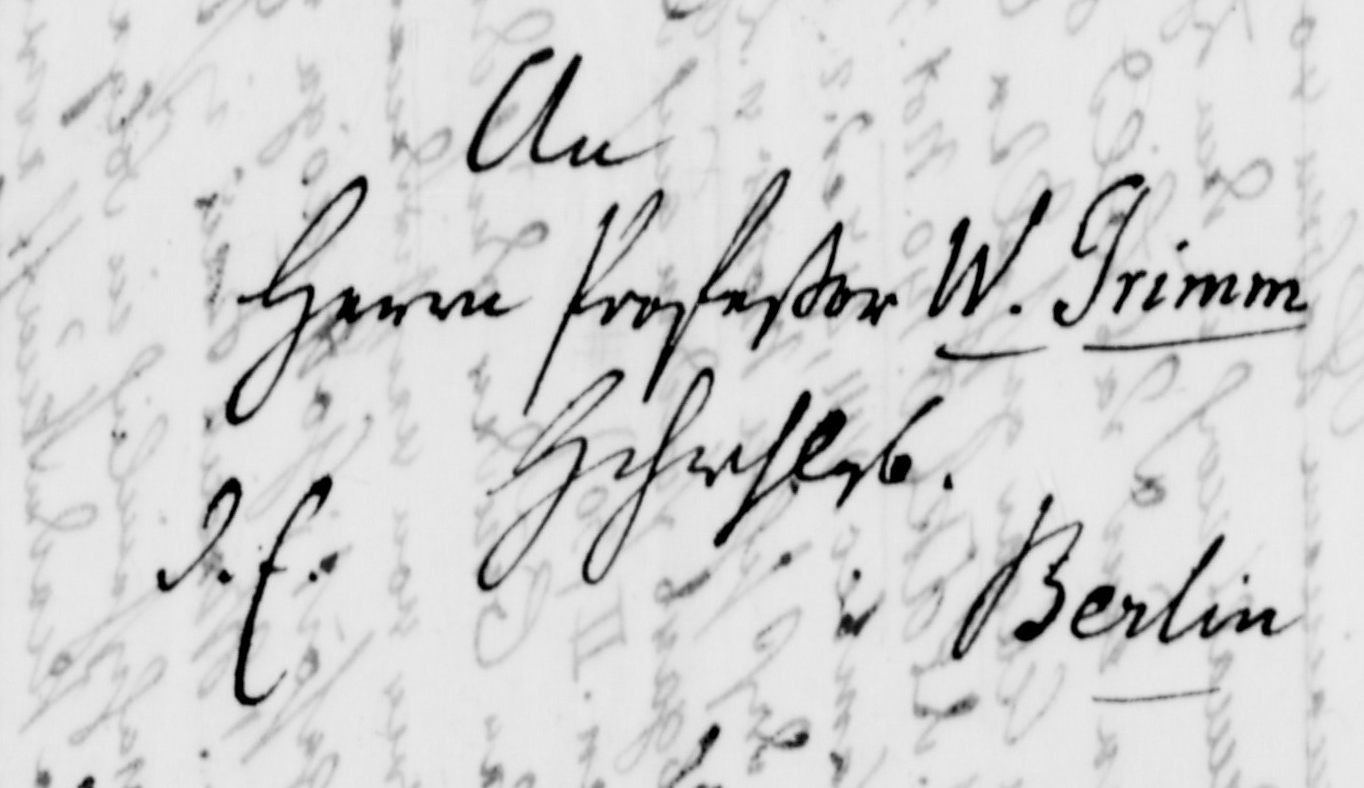} &
\includegraphics[width=0.3\linewidth]{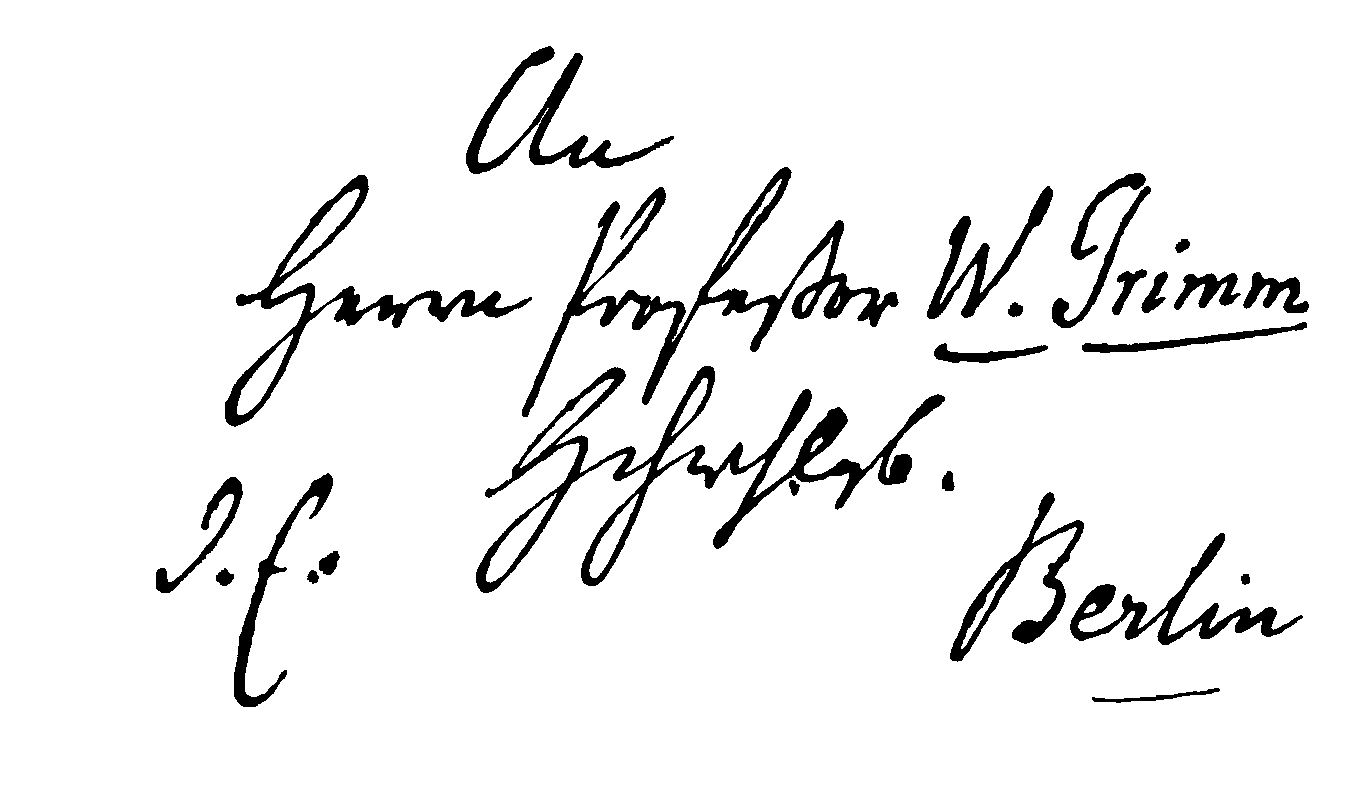} &
\includegraphics[width=0.3\linewidth]{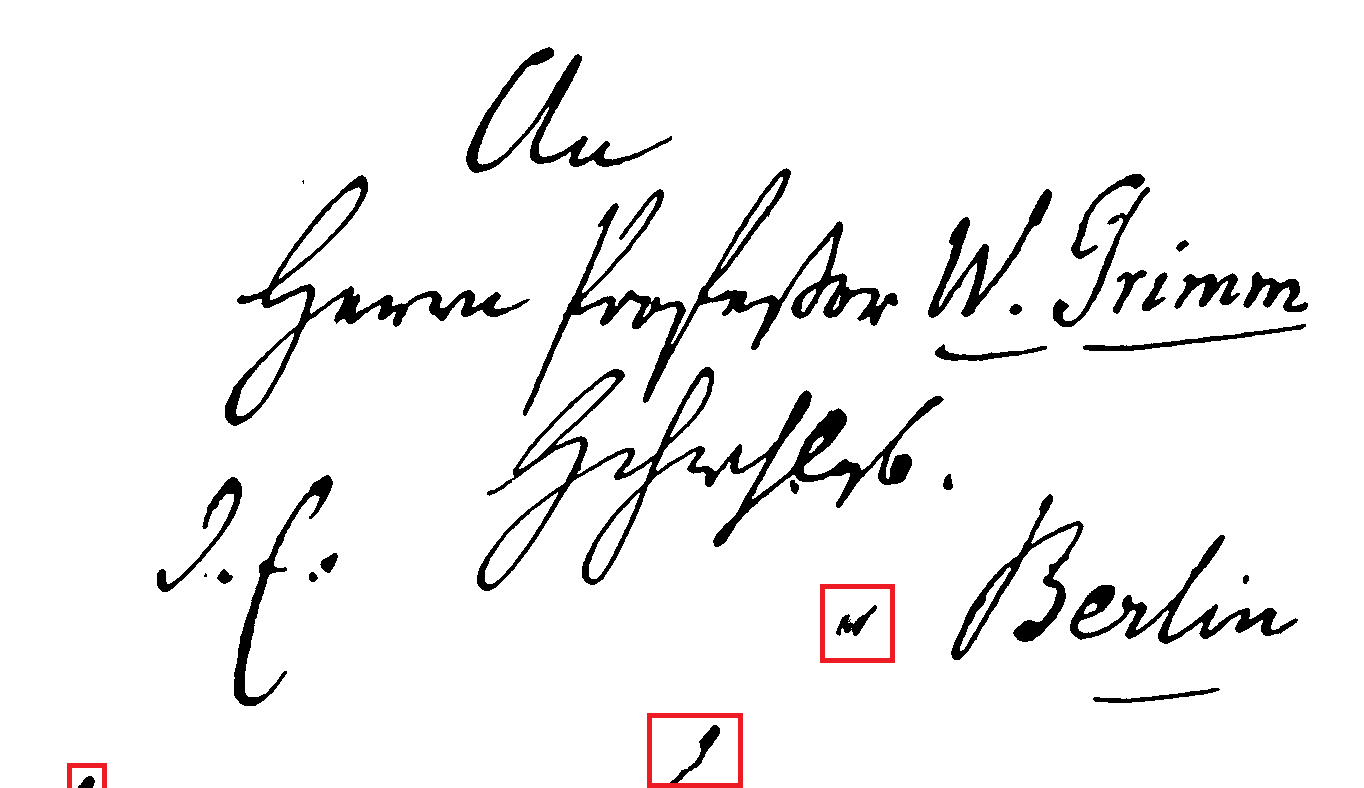} \\
& &  (\textcolor{blue}{PSNR = 19.01}) \\
\end{tabular}
\caption{Qualitative analysis on output generated by \textbf{DocBinFormer} on various \textbf{handwritten} DIBCO and H-DIBCO Images. Images from left to right are: Original, Ground Truth, and Binarized output.}
\label{fig:Output_Handwritten}
\end{figure*}

\begin{figure*}[!ht]
\centering
\captionsetup{justification=centering}
\begin{tabular}{ccc}
\includegraphics[width=0.3\linewidth]{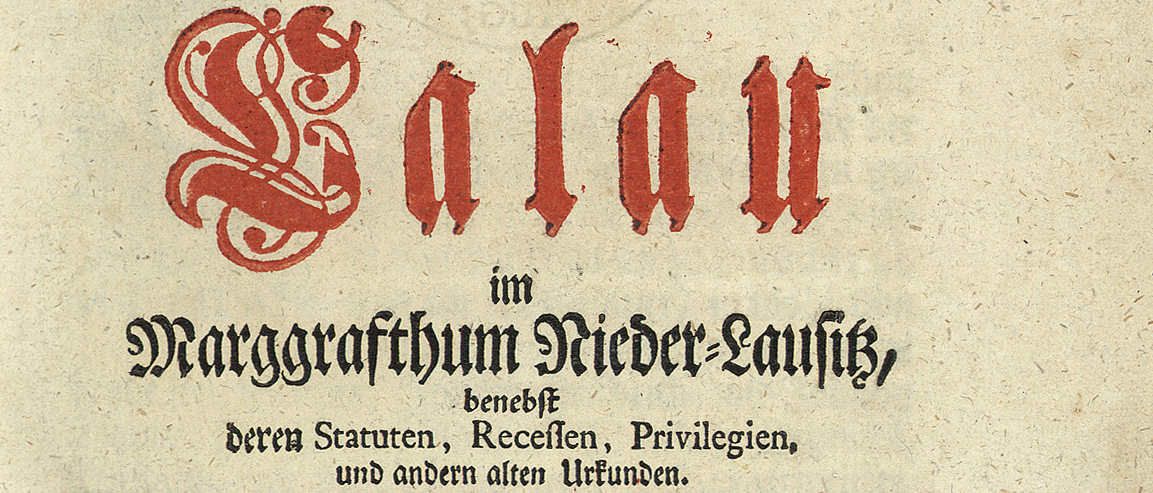} &
\includegraphics[width=0.3\linewidth]{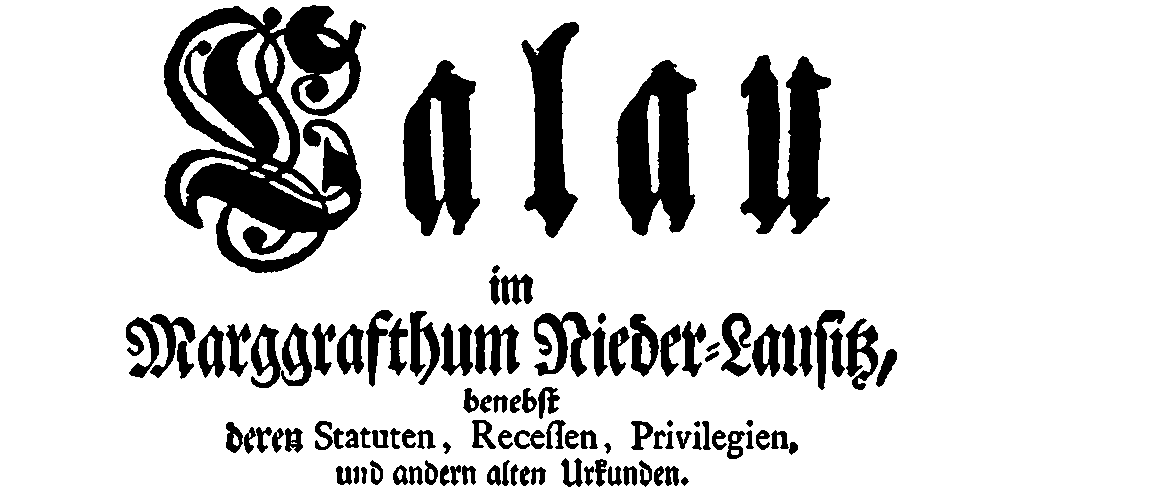} &
\includegraphics[width=0.3\linewidth]{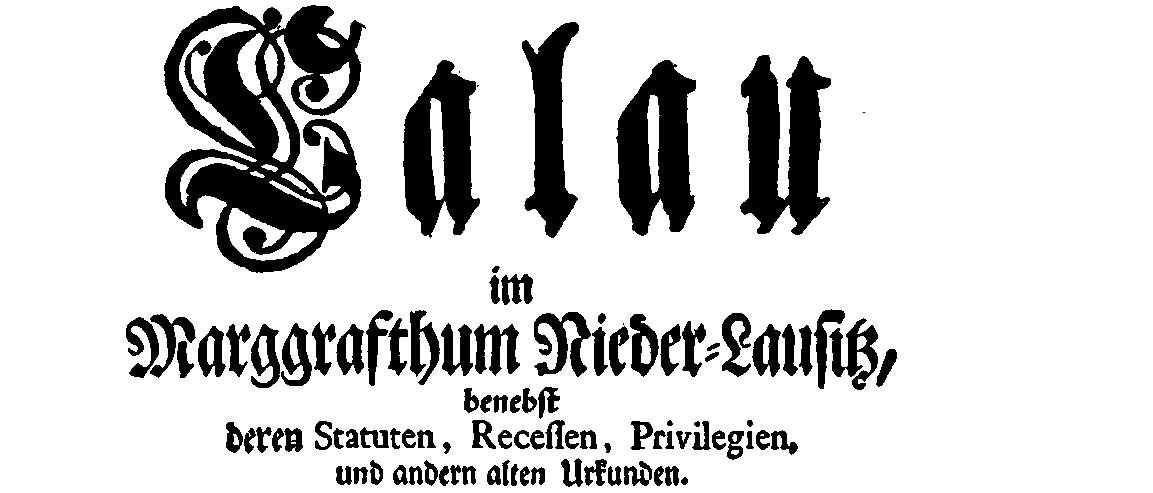} \\ \\
\includegraphics[width=0.3\linewidth]{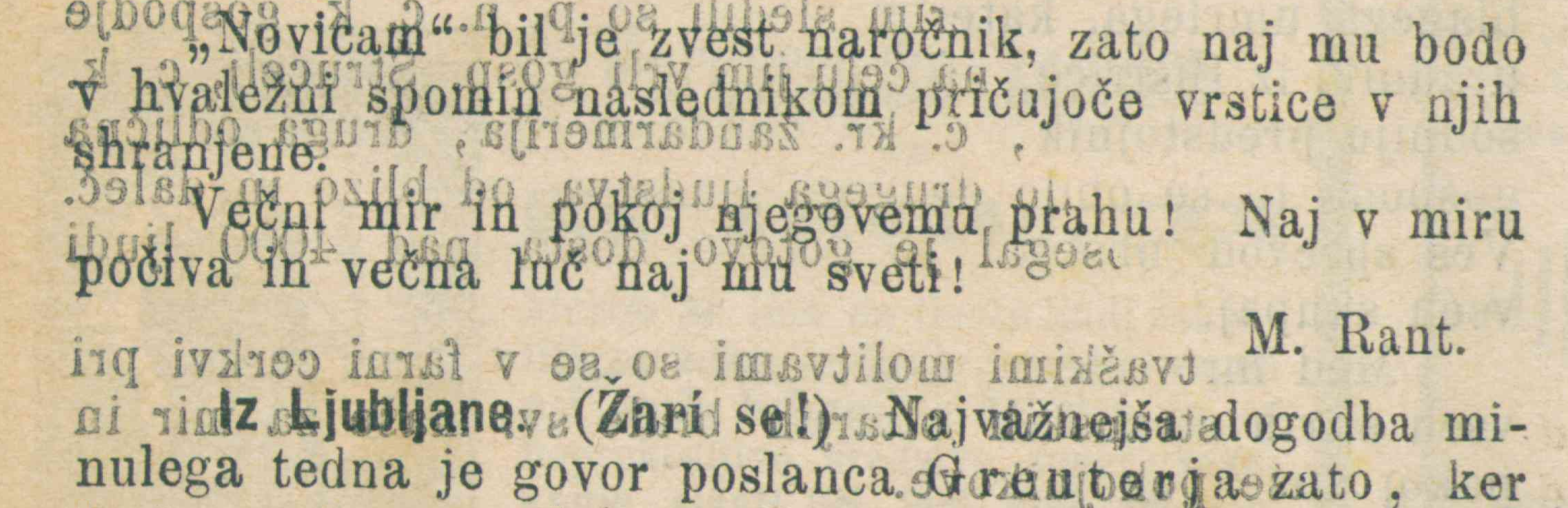} &
\includegraphics[width=0.3\linewidth]{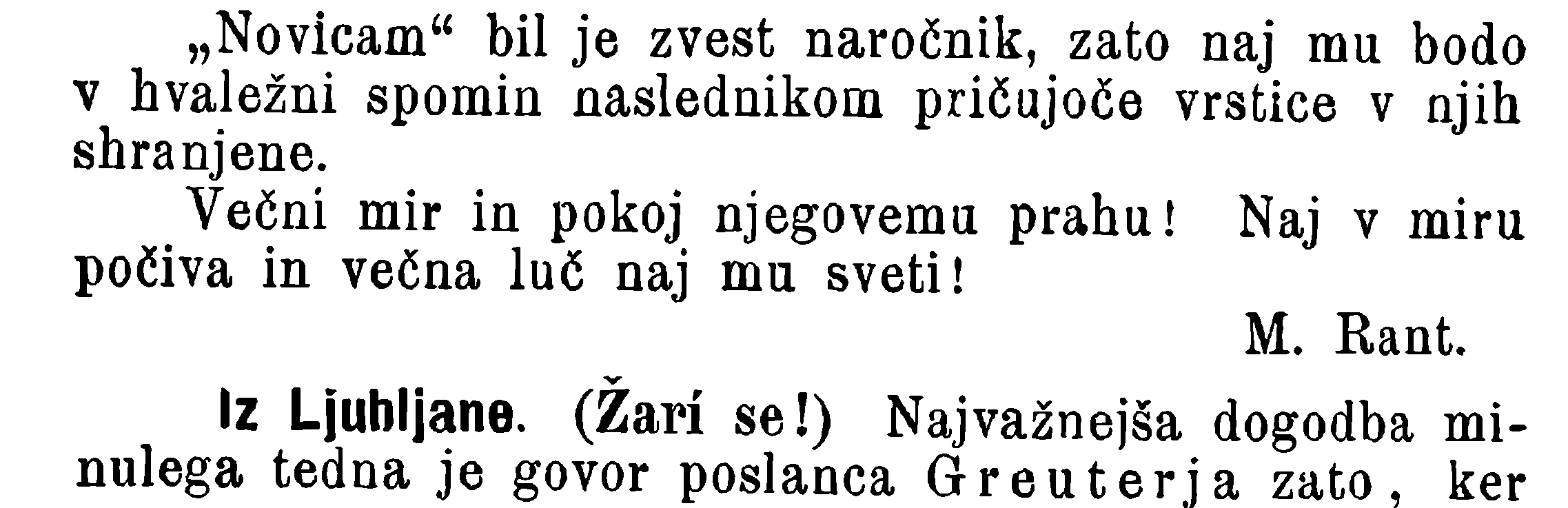} &
\includegraphics[width=0.3\linewidth]{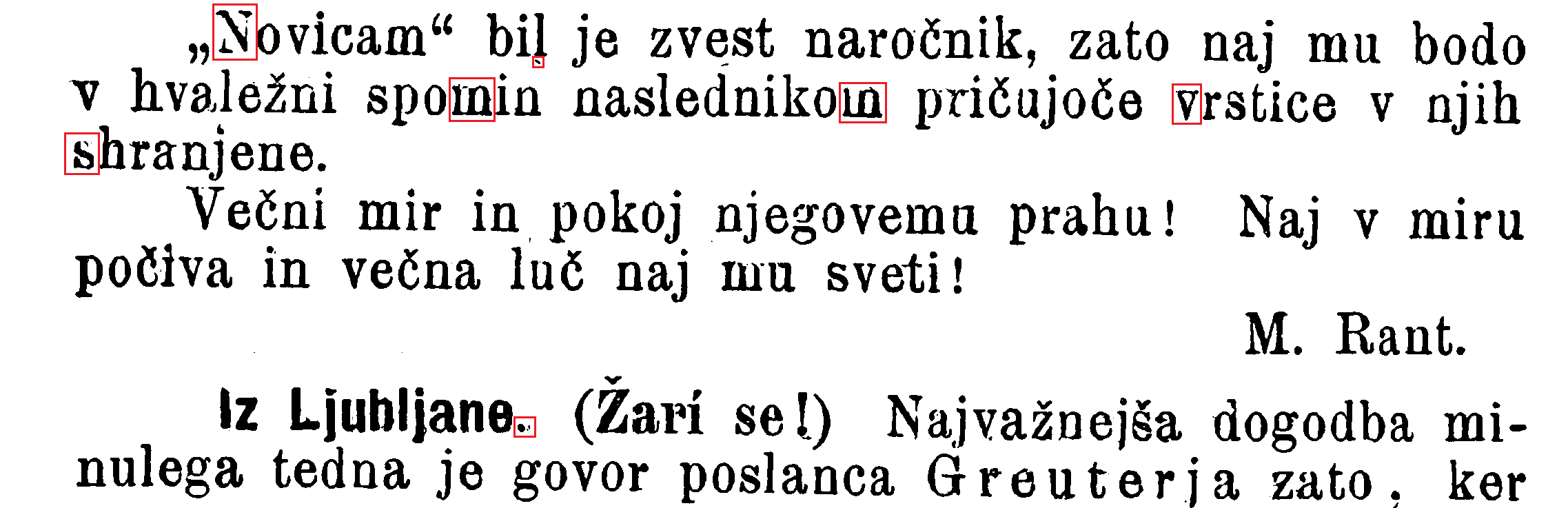} \\
\end{tabular}
\caption{Qualitative Analysis on output generated by \textbf{DocBinFormer} on various \textbf{machine-typed} DIBCO and H-DIBCO Images. Images from left to right are: Original, Ground Truth, and Binarized output.}
\label{fig:Output_Typed}
\end{figure*}

\begin{figure*}[!ht]
\begin{center}
\captionsetup{justification=centering}
\begin{tabular}{ccccc}
\includegraphics[width=0.35\columnwidth]{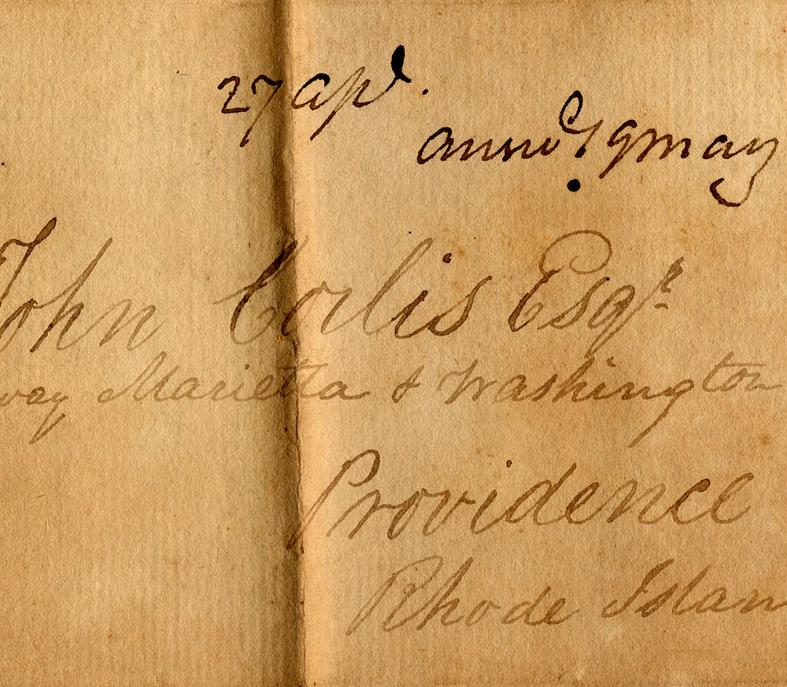} &
\includegraphics[width=0.35\columnwidth]{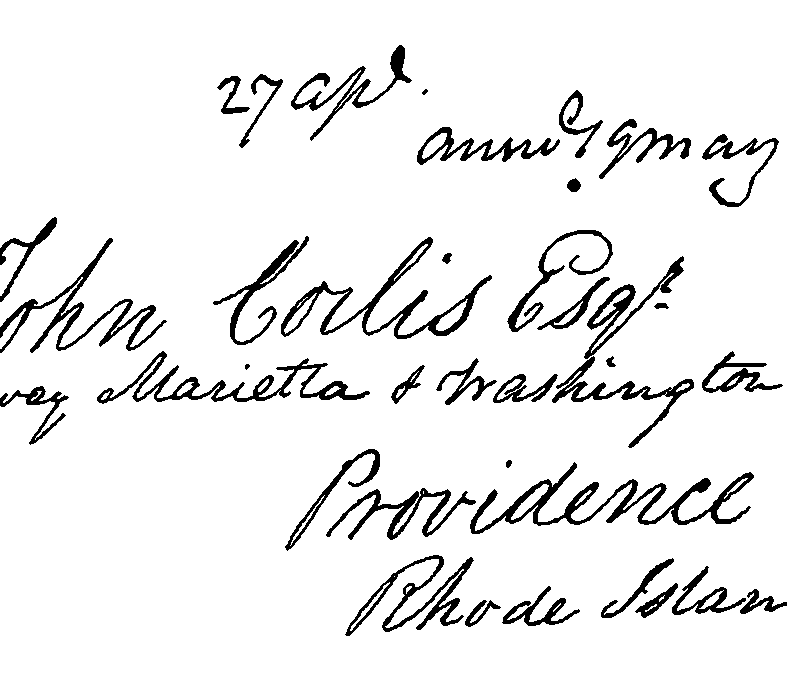} &
\includegraphics[width=0.35\columnwidth]{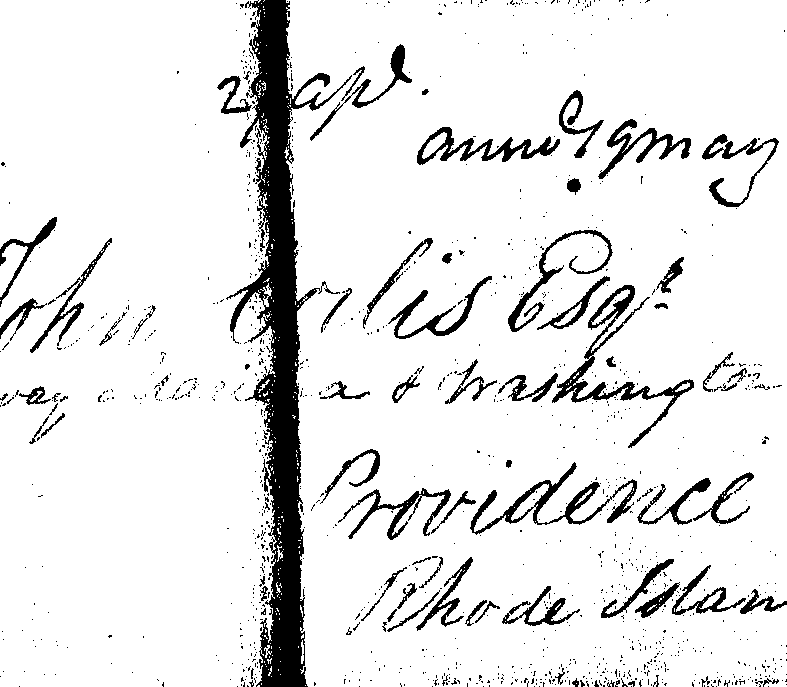} & 
\includegraphics[width=0.35\columnwidth]{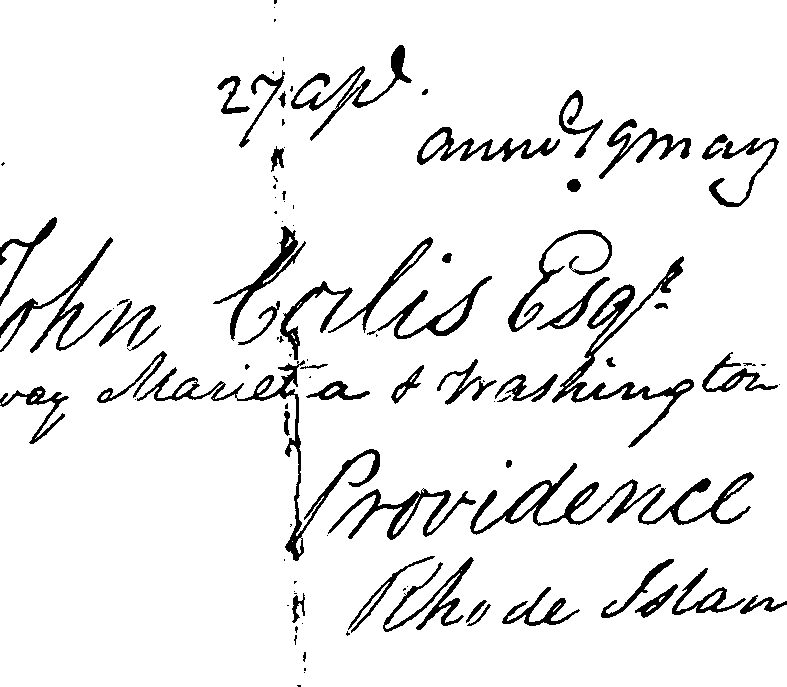} \\
\textit{a}) Original &
\textit{b}) GT &
\textit{c}) Otsu~\cite{otsu1979threshold}   & 
\textit{e}) SauvolaNet~\cite{li2021sauvolanet}  & \\
& & (\textcolor{blue}{PSNR = 12.18}) &  (\textcolor{blue}{PSNR = 19.44}) \\
\includegraphics[width=0.35\columnwidth]{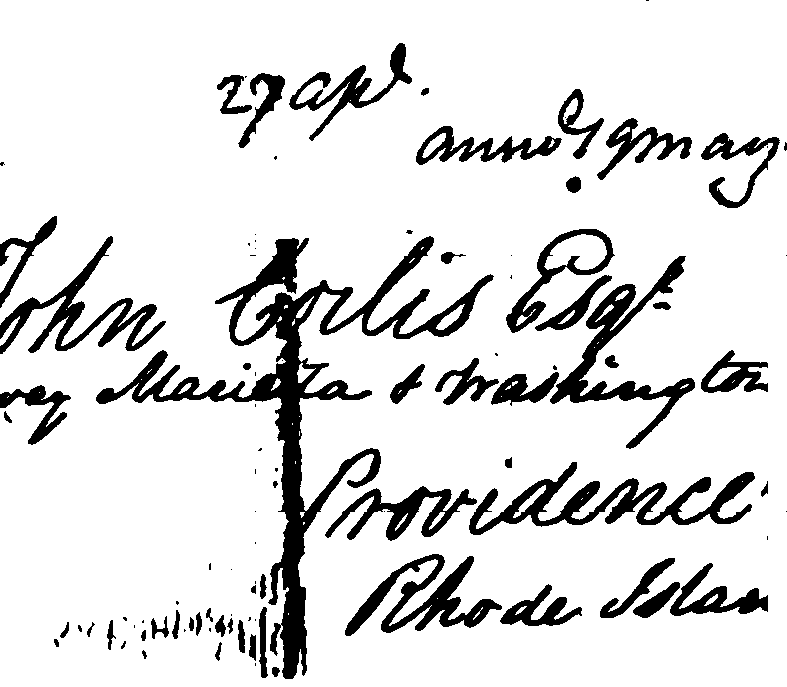} &
\includegraphics[width=0.35\columnwidth]{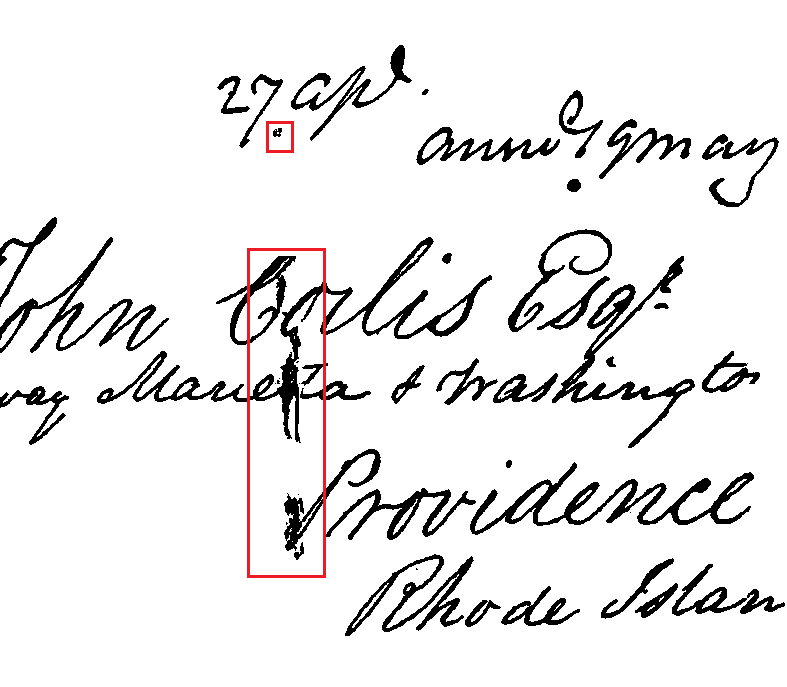} &
\includegraphics[width=0.35\columnwidth]{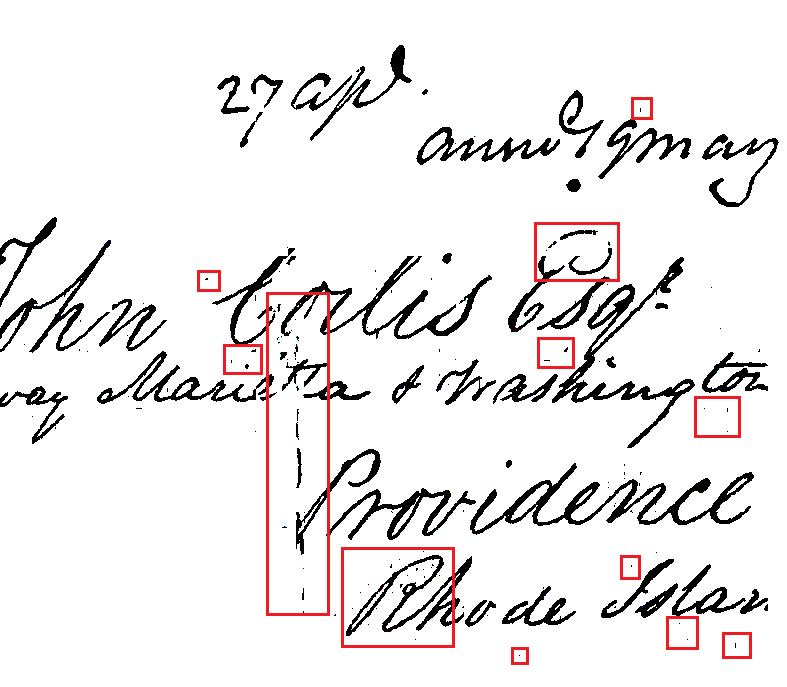} &
\includegraphics[width=0.35\columnwidth]{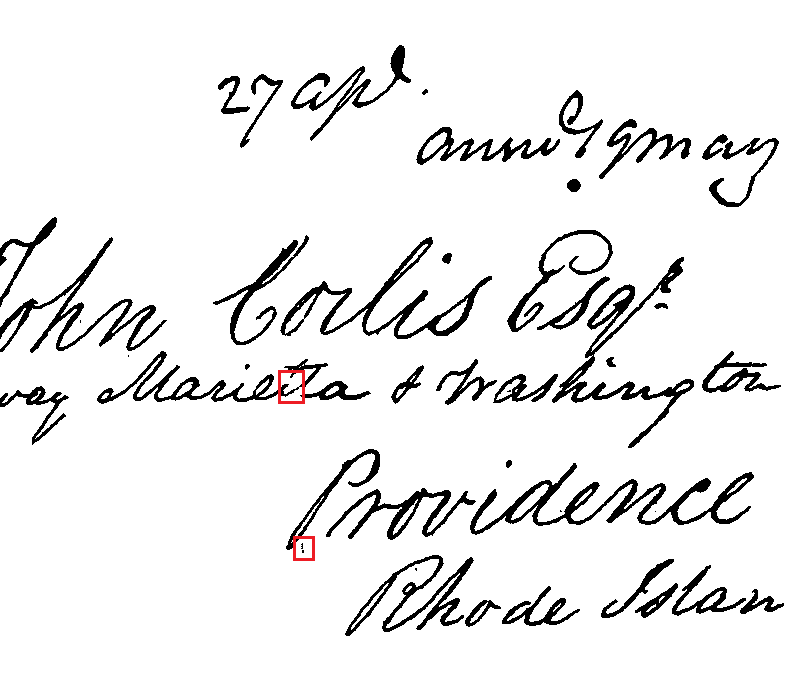} & \\ 
\textit{f}) DE GAN~\cite{souibgui2020gan} & 
\textit{g}) Jemni~\textit{et al.}~\cite{jemni2022enhance}  &
\textit{h}) DocEnTr~\cite{souibgui2022docentr}  &
\textit{i}) \textbf{DocBinFormer}  & \\
(\textcolor{blue}{PSNR = 12.31}) & (\textcolor{blue}{PSNR = 19.52}) & (\textcolor{blue}{PSNR = 19.93}) & (\textcolor{blue}{PSNR = 21.89})\\ 
\end{tabular}
\caption{Qualitative performance of the various binarization techniques on sample no. 6 (shown in (a)) from the DIBCO 2011 dataset.}
\label{fig:compDIBCO2011_2}
\end{center}
\end{figure*}

\subsection{Qualitative Analysis and Comparison}
\begin{figure*}[!ht]
\begin{center}
\captionsetup{justification=centering}
\begin{tabular}{cccc}
\includegraphics[width=0.35\columnwidth]{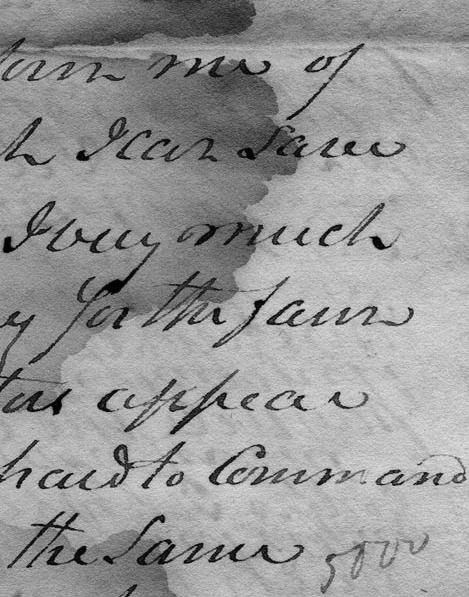} &
\includegraphics[width=0.35\columnwidth]{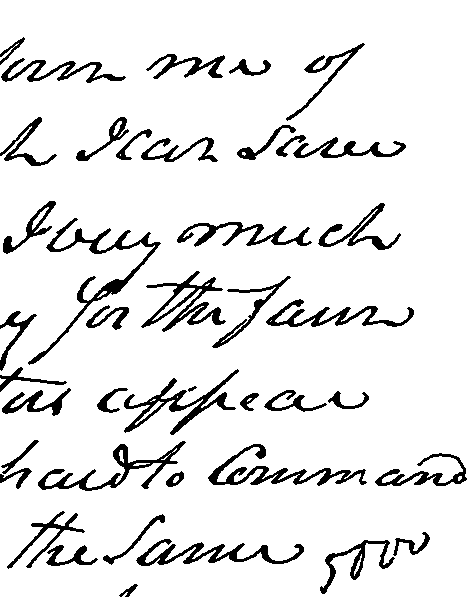} & 
\includegraphics[width=0.35\columnwidth]{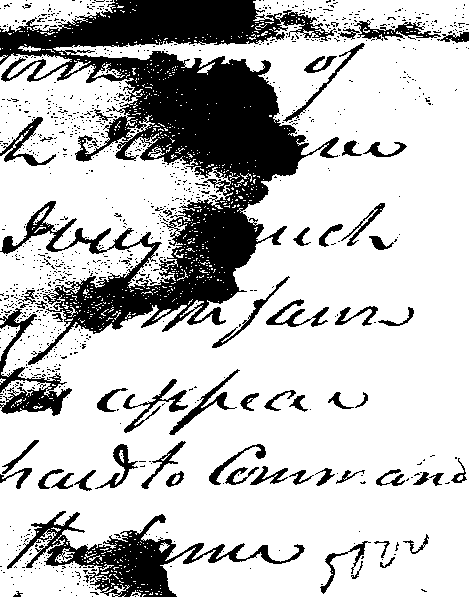} & 
\includegraphics[width=0.35\columnwidth]{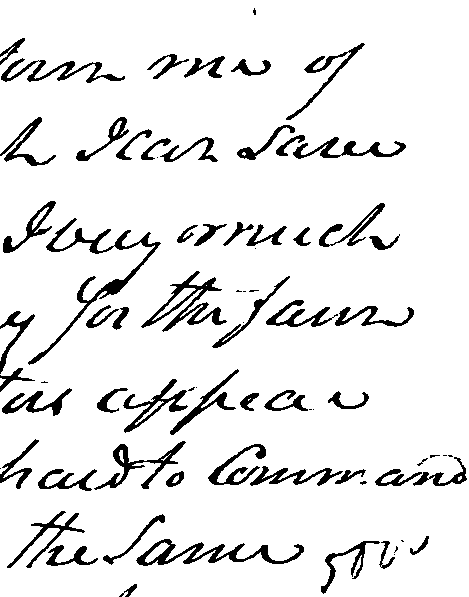} \\
\textit{a}) Original &
\textit{b}) GT & 
\textit{c}) Otsu~\cite{otsu1979threshold}  & 
\textit{d}) SauvolaNet~\cite{li2021sauvolanet} \\
& & (\textcolor{blue}{PSNR = 7.78}) &  (\textcolor{blue}{PSNR = 19.90}) \\
\includegraphics[width=0.35\columnwidth]{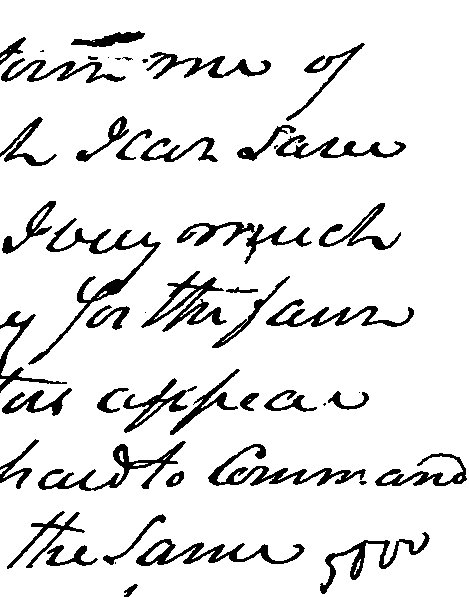} &
\includegraphics[width=0.35\columnwidth]{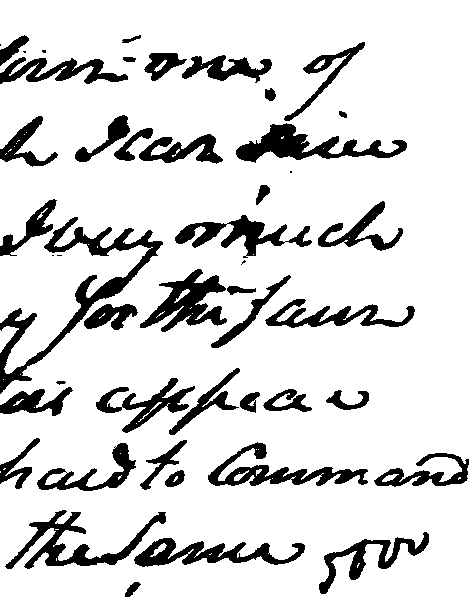} & 
\includegraphics[width=0.35\columnwidth]{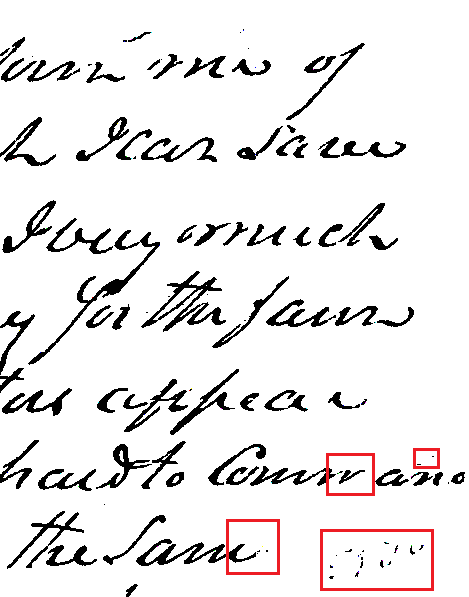} &
\includegraphics[width=0.35\columnwidth]{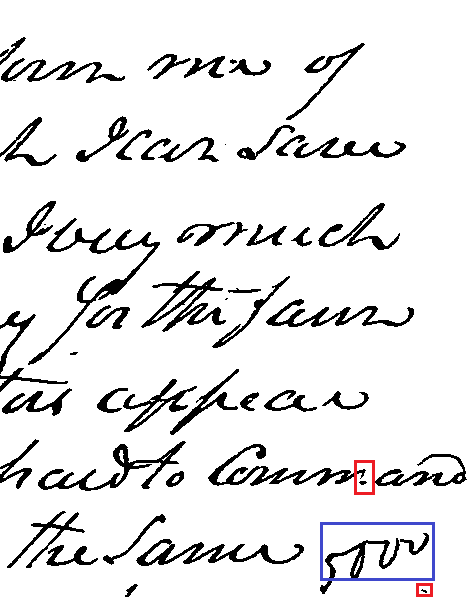}  \\
\textit{e}) Jemni \textit{et al.}~\cite{jemni2022enhance} &
\textit{f}) DE GAN~\cite{souibgui2020gan}    & 
\textit{g}) DocEnTr~\cite{souibgui2022docentr}   & 
\textit{h}) \textbf{DocBinFormer}  \\
(\textcolor{blue}{PSNR = 18.99}) & (\textcolor{blue}{PSNR = 12.22}) & (\textcolor{blue}{PSNR = 17.73}) & (\textcolor{blue}{PSNR = 20.59})\\ 
\end{tabular}
\caption{Qualitative performance of the various binarization techniques on sample no. 4 (shown in (a)) from DIBCO 2011.}
\label{fig:compDIBCO2011_1}
\end{center}
\end{figure*}

Subsection~\ref{sub_sec:QuantAnalysis} provided the quantitative results produced by the proposed model and did a comparative analysis. In this subsection, we will provide a qualitative evaluation. We begin the qualitative analysis by providing an intuitive comparison between the ground truth (GT) image and the output of the proposed model on various handwritten and machine-typed document images taken from different DIBCO datasets. The illustration is shown in Fig.~\ref{fig:Output_Handwritten},~\ref{fig:Output_Typed} and~\ref{fig:compDIBCO2011_1}, respectively. In the qualitative analysis, we have used \textcolor{red}{red bounding boxes} in the output images to highlight the areas where a particular model failed to generate effective binary output. We also highlighted the areas on the output generated by DocBinFormer with \textcolor{blue}{blue bounding boxes}, where the majority of the other models failed. Next, we performed a comparative evaluation of the output produced by \textbf{DocBinFormer} in contrast to the related methodologies on two sample images taken from the DIBCO 2011 and a sample taken from the DIBCO 2012 dataset. The comparative evaluation is shown in Figs.~\ref{fig:compDIBCO2011_1} and \ref{fig:compDIBCO2011_2}, respectively. 

\begin{figure*}[ht!]
\centering
\begin{tabular}{ccccc}
{\includegraphics[width = 0.45\columnwidth]{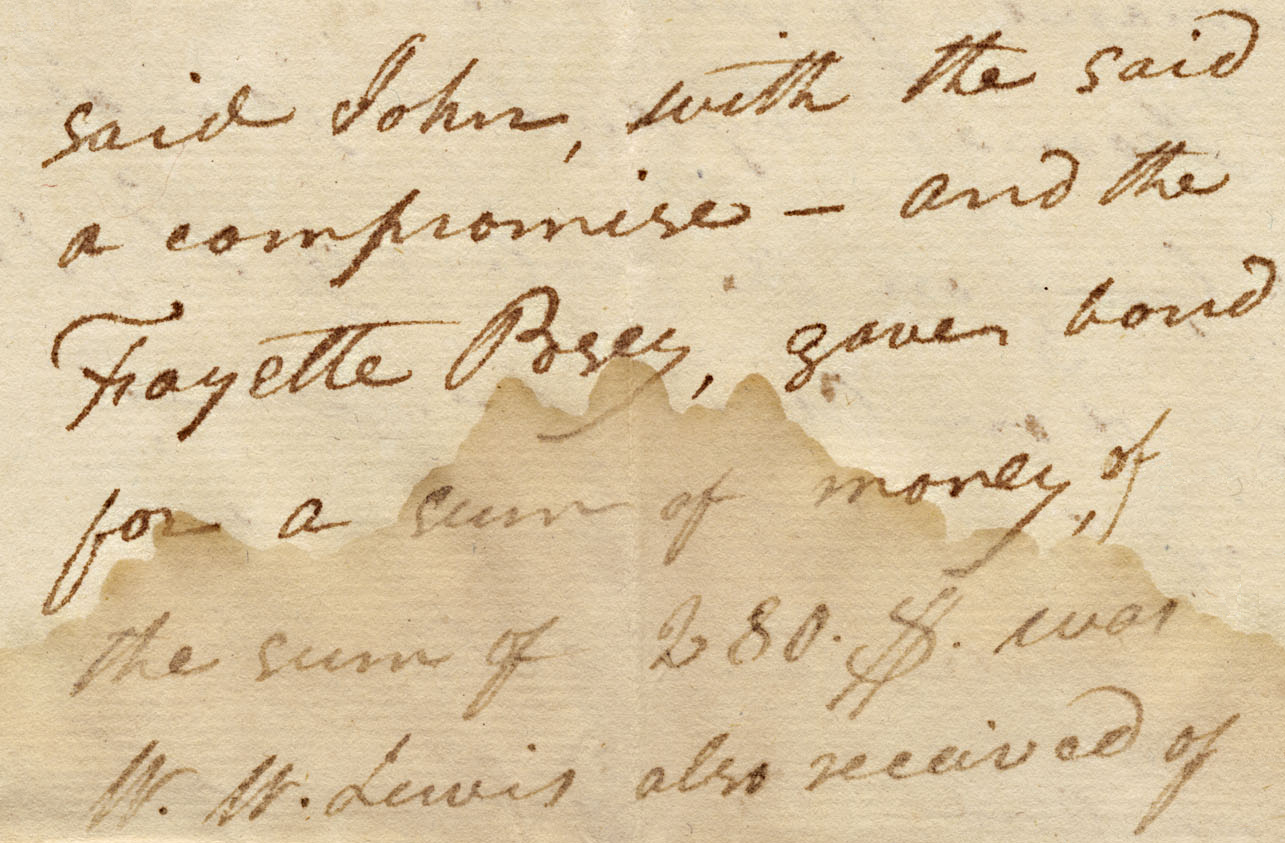}} &
{\includegraphics[width=0.45\columnwidth]{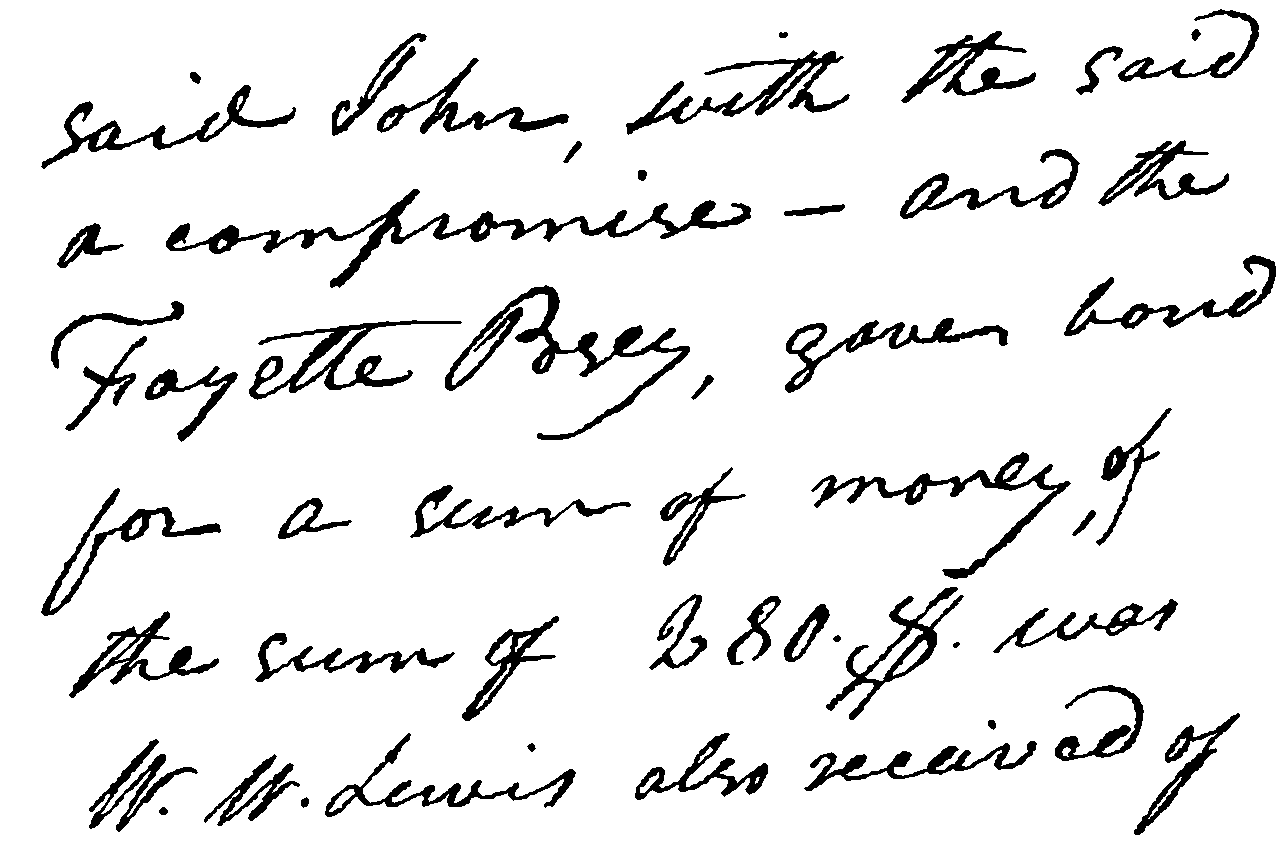}} &
{\includegraphics[width=0.45\columnwidth]{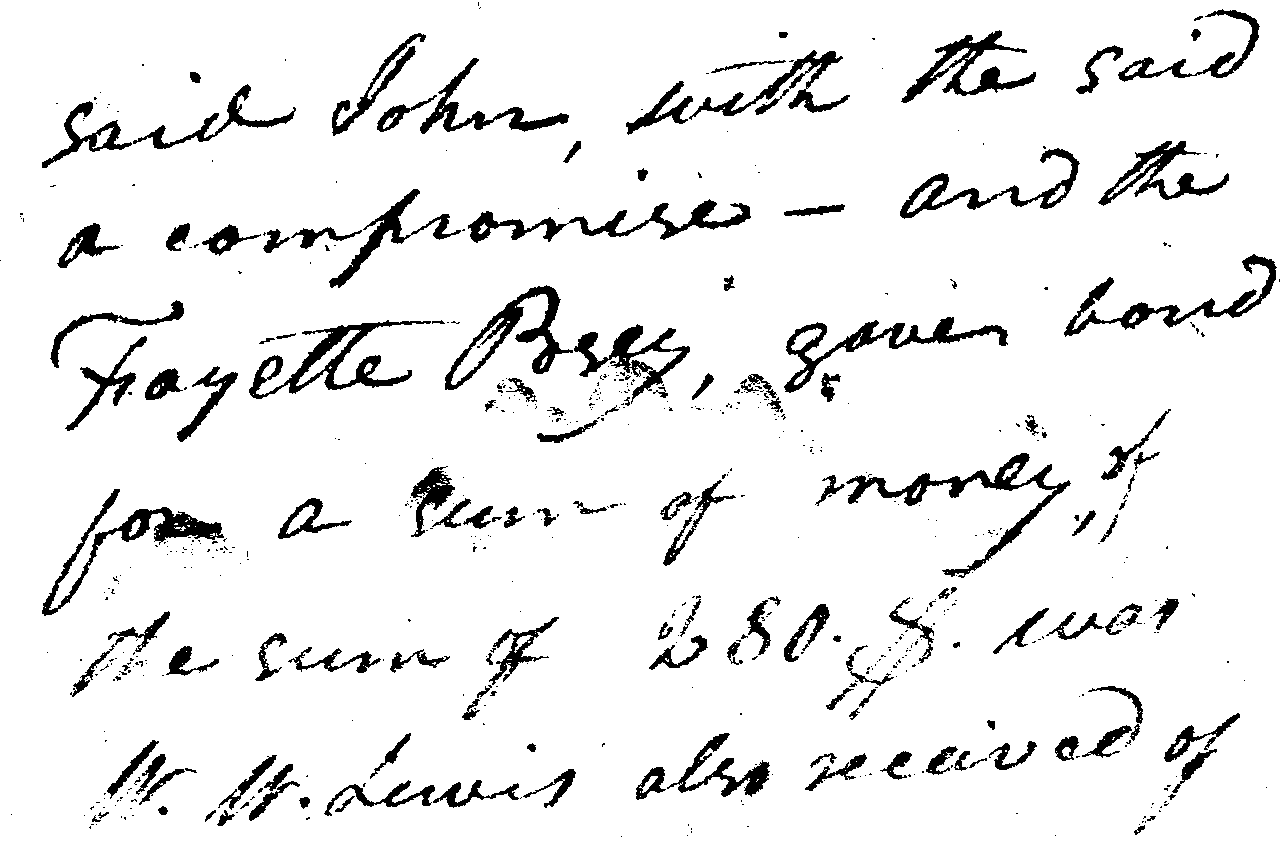}}&
{\includegraphics[width = 0.45\columnwidth]{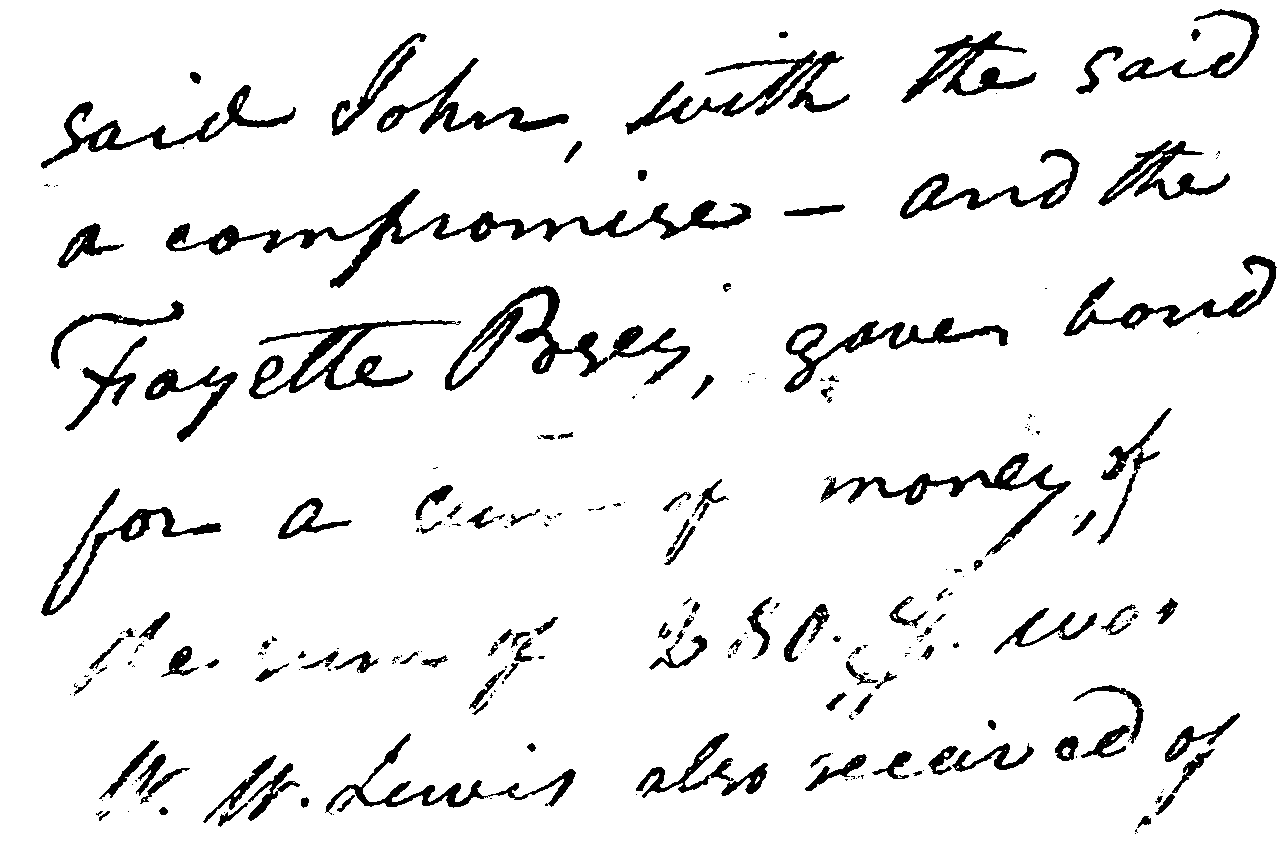}} & \\
\scriptsize a) Original &  \scriptsize b) GT  &  \scriptsize c) Otsu~\cite{otsu1979threshold}  &  \scriptsize d) SauvolaNet~\cite{li2021sauvolanet}\\ 
& & (\textcolor{blue}{PSNR = 16.78 }) &  (\textcolor{blue}{PSNR = 18.14}) \\
{\includegraphics[width=0.45\columnwidth]{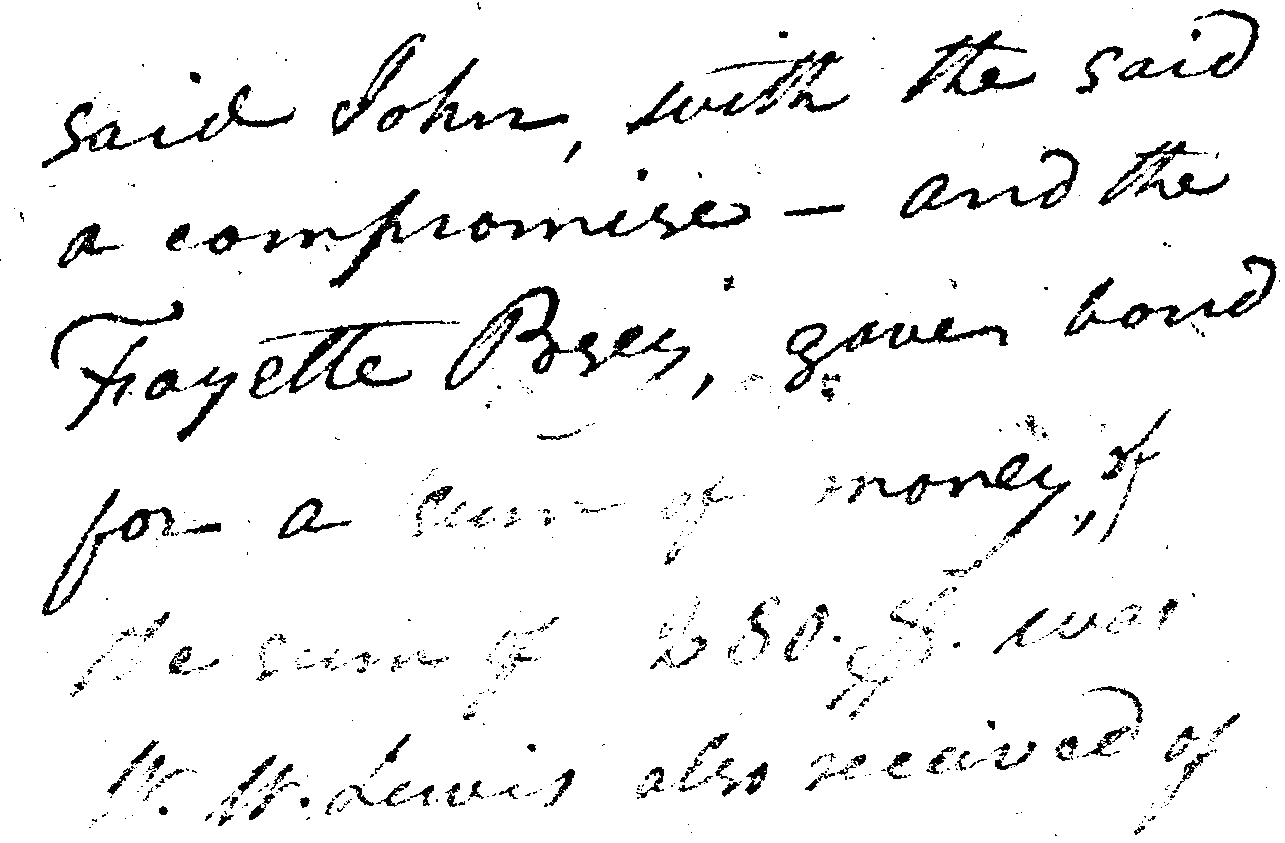}} &
{\includegraphics[width=0.45\columnwidth]{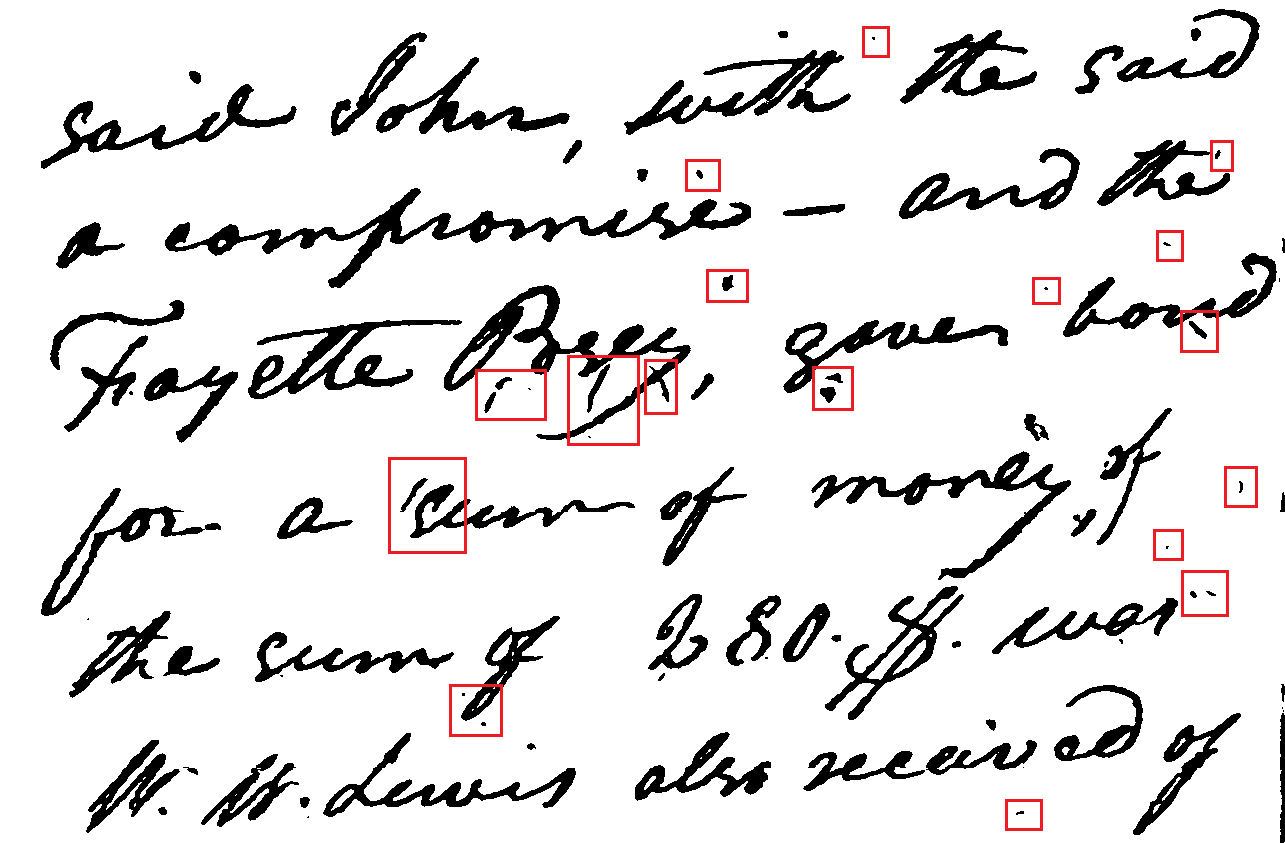}}&
{\includegraphics[width=0.45\columnwidth]{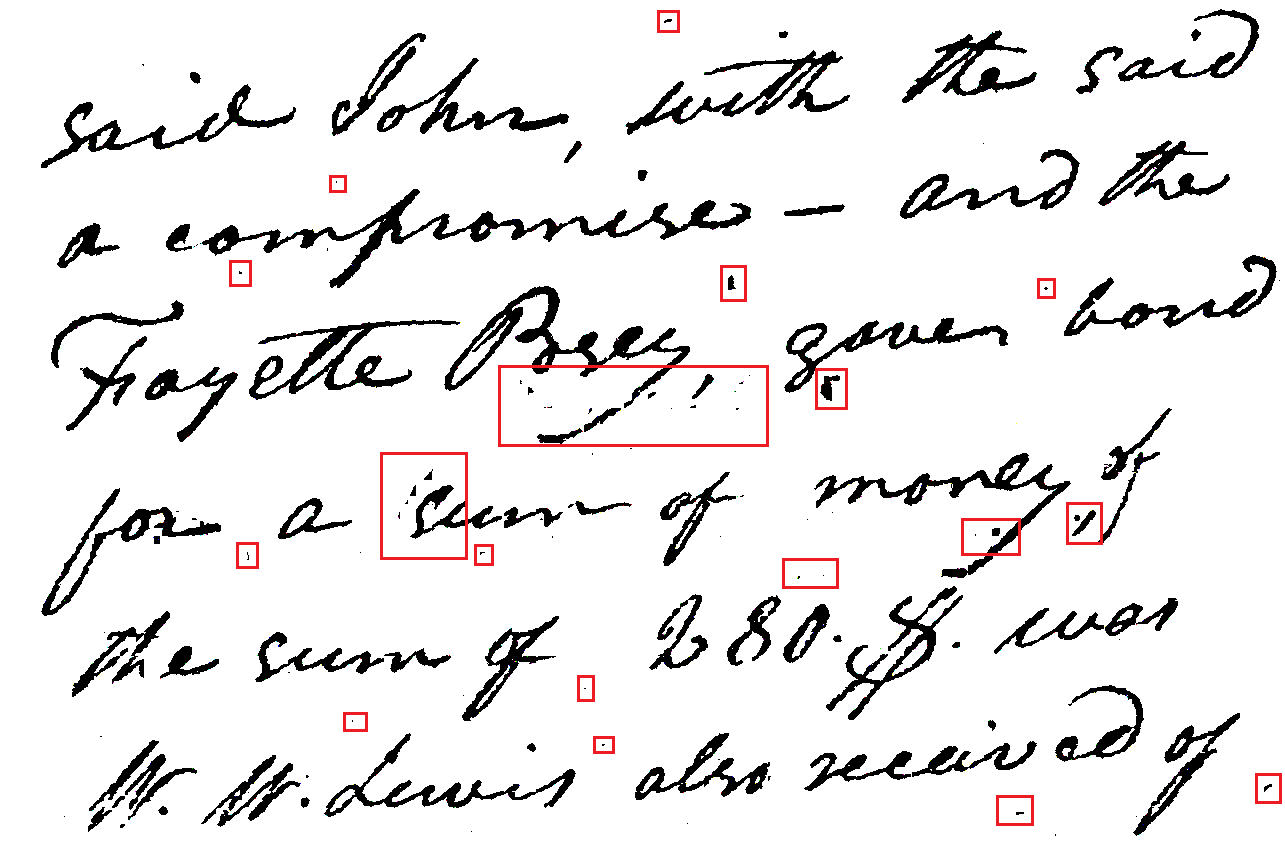}} &
{\includegraphics[width = 0.45\columnwidth]{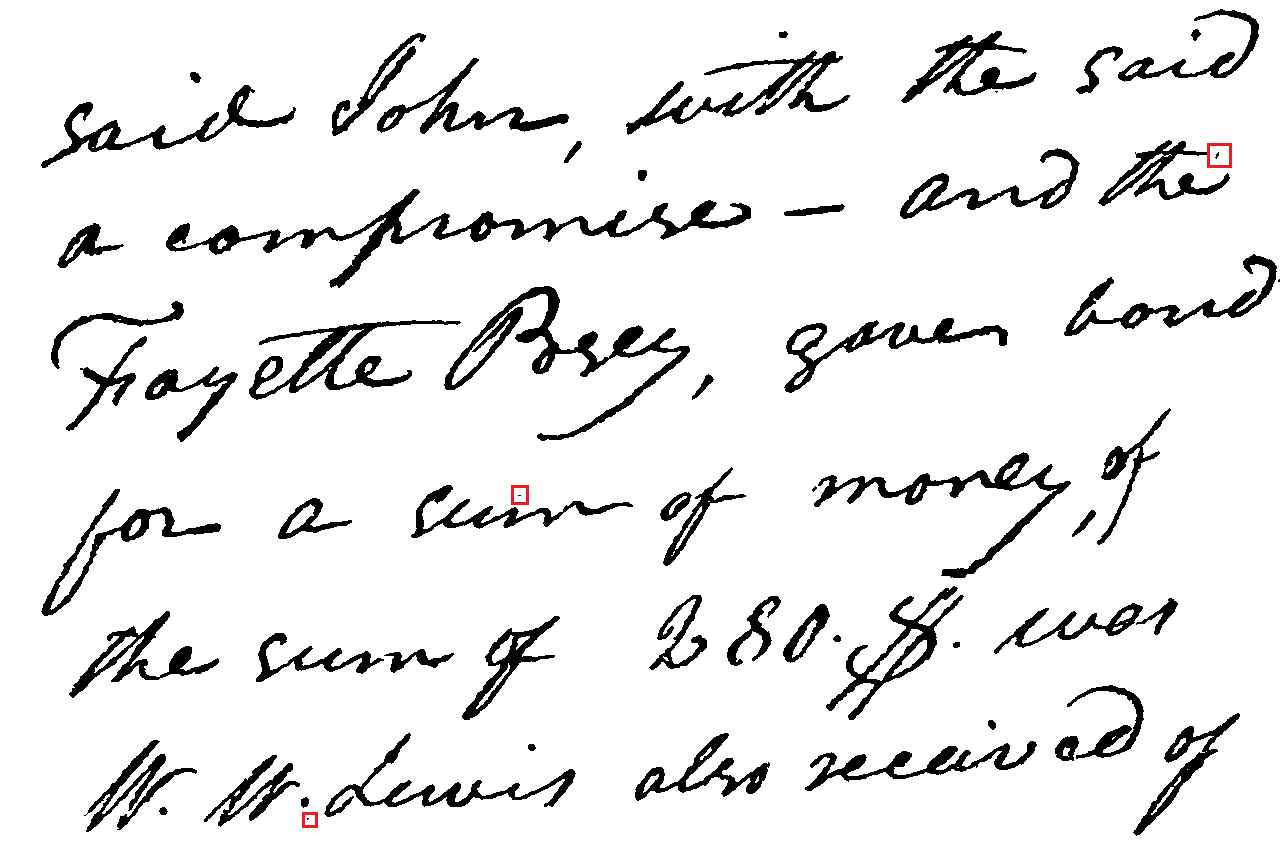}} &\\
\scriptsize e) Bradley~\cite{bradley2007adaptive} & \scriptsize f) DE-GAN~\cite{souibgui2020gan}  & \scriptsize g) DocEnTr~\cite{souibgui2022docentr}  &  \scriptsize h) \textbf{DocBinFormer}\\
(\textcolor{blue}{PSNR = 16.88}) & (\textcolor{blue}{PSNR = 15.87}) & (\textcolor{blue}{PSNR = 20.13}) & (\textcolor{blue}{PSNR = 21.43})\\
\end{tabular}
\caption{Qualitative performance of the various binarization techniques on sample no. 2 (shown in (a)) from DIBCO 2012.}
\label{fig:compDIBCO2012_2}
\end{figure*}
\begin{figure*}[!ht]
\centering
\begin{tabular}{ccccc}
{\includegraphics[width = 0.45\columnwidth]{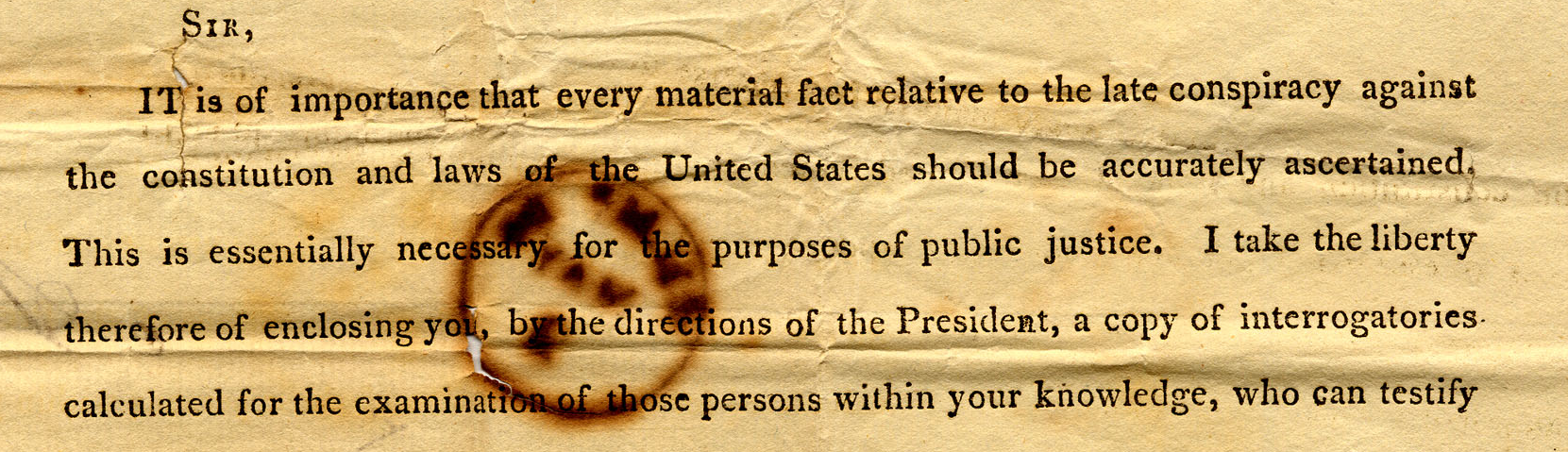}} &
{\includegraphics[width=0.45\columnwidth]{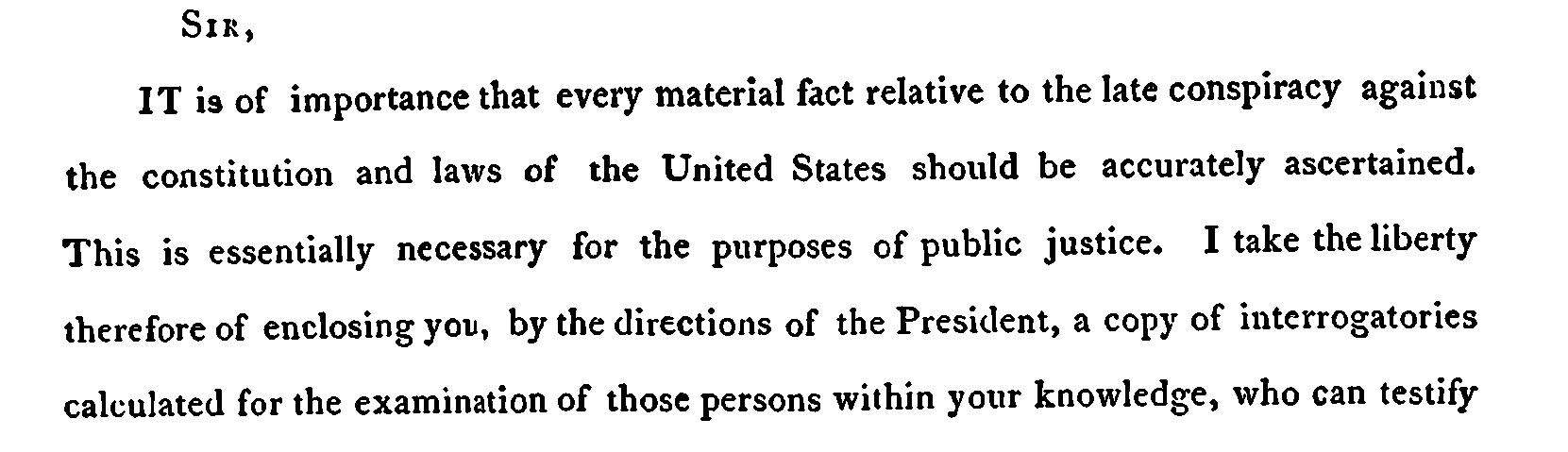}} &
{\includegraphics[width=0.45\columnwidth]{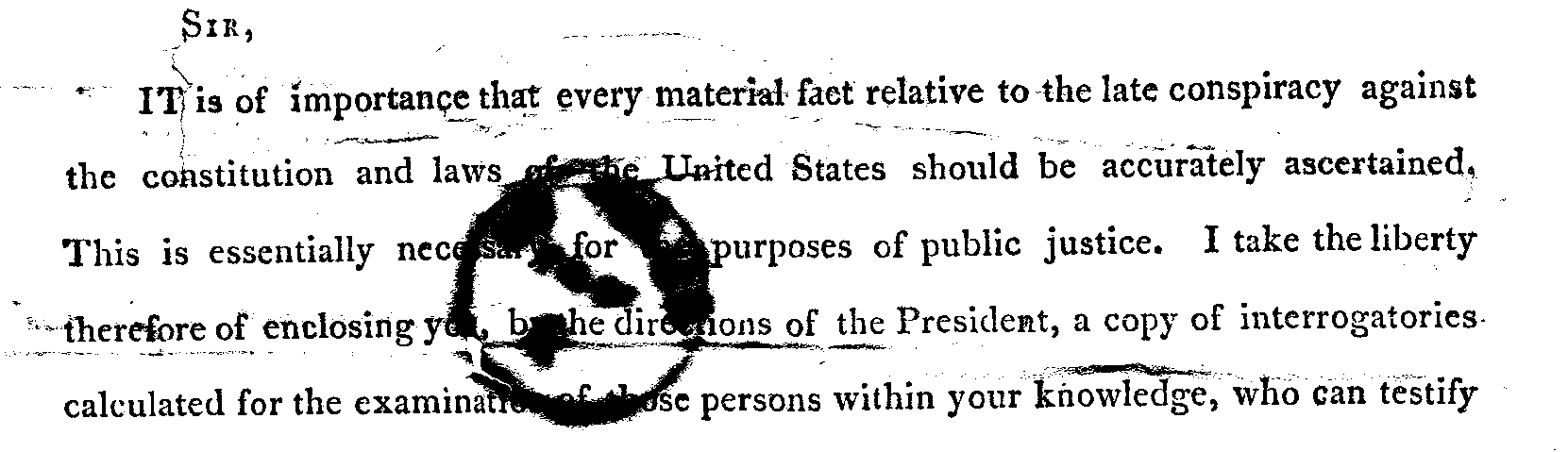}}&
{\includegraphics[width = 0.45\columnwidth]{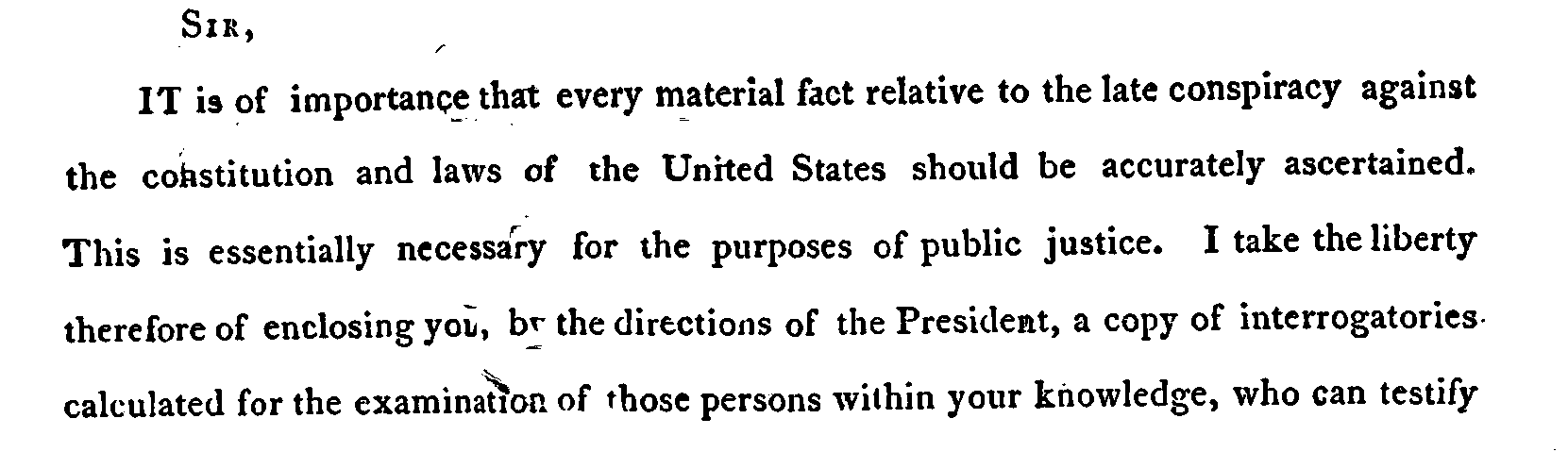}} & \\
\scriptsize a) Original &  \scriptsize b) GT  &  \scriptsize c) Otsu~\cite{otsu1979threshold}  &  \scriptsize d) SauvolaNet~\cite{li2021sauvolanet}\\ 
& & (\textcolor{blue}{PSNR = 12.79}) &  (\textcolor{blue}{PSNR = 22.20}) \\
{\includegraphics[width=0.45\columnwidth]{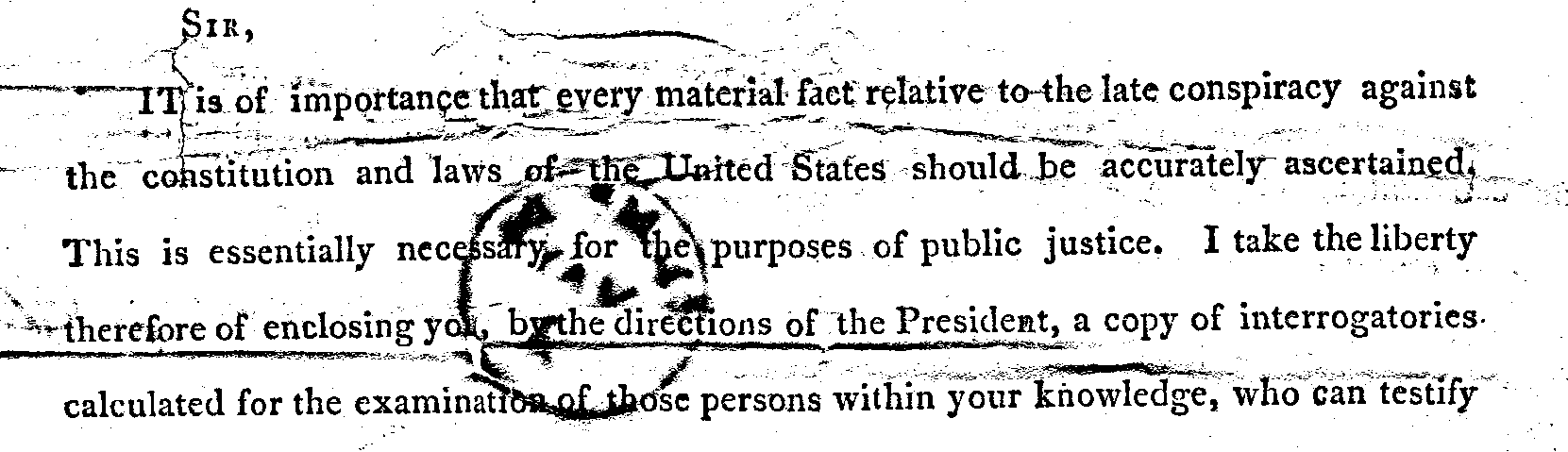}} &
{\includegraphics[width=0.45\columnwidth]{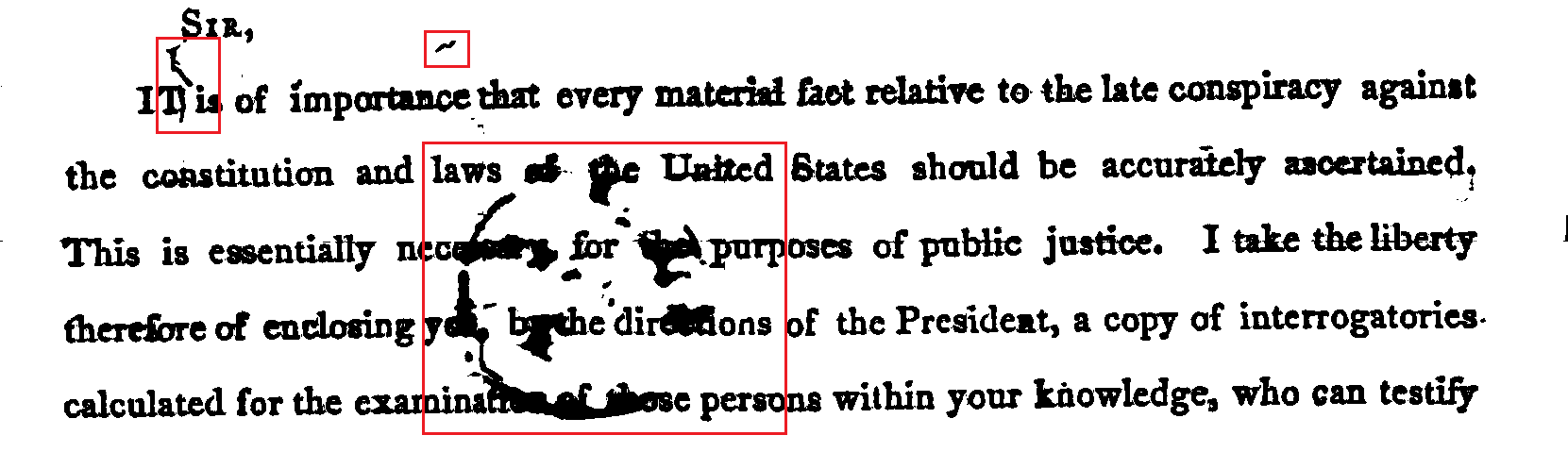}}&
{\includegraphics[width=0.45\columnwidth]{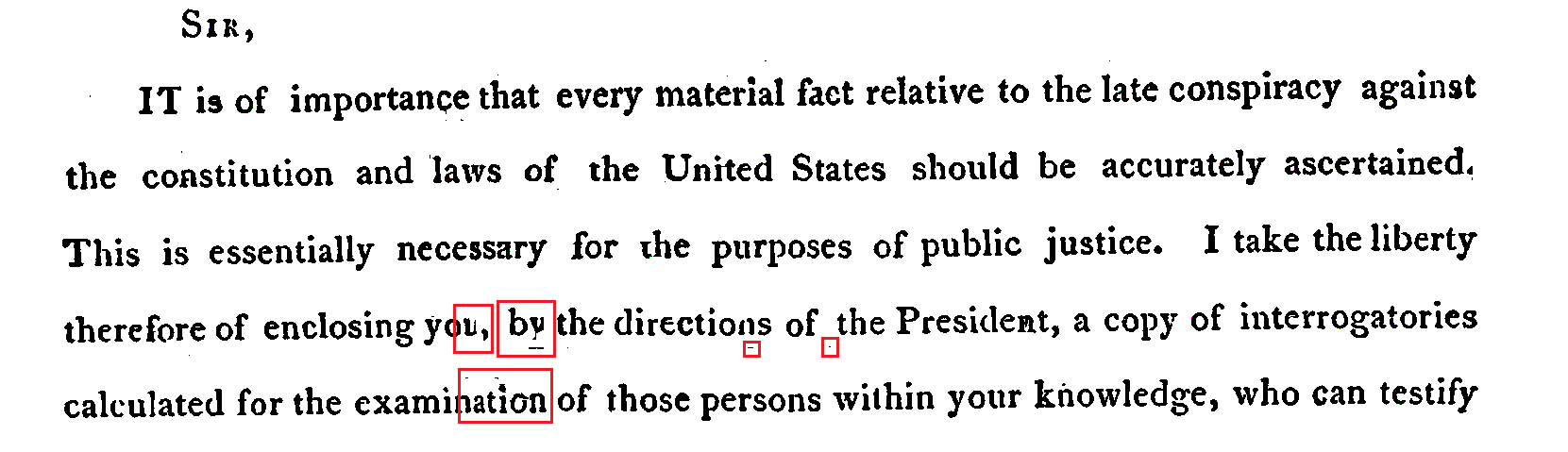}} &
{\includegraphics[width = 0.45\columnwidth]{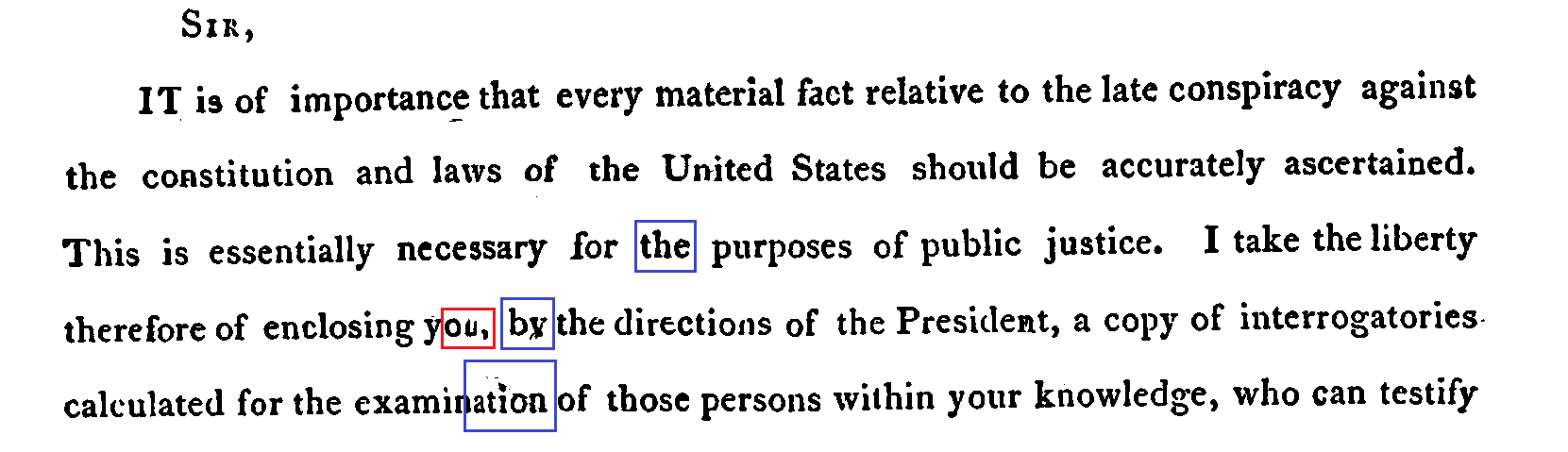}} &\\
\scriptsize e) Bradley~\cite{bradley2007adaptive} & \scriptsize f) DE-GAN~\cite{souibgui2020gan}  & \scriptsize g) DocEnTr~\cite{souibgui2022docentr}  &  \scriptsize h) \textbf{DocBinFormer}\\
(\textcolor{blue}{PSNR = 11.10}) & (\textcolor{blue}{PSNR = 13.39}) & (\textcolor{blue}{PSNR = 21.89}) & (\textcolor{blue}{PSNR = 22.12})\\
\end{tabular}
\caption{Qualitative performance of the various binarization techniques on sample no. 16 (shown in (a)) from DIBCO 2013.}
\label{fig:compDIBCO2013_1}
\end{figure*}
\begin{figure*}[!ht]
\centering
\begin{tabular}{ccccc}
{\includegraphics[width = 0.45\columnwidth]{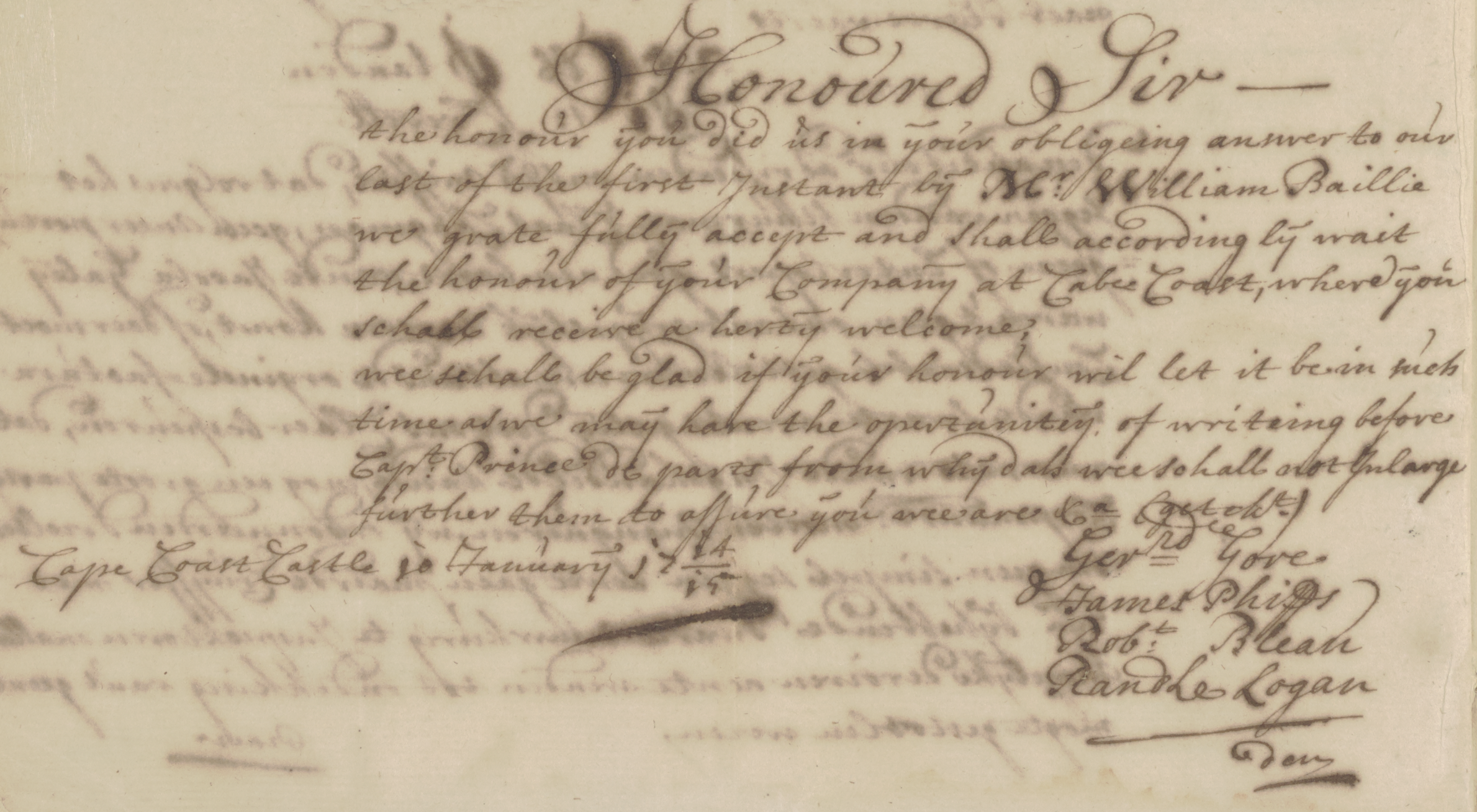}} &
{\includegraphics[width=0.45\columnwidth]{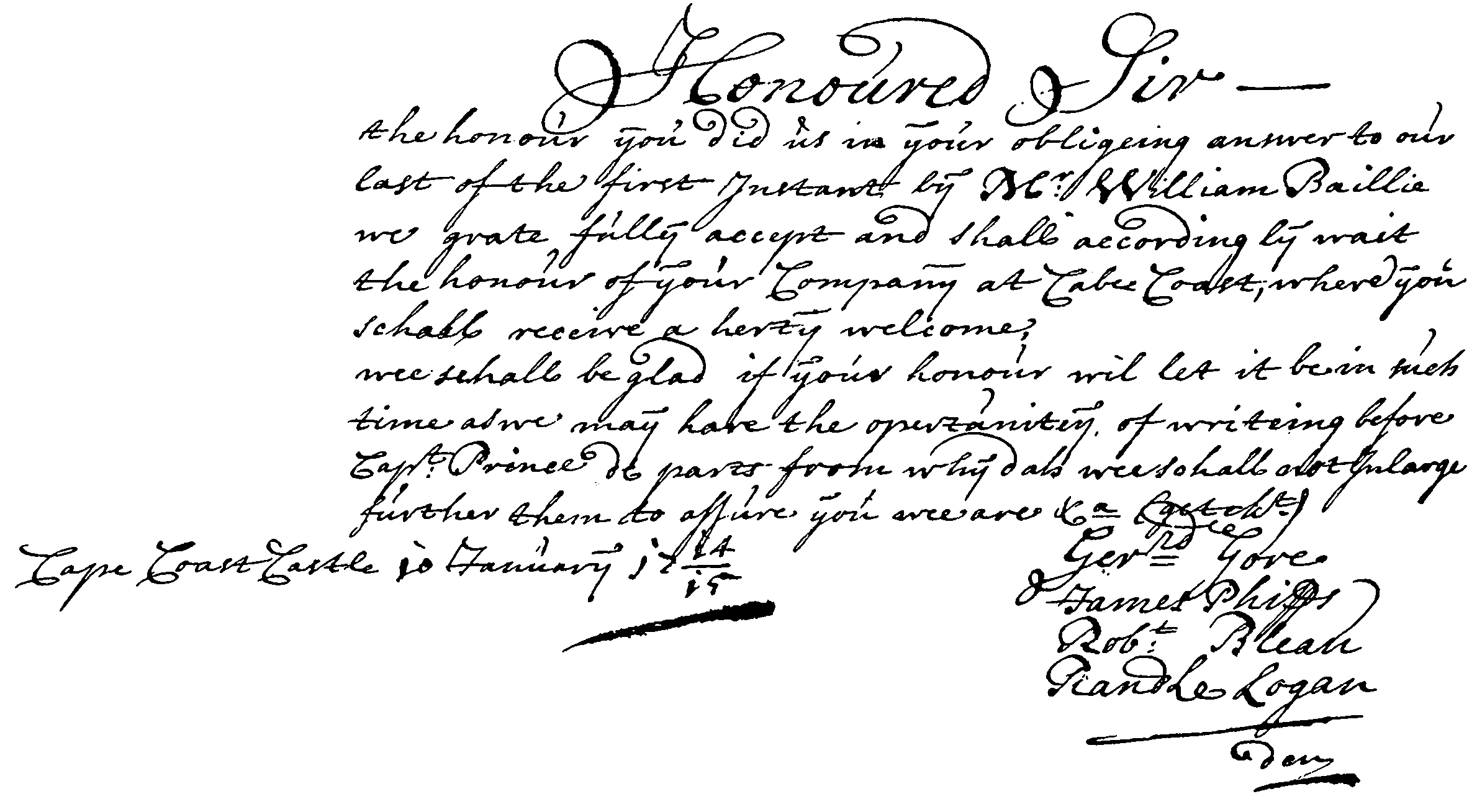}} &
{\includegraphics[width=0.45\columnwidth]{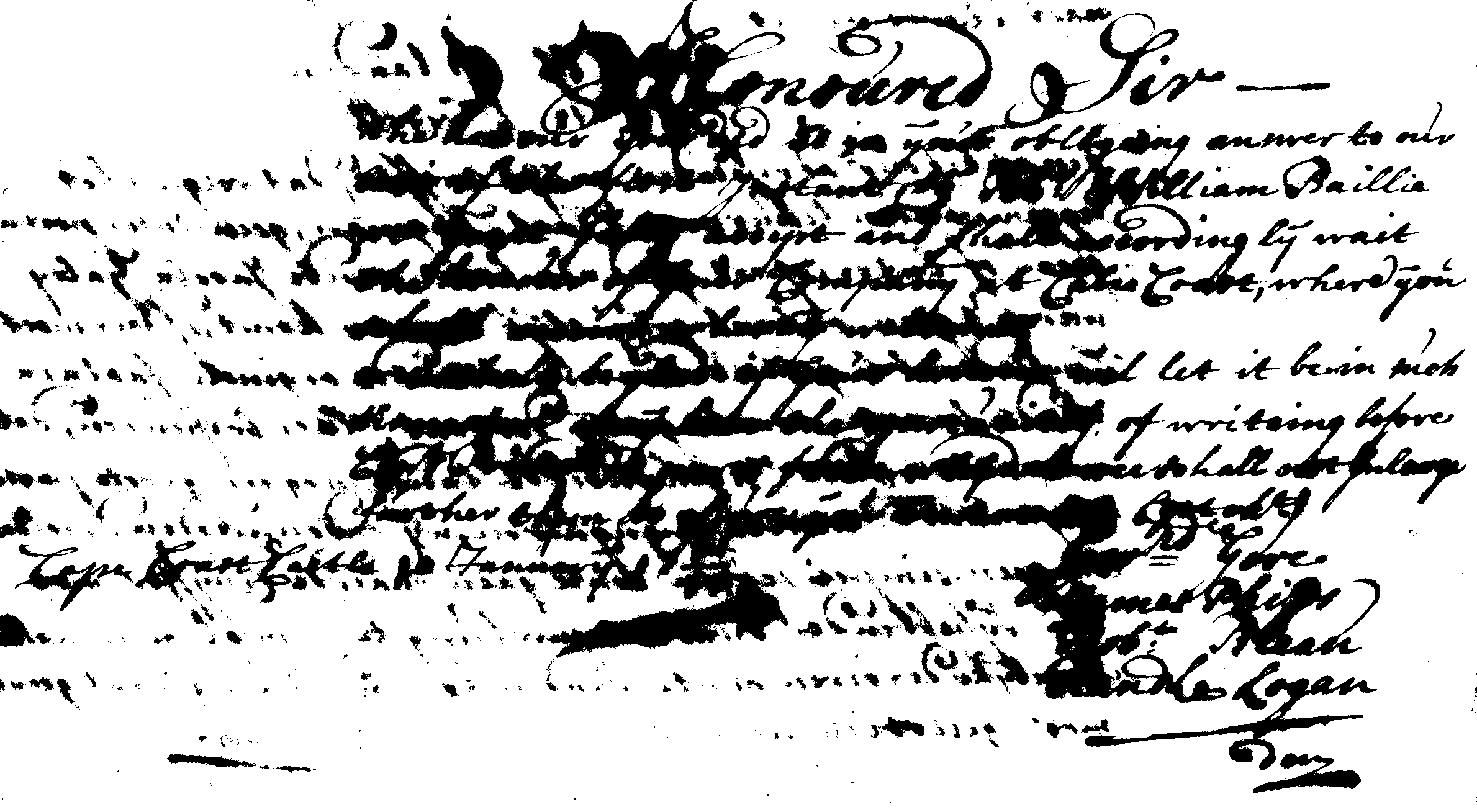}}&
{\includegraphics[width = 0.45\columnwidth]{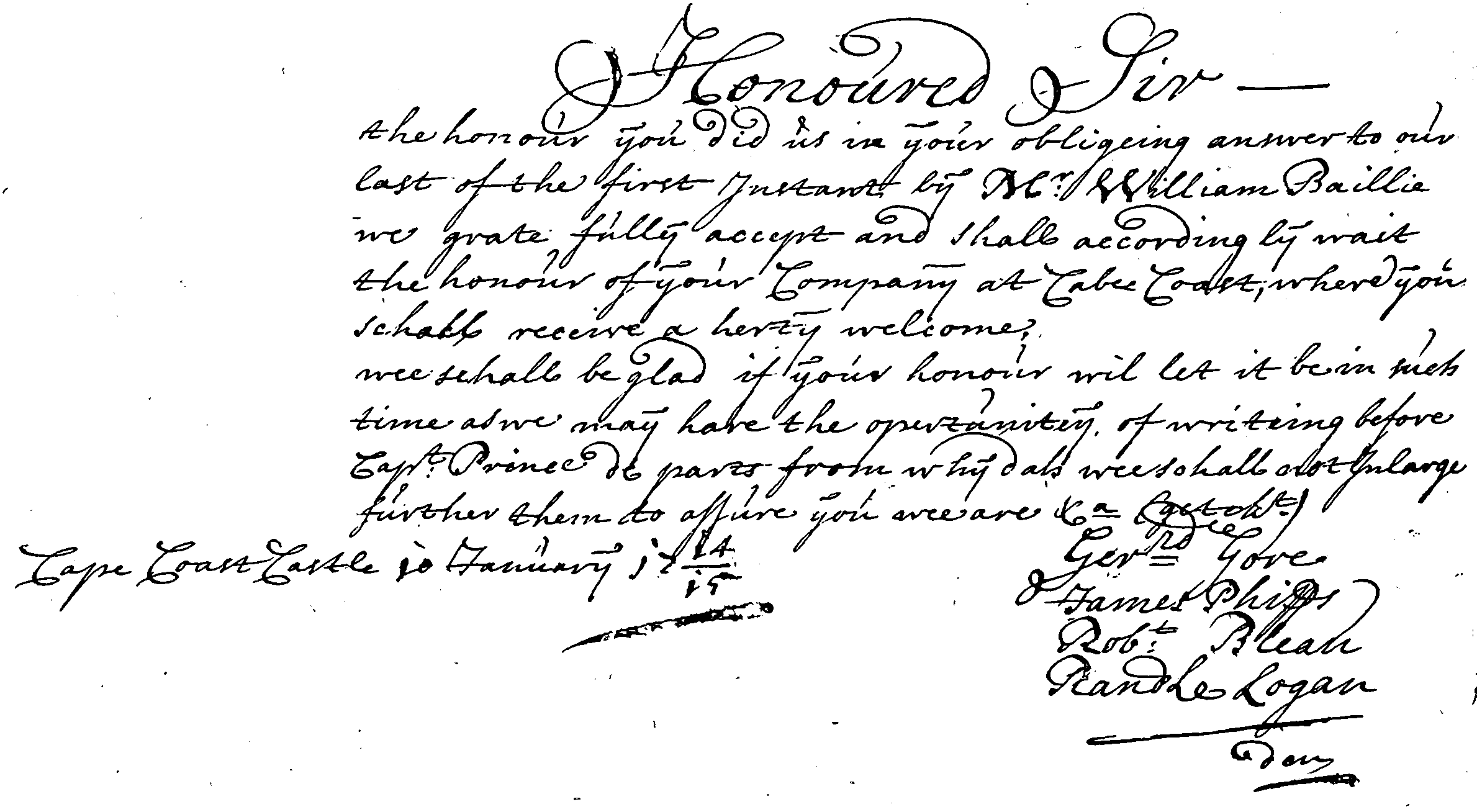}} & \\
\scriptsize a) Original &  \scriptsize b) GT  &  \scriptsize c) Otsu~\cite{otsu1979threshold}  &  \scriptsize d) SauvolaNet~\cite{li2021sauvolanet}\\  
& & (\textcolor{blue}{PSNR = 7.82}) &  (\textcolor{blue}{PSNR = 20.43}) \\
{\includegraphics[width=0.45\columnwidth]{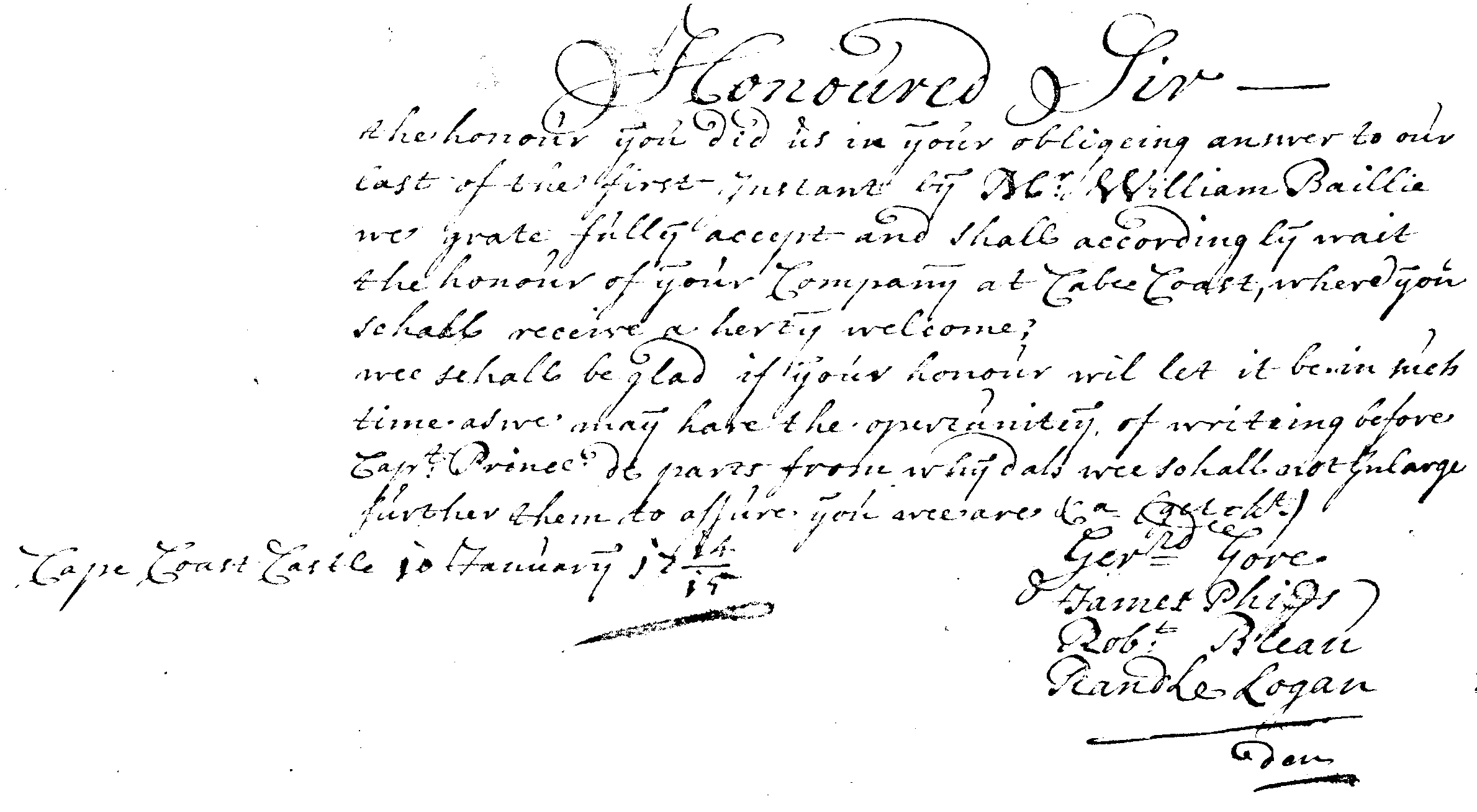}} &
{\includegraphics[width=0.45\columnwidth]{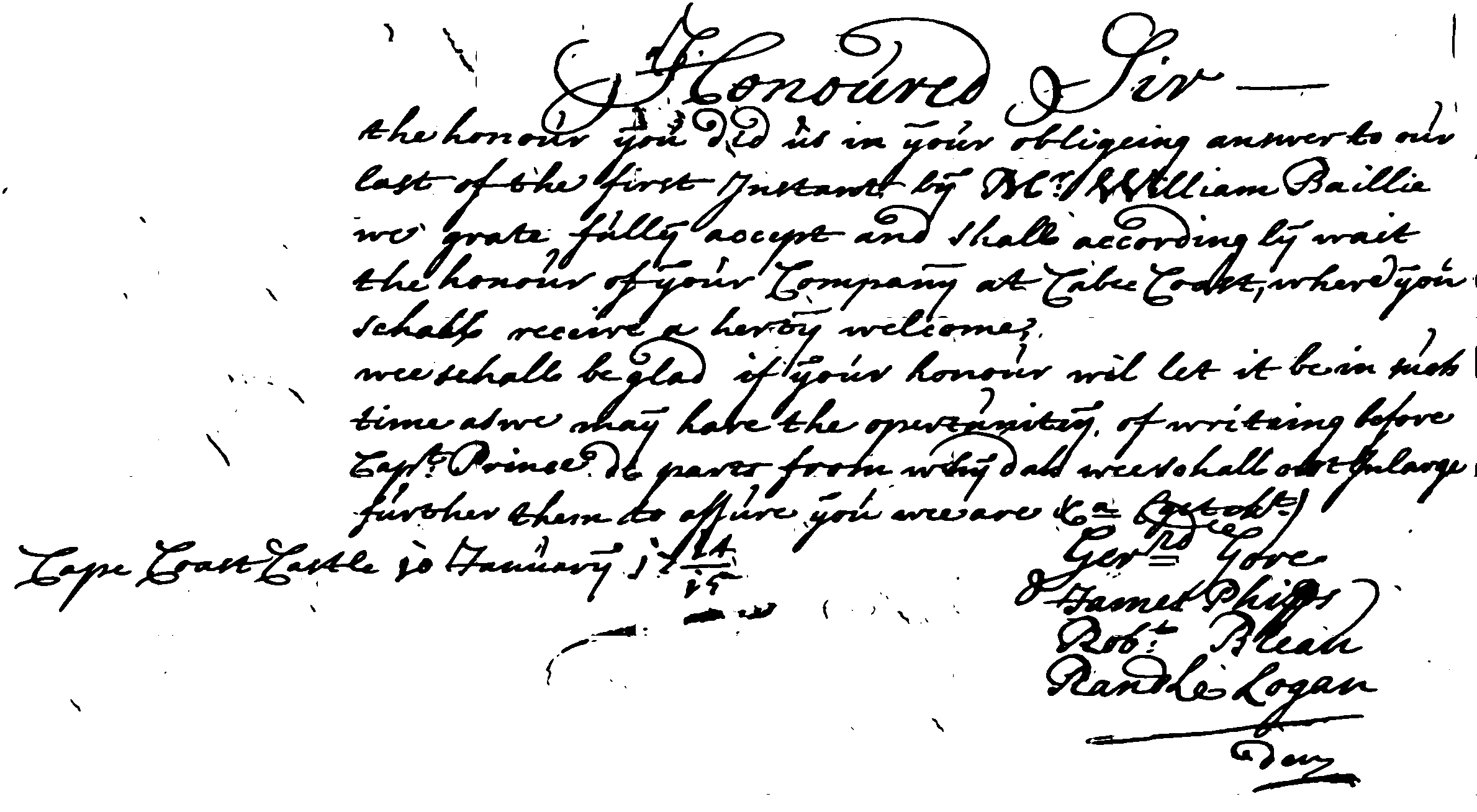}}&
{\includegraphics[width=0.45\columnwidth]{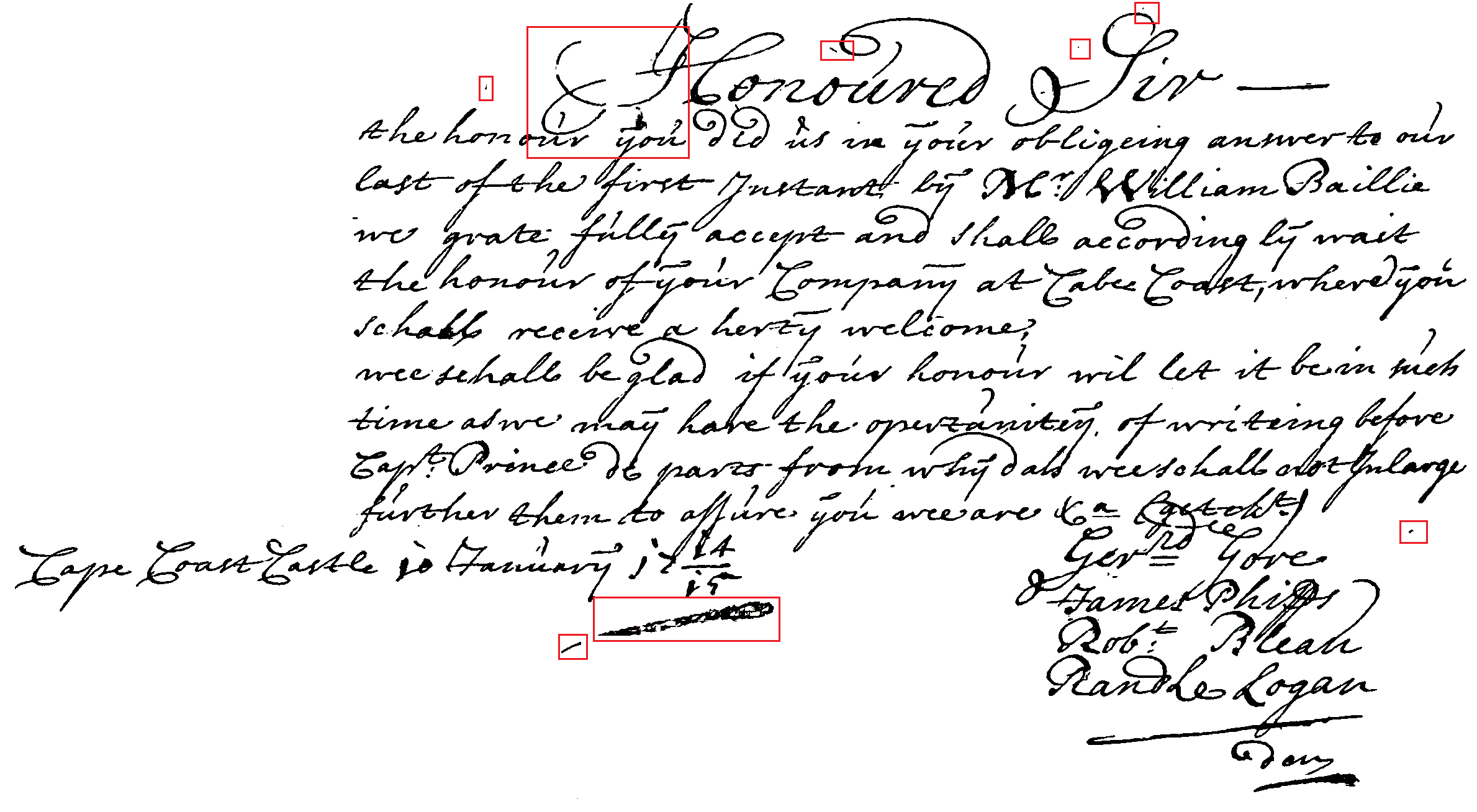}} &
{\includegraphics[width = 0.45\columnwidth]{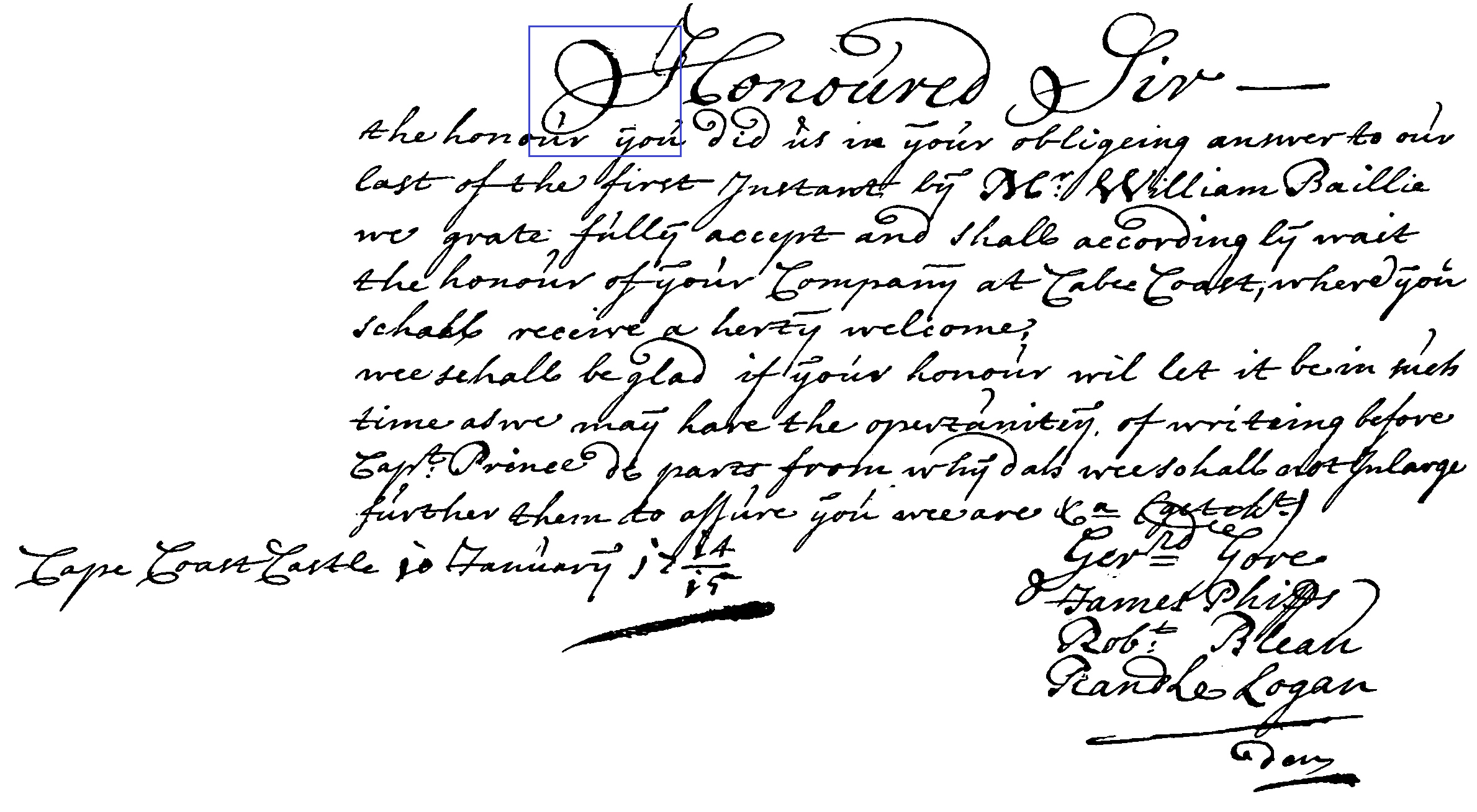}} &\\
\scriptsize e) Bradley~\cite{bradley2007adaptive} & \scriptsize f) DE-GAN~\cite{souibgui2020gan}  & \scriptsize g) DocEnTr~\cite{souibgui2022docentr}  &  \scriptsize h) \textbf{DocBinFormer}\\
(\textcolor{blue}{PSNR = 13.65}) & (\textcolor{blue}{PSNR = 14.23}) & (\textcolor{blue}{PSNR = 19.60}) & (\textcolor{blue}{PSNR = 21.55})\\
\end{tabular}
\caption{Qualitative performance of the various binarization techniques on sample No. 14 (shown in (a)) from DIBCO 2017.}
\label{fig:compDIBCO2017}
\end{figure*}
\begin{figure*}[ht!]
\centering
\begin{tabular}{ccccc}
{\includegraphics[width = 0.45\columnwidth]{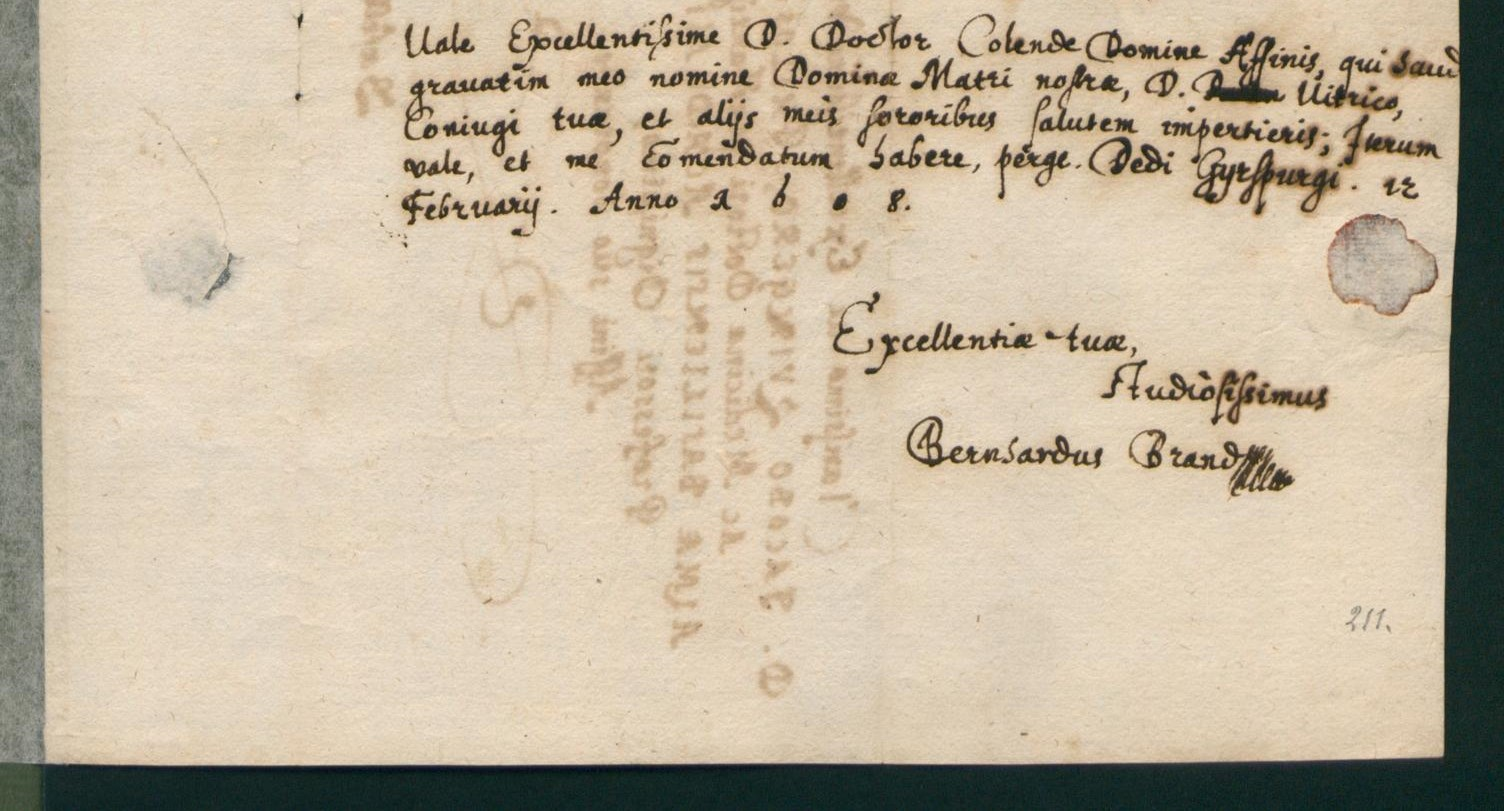}} &
{\includegraphics[width=0.45\columnwidth]{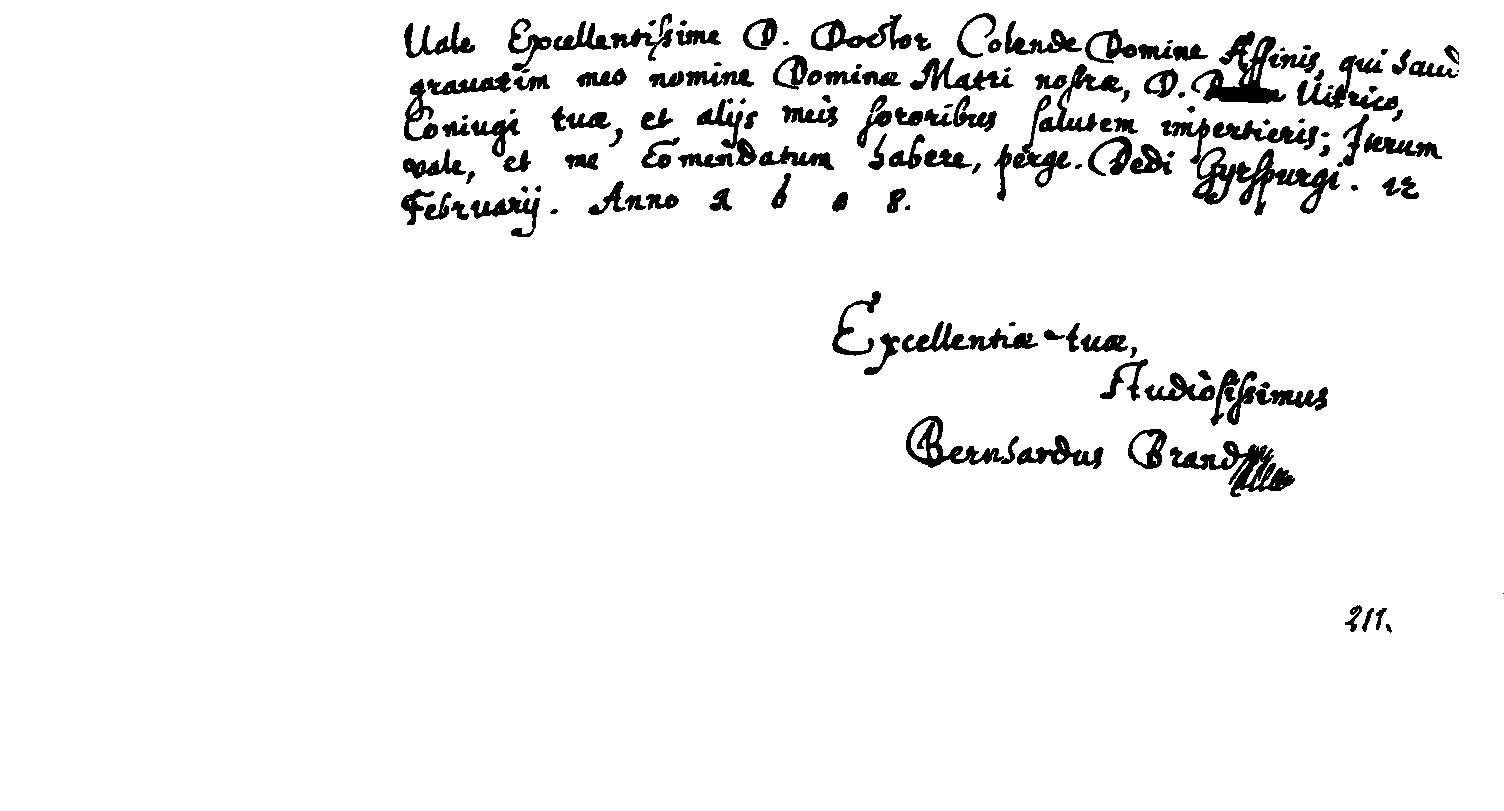}} &
{\includegraphics[width=0.45\columnwidth]{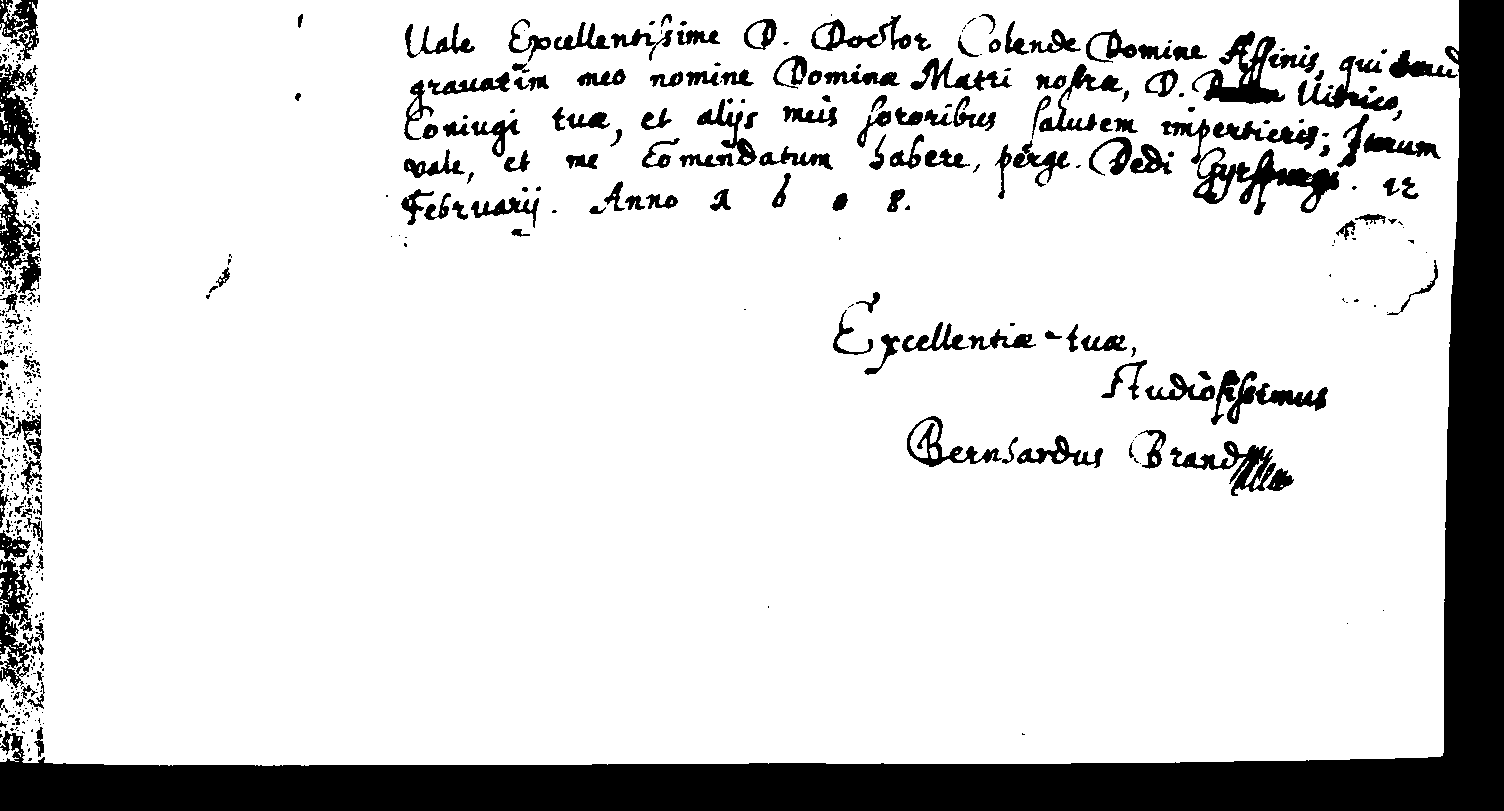}}&
{\includegraphics[width = 0.45\columnwidth]{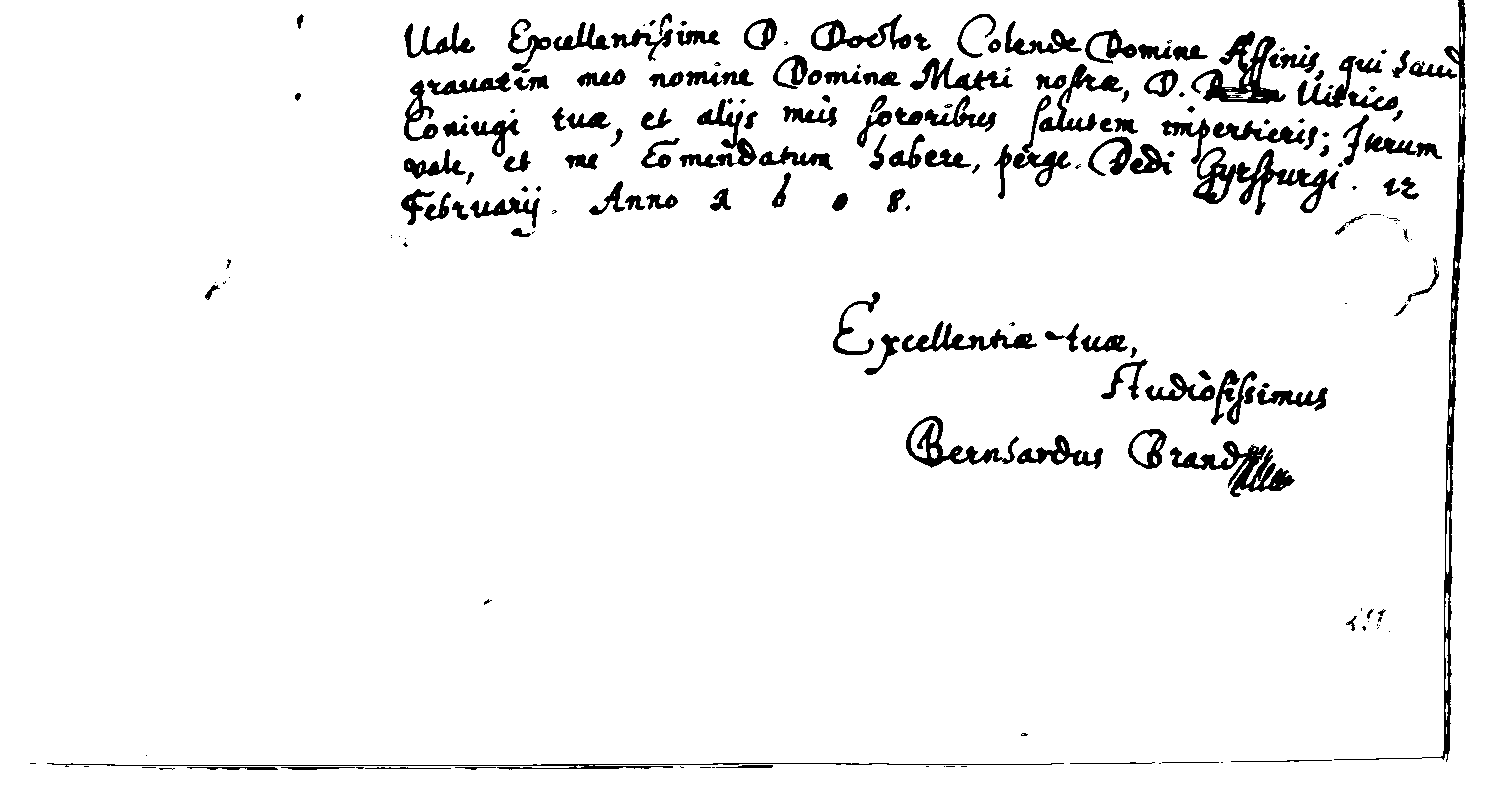}} & \\
\scriptsize a) Original &  \scriptsize b) GT  &  \scriptsize c) Otsu~\cite{otsu1979threshold}  &  \scriptsize d) SauvolaNet~\cite{li2021sauvolanet}\\ 
& & (\textcolor{blue}{PSNR = 9.45}) &  (\textcolor{blue}{PSNR = 19.21}) \\
{\includegraphics[width=0.45\columnwidth]{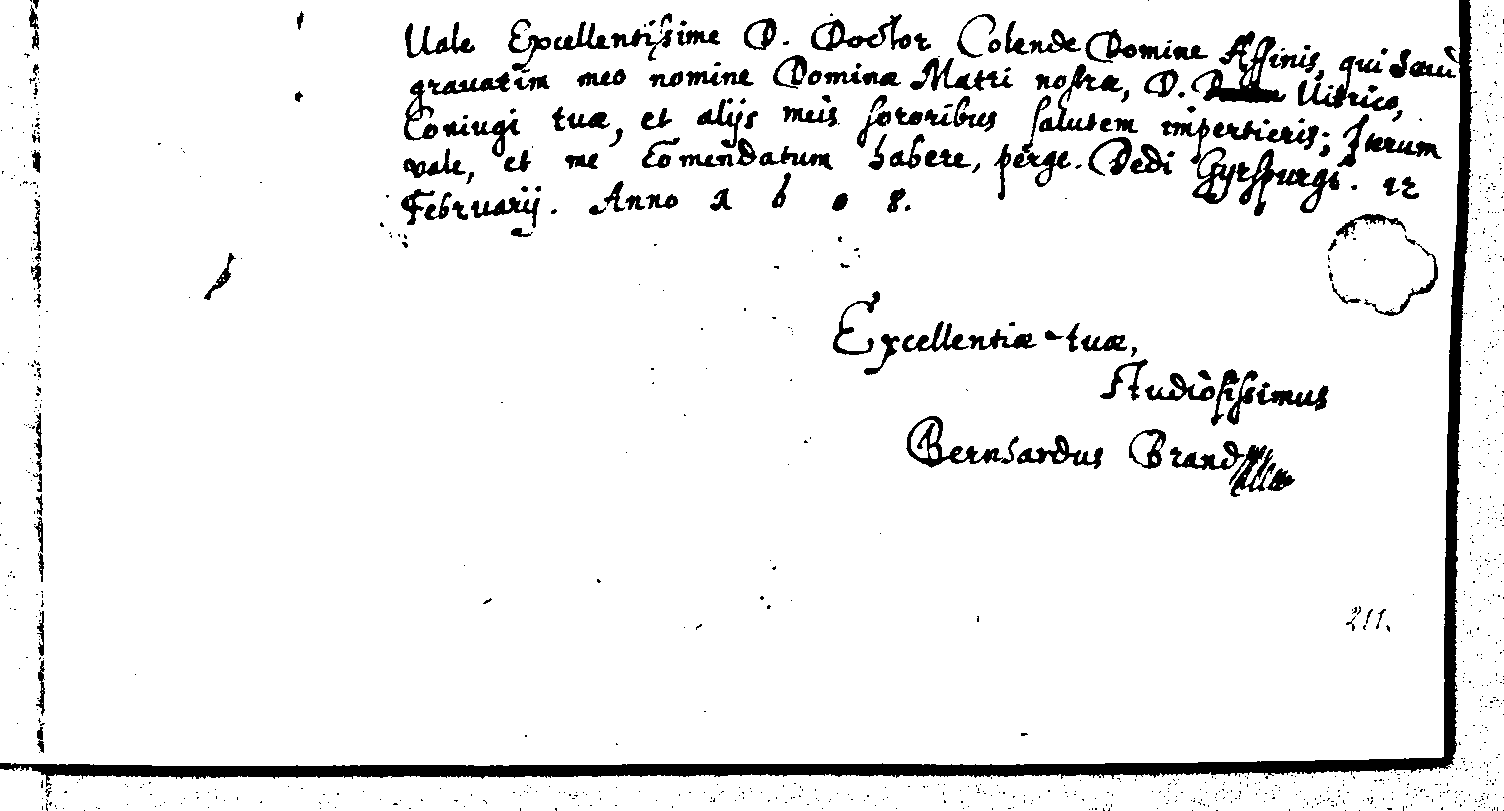}} &
{\includegraphics[width=0.45\columnwidth]{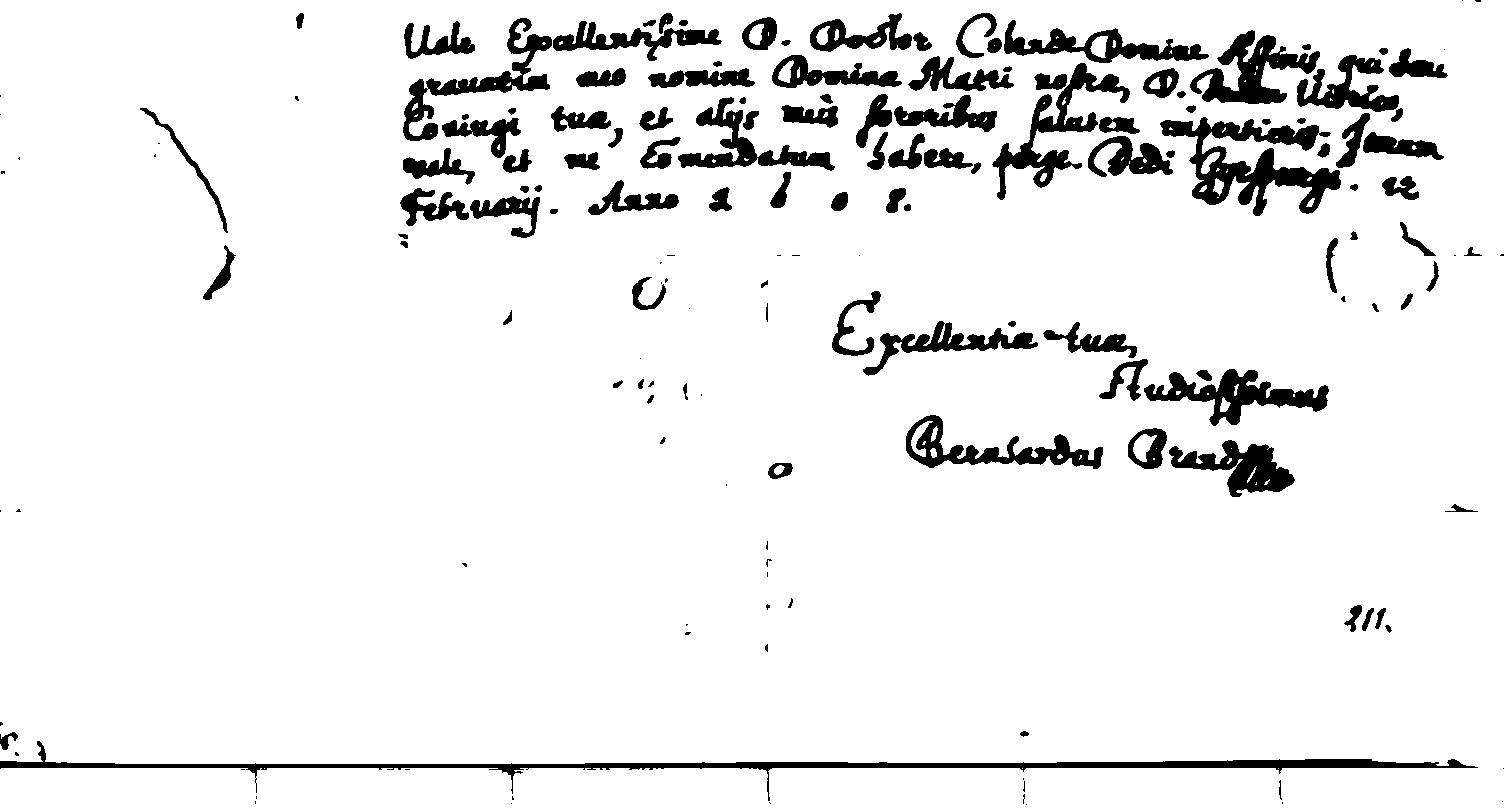}}&
{\includegraphics[width=0.45\columnwidth]{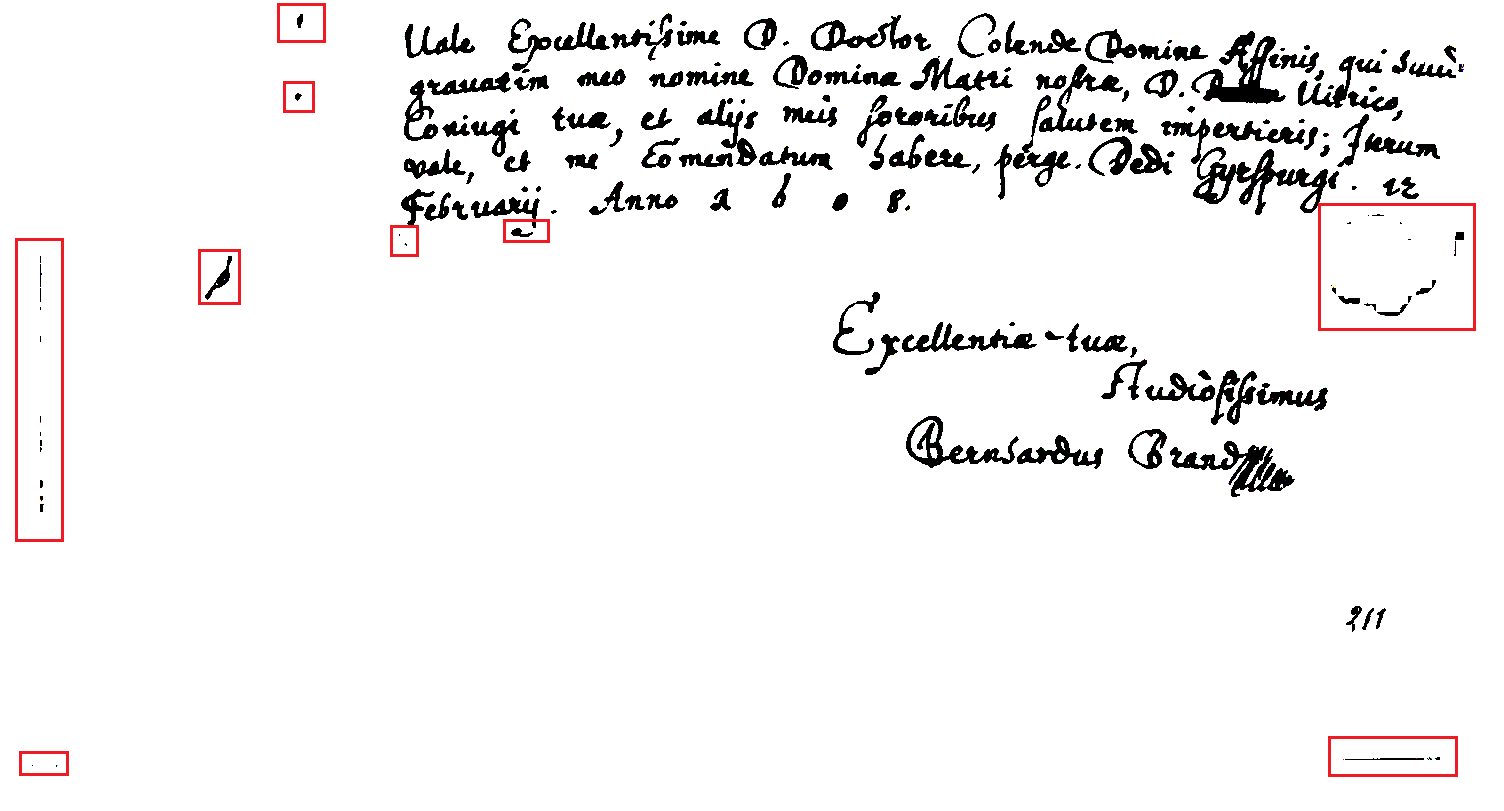}} &
{\includegraphics[width = 0.45\columnwidth]{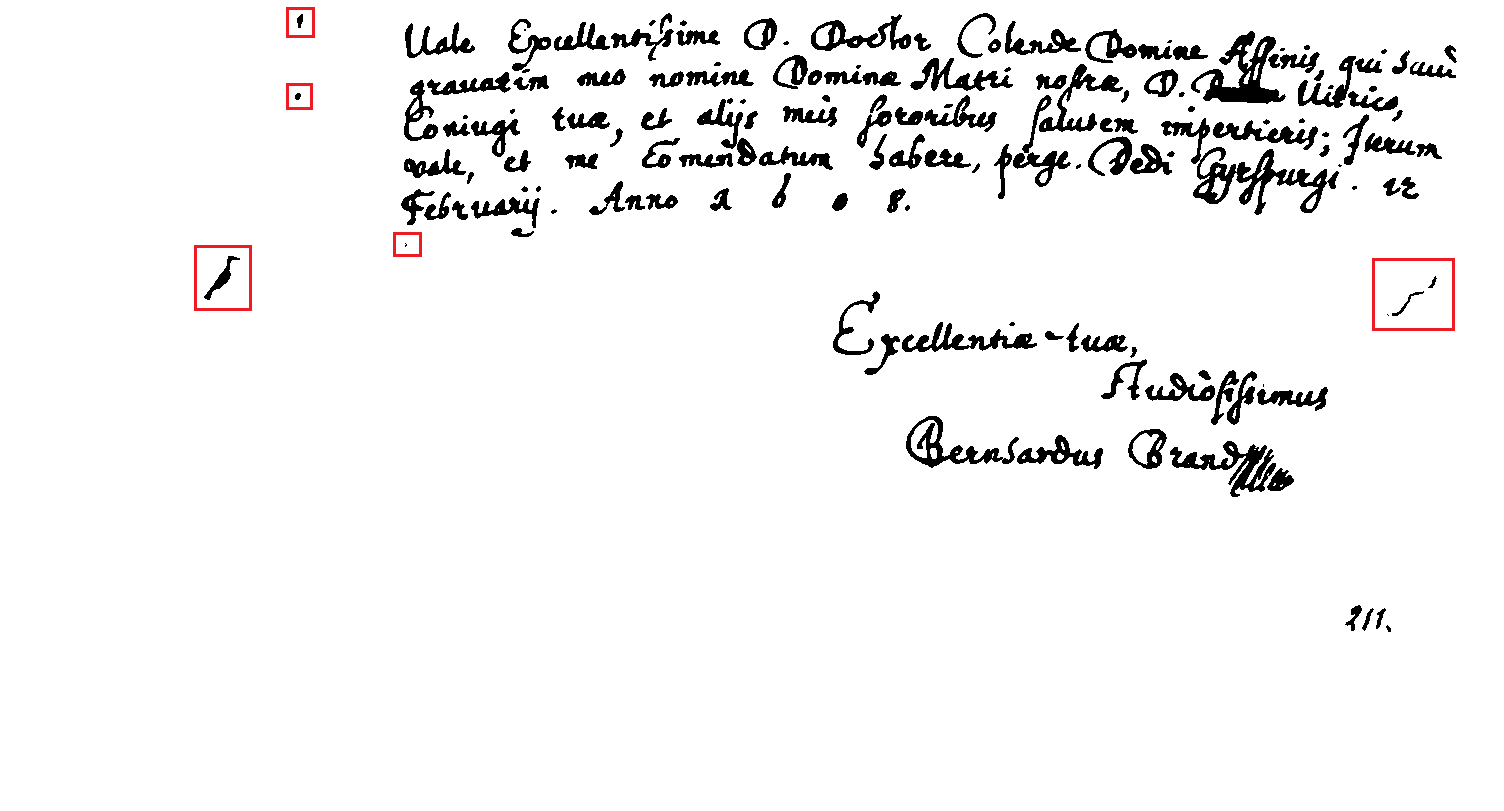}} &\\
\scriptsize e) Bradley~\cite{bradley2007adaptive} & \scriptsize f) DE-GAN~\cite{souibgui2020gan}  & \scriptsize g) DocEnTr~\cite{souibgui2022docentr}  &  \scriptsize h) \textbf{DocBinFormer}\\
(\textcolor{blue}{PSNR = 8.97}) & (\textcolor{blue}{PSNR = 16.09}) & (\textcolor{blue}{PSNR = 21.67}) & (\textcolor{blue}{PSNR = 22.08})\\
\end{tabular}
\caption{Qualitative performance of the various binarization techniques on sample no. 2 (shown in (a)) from DIBCO 2018.}
\label{fig:compDIBCO2018}
\end{figure*}
\begin{figure*}[htbp]
\centering
\captionsetup{justification=centering}
\begin{tabular}{cccc}
\includegraphics[width=0.45\columnwidth,scale=0.25]{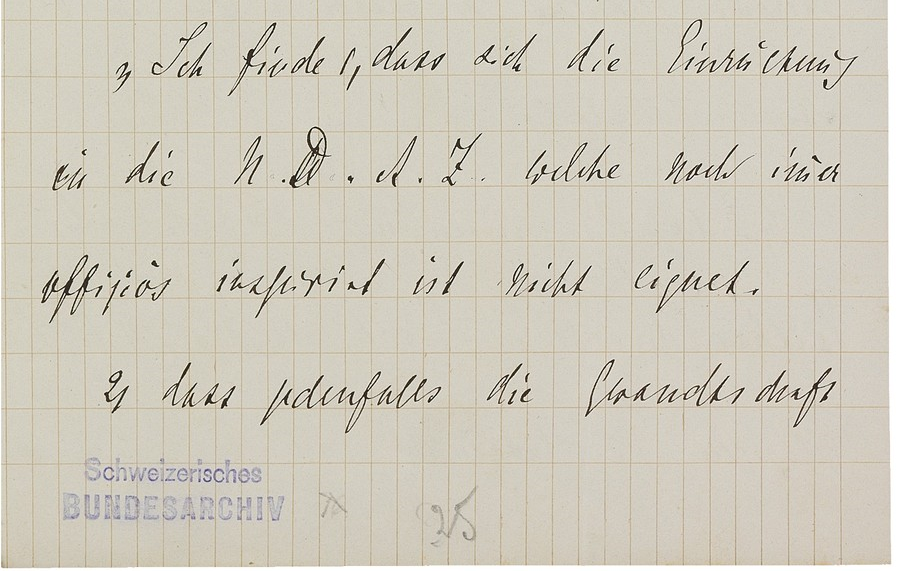} &
\includegraphics[width=0.45\columnwidth,scale=0.25]{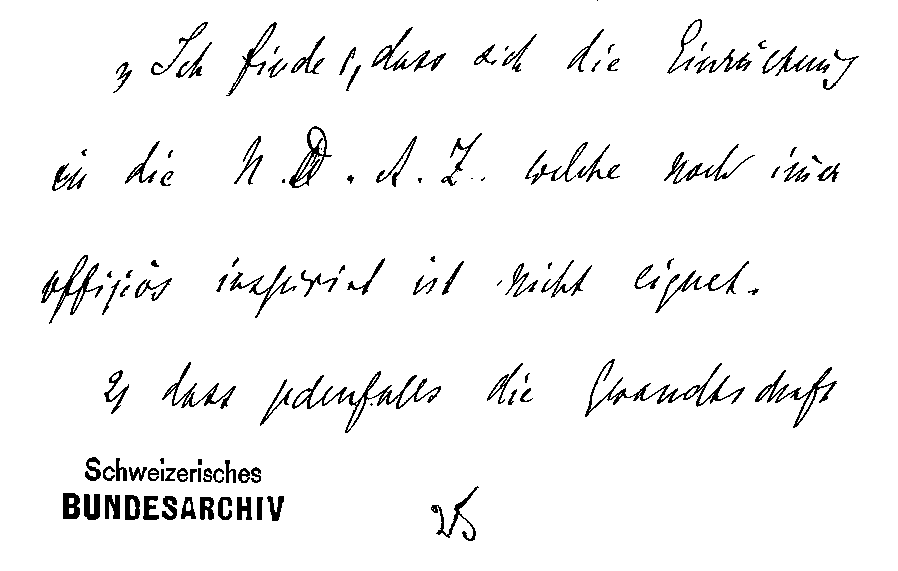} &
\includegraphics[width=0.45\columnwidth,scale=0.25]{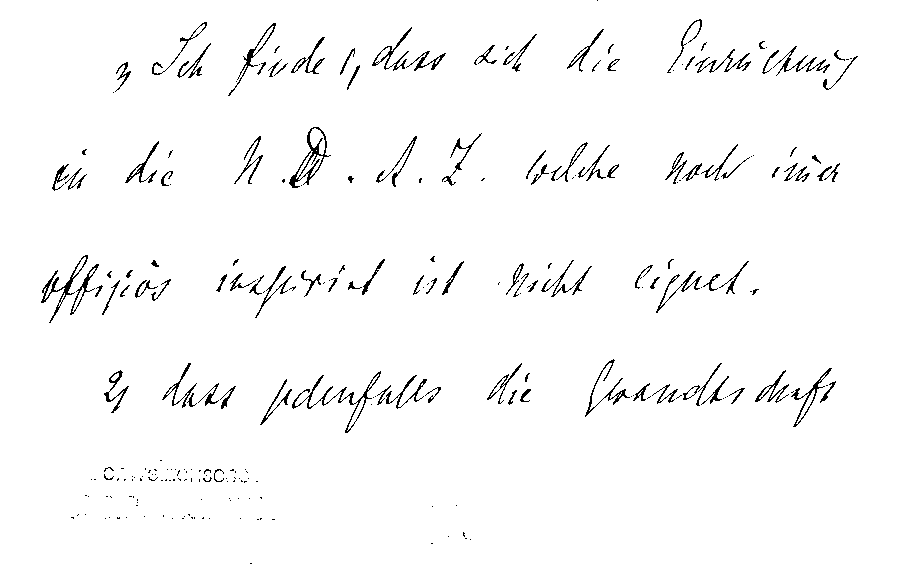} &
\includegraphics[width=0.45\columnwidth,scale=0.25]{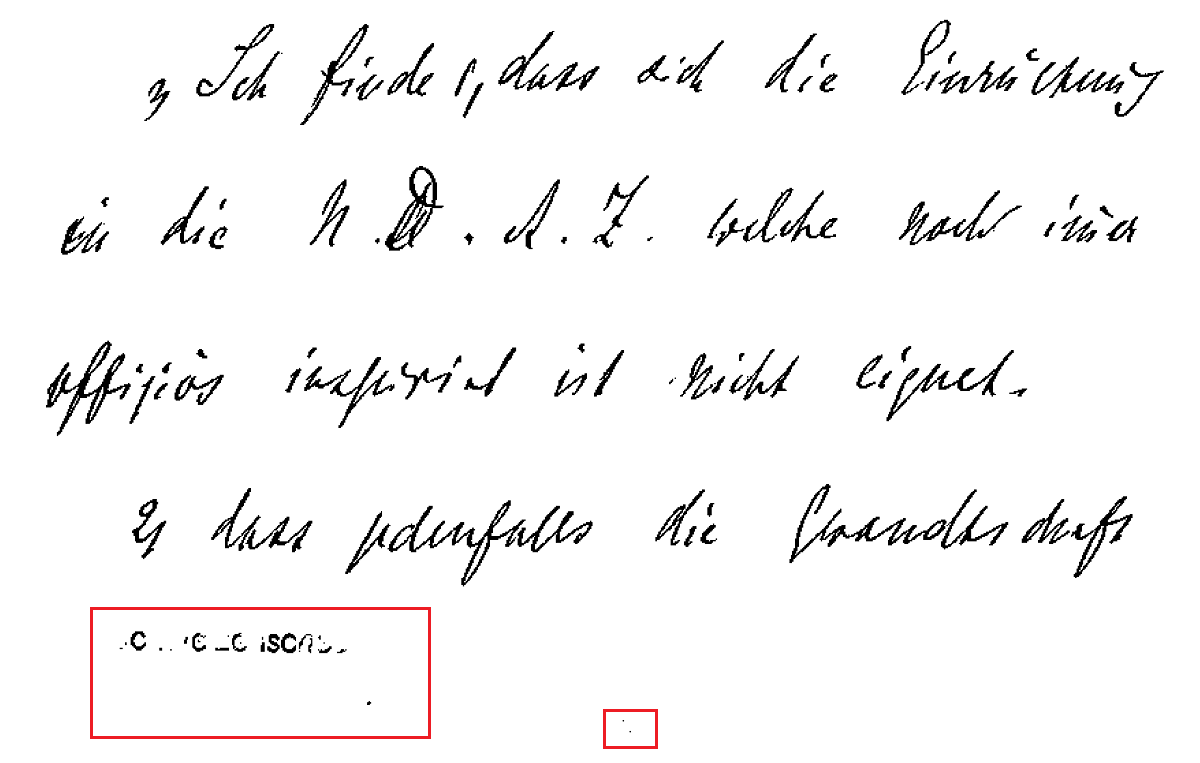} 
\\  \scriptsize a) Original &  \scriptsize b) GT  &  \scriptsize c) Otsu~\cite{otsu1979threshold}  &  \scriptsize d) SauvolaNet~\cite{li2021sauvolanet}\\  
& & (\textcolor{blue}{PSNR = 17.62}) &  (\textcolor{blue}{PSNR = 18.11}) \\
\includegraphics[width=0.45\columnwidth,scale=0.25]{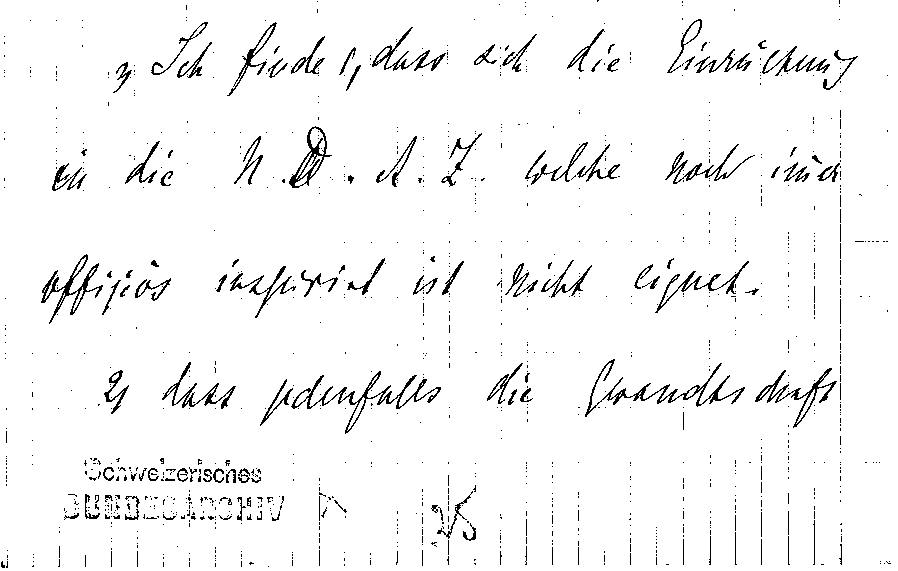} &
\includegraphics[width=0.45\columnwidth,scale=0.25]{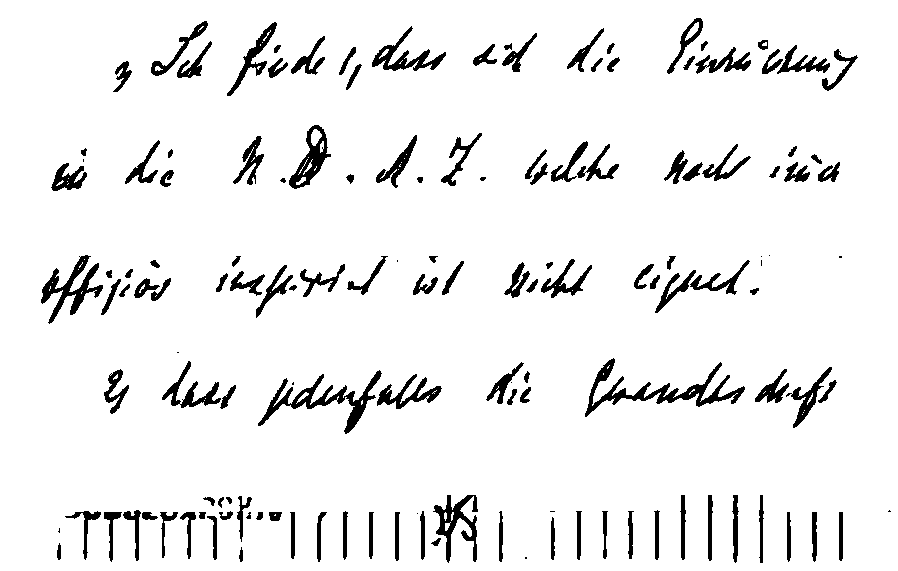} &
\includegraphics[width=0.45\columnwidth,scale=0.25]{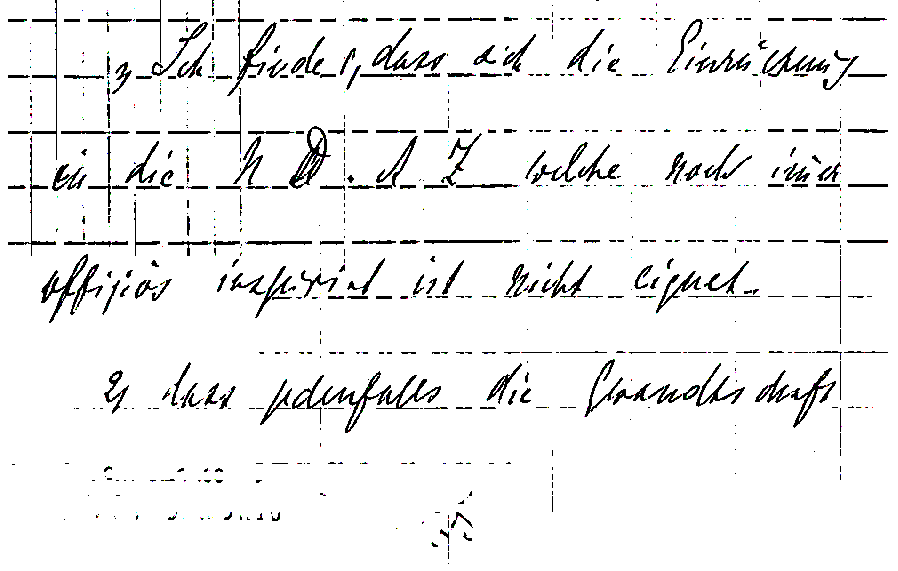} &
\includegraphics[width=0.45\columnwidth,scale=0.25]{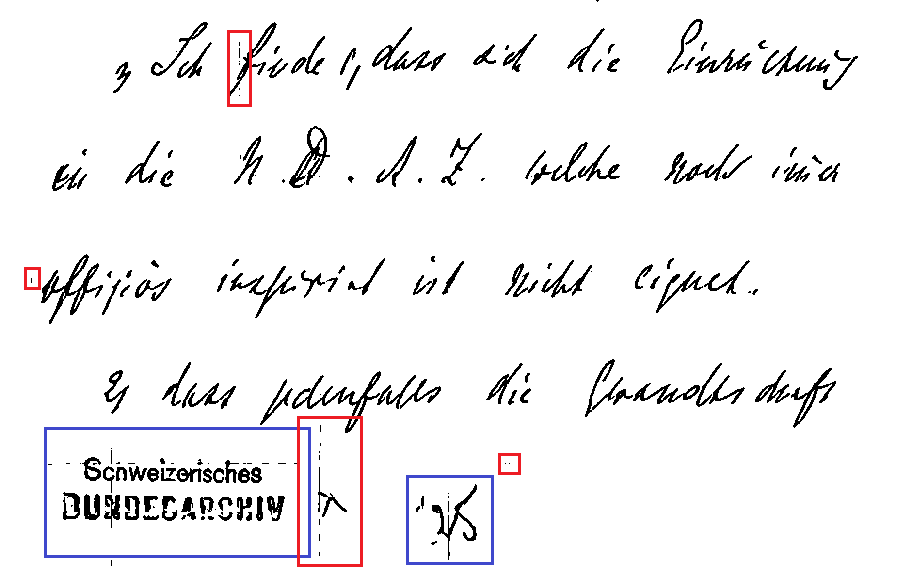} 
\\ \scriptsize e) Bradley~\cite{bradley2007adaptive} & \scriptsize f) DE-GAN~\cite{souibgui2020gan}  & \scriptsize g) DocEnTr~\cite{souibgui2022docentr}  &  \scriptsize h) \textbf{DocBinFormer}\\
(\textcolor{blue}{PSNR = 18.68}) & (\textcolor{blue}{PSNR = 14.05}) & (\textcolor{blue}{PSNR = 17.55}) & (\textcolor{blue}{PSNR = 18.90})\\
\end{tabular}
\caption{Qualitative performance of the various binarization techniques on sample no. 15 (shown in (a)) from DIBCO 2019.}
\label{fig:compDIBCO2019}
\end{figure*}

From Figs.~\ref{fig:compDIBCO2011_1} and \ref{fig:compDIBCO2011_2}, it can be inferred that our model produces more promising results in comparison to the results produced by the existing transformer-based model for document image enhancement. We can be fairly conclusive, that the proposed network makes good use of the local transformer encoder block to gather pixel-level information from the input image, and this information proved to be a valuable input to the overall encoded feature vector. 
Next, we perform a qualitative comparative analysis on example images taken from the DIBCO 2013 dataset. The binarization results of the proposed model and the other techniques are illustrated side-by-side in Fig.~\ref{fig:compDIBCO2013_1}. The comparisons shown in this example are a good example of stamp stain degradation, respectively. Next, we further test our model on images taken from the DIBCO 2017, 2018, and the most recent 2019 datasets. The output of the proposed model and the other techniques is shown in Figs.~\ref{fig:compDIBCO2017}, \ref{fig:compDIBCO2018}, \ref{fig:compDIBCO2019}, respectively. Based on the comparison shown in Figs.~\ref{fig:compDIBCO2017}, \ref{fig:compDIBCO2018} and \ref{fig:compDIBCO2019}, we can evidently see how the two-level transformer encoder captures the features more efficiently in comparison to the existing ViT-based model. While the thresholding, GAN, and CNN-based models captured a lot of noise, the previously proposed simple ViT-based model ensured that not capture noise but discarded the useful information. 

The combination of the patch and the sub-patch level feature representations ensured that the proposed method effectively recorded just the relevant information and successfully eliminated noise or deterioration. Thus the proposed network produces more efficient results in binarization tasks in comparison to the existing ViT-based technique. In conclusion, the qualitative comparison shown in Fig.\ref{fig:Output_Handwritten} to \ref{fig:compDIBCO2019} clearly demonstrates the proposed model’s superiority in recovering severely damaged document images over peer models. 
\begin{table*}[ht!]
\centering
\caption{Ablation study of the effectiveness of the proposed two-level network on the DIBCO 2017 dataset.}
\footnotesize
\begin{tabular}{|l|c|c|c|c|c|r|}
\toprule
    \textbf{Patch Size}   &\textbf{Sub-Patch Size}     &\textbf{Global Dim}        &\textbf{Local Dim}       &\textbf{Global Layers}   &\textbf{Local Layers}    &\textbf{PSNR$\uparrow$} \\ \midrule
    $16\times16$     &$8\times8$     &1024      &256     &12   &4    &20.70\\
    $16\times16$     &$8\times8$     &768      &256     &6   &12    &20.54\\
    $\mathbf{16\times16}$     &$\mathbf{8\times8}$    &\textbf{768}      &\textbf{256}     &\textbf{6}   &\textbf{4}    &\textbf{20.93}\\
    $16\times16$     &$4\times4$     &256      &256     &6   &4    &18.25\\
    $8\times8$     &$4\times4$     &768      &768     &6   &4    &18.96\\
    \hline
\end{tabular}
\label{Tab:Table9}
\end{table*}
\subsection{Ablation Study}
\label{sub_sec:Ablation}
For ablation experiments, we used sample images from the DIBCO 2017 for testing while the other DIBCO and H-DIBCO datasets were the training samples. \\
\textbf{Improvement on the multi-head attention.} From the experiments, it was observed that higher embedding dimensions produced better results with other factors remaining constant. Not many changes were noticed in increasing or decreasing the layers of encoders used, and a reasonable balance between performance and computational complexity was found by using 6 layers for the global encoder and 4 for the local encoder. \\
\textbf{Improvement on the patch and sub-patch.} As far as the patch size is concerned not much could be tested due to computational limitations, but from the limited testing that was done we found that decreasing the patch size below 16 had an adverse effect on the performance of the proposed architecture. We speculate that the reason for this inference is the lack of spatial context within the extremely small sub-patches of size $(4\times4)$. Such small sub-patches do not get the exposure required to learn the subtle differences that separate signal from noise, which later on manifests in the produced results as misinterpretation of the signal as noise or vice-versa. This reduces the overall quality of the binarized results. 

Furthermore, experimentation can be carried out by varying the above hyper-parameters. The sub-patches can also be modified for conducting experimental analysis to say, ($2\times2$) but it can be computationally very expensive to train such a large model if we create too small sub-patches. 
\section{Conclusion}
\label{sec:Con}
In this article, we proposed a novel two-stage - global and local transformer network for document image binarization, called \textbf{DocBinFormer}. The proposed network uses a transformer-in-transformer architecture for efficient binarization of degraded document images which is trained in a complete end-to-end manner. We discussed in detail how the model captures both patch-level as well as local pixel-level feature dependencies through a widely used attention mechanism from an image to improve the quality of binarization. We also discussed the importance of local attention in ViT in-depth through our study. We illustrated the entire proposed network both graphically and mathematically and provided a step-by-step explanation of the working principles of the model. To the best of our knowledge, \textbf{DocBinFormer} is the first two-level ViT-based approach that can capture and utilize both global and local attention. We carried out both quantitative and qualitative experimental analysis over the various real-world document image binarization datasets, each with its unique set of challenges, and the proposed model was successfully able to tackle most of them and achieved state-of-the-art results. Extensive experiments illustrate that the \textbf{DocBinFormer} model yields significantly better results in comparison with state-of-the-art.
\subsection{Acknowledgments}
The authors would like to express their sincere gratitude to Mr. Bappaditya Shome for helping in proofreading and revising this paper. The authors also thank Mr. Preetam Ghosh for helping with PyTorch coding. We thank Mohamed Ali Souibgui and Sana Khamekhem Jemni for sharing the experimental results.

\bibliographystyle{IEEEtran}
\bibliography{Refs}
\end{document}